\documentclass{article}
\usepackage{microtype}
\usepackage{graphicx}
\usepackage{subfigure}
\usepackage{booktabs} %

\usepackage{hyperref}

\usepackage[accepted]{icml2023}

\usepackage{amsmath}
\usepackage{amssymb}
\usepackage{mathtools}
\usepackage{amsthm}

\usepackage{enumitem}

\usepackage[capitalize,noabbrev]{cleveref}

\usepackage[textsize=tiny]{todonotes}

\usepackage{threeparttable}
\usepackage{multicol,multirow}
\usepackage{booktabs} %
\usepackage{bbding}

\usepackage{xspace}

\usepackage{bbm}

\usepackage{balance}

\usepackage{enumitem}

\usepackage{booktabs}
\usepackage{makecell}
\usepackage{multirow}
\usepackage{sail}
\usepackage{diagbox}

\def\yyf{\textcolor{black}}

\def \mymethod{{EPS-AD}}
\def \mymethodwithnorm{{EPS-N}}
\def \yoonwithnorm{{S-N}}

\def\mytitle{Detecting Adversarial Data by Probing Multiple Perturbations \\ Using Expected Perturbation Score}

\icmltitlerunning{Detecting Adversarial Data by Probing Multiple Perturbations  Using Expected Perturbation Score}
\begin{document}

\twocolumn[
\icmltitle{\mytitle}

\icmlsetsymbol{equal}{*}

\begin{icmlauthorlist}
\icmlauthor{Shuhai Zhang}{equal,scut,pz}
\icmlauthor{Feng Liu}{equal,unimelb}
\icmlauthor{Jiahao Yang}{scut} 
\icmlauthor{Yifan Yang}{scut}
\icmlauthor{Changsheng Li}{BIT}
\icmlauthor{Bo Han}{HKBU}
\icmlauthor{Mingkui Tan}{scut,edu}
\end{icmlauthorlist}

\icmlaffiliation{scut}{School of Software Engineering, South China University of Technology, China}
\icmlaffiliation{unimelb}{The University of Melbourne}
\icmlaffiliation{pz}{Pazhou Laboratory, China}
\icmlaffiliation{edu}{Key Laboratory of Big Data and Intelligent Robot, Ministry of Education, China}
\icmlaffiliation{HKBU}{Department of Computer Science, Hong Kong Baptist University}
\icmlaffiliation{BIT}{Beijing Institute of Technology, Beijing, China}
\icmlcorrespondingauthor{Mingkui Tan}{mingkuitan@scut.edu.cn}
\icmlcorrespondingauthor{Bo Han}{bhanml@comp.hkbu.edu.hk}
\icmlcorrespondingauthor{Changsheng Li}{lcs@bit.edu.cn}
\icmlkeywords{Machine Learning, ICML}

\vskip 0.3in
]

\printAffiliationsAndNotice{\icmlEqualContribution} %

\begin{abstract}
Adversarial detection aims to determine whether a given sample is an adversarial one based on the discrepancy between natural and adversarial distributions. Unfortunately, estimating or comparing two data distributions is extremely difficult,  {especially in high-dimension spaces}. 
Recently, the  {\emph{gradient}} of log probability density ({a.k.a.}, score) \wrt  {the sample}  is used as an alternative statistic to compute.
However, we find that the score is sensitive in identifying adversarial samples due to insufficient information with one sample only. 
In this paper, we propose a new statistic called \textit{expected perturbation score} (EPS), which  is essentially the expected score of a sample after various perturbations. Specifically, to obtain adequate information regarding one sample, we perturb it by adding various noises
to capture its multi-view observations.
We theoretically prove that EPS is a proper statistic to compute the discrepancy between two  {samples} under mild conditions.  
In practice, we can use a pre-trained diffusion model to estimate EPS for each sample. 
Last, we propose an EPS-based adversarial detection (EPS-AD) method, in which we develop EPS-based \emph{maximum mean discrepancy} (MMD) as a metric to measure  the discrepancy  between the test sample and natural samples. 
We also prove that the EPS-based MMD between natural and adversarial samples is larger than that among natural samples. 
Extensive experiments show the superior adversarial detection performance of our \mymethod.

\end{abstract}

\section{Introduction}

Deep neural networks (DNNs) are known to be sensitive to adversarial samples that are generated by adding imperceptible perturbations to the input but may mislead the model to make unexpected predictions \citep{szegedy2013intriguing, goodfellow2014explaining,niu2022efficient,niu2023towards,li2023learning}. Adversarial samples threaten widespread machine learning systems \citep{li2014feature,ozbulak2019impact}, which raises an urgent requirement for advanced techniques to improve the robustness of models. Among them, \textit{adversarial training} introduces adversarial data into training to improve the robustness of models but suffers from significant performance degradation and high computational complexity \citep{laidlaw2020perceptual,wong2020fast}; \textit{adversarial purification} relies on generative models to purify adversarial data before classification, which still has to compromise on unsatisfactory natural and adversarial accuracy \citep{shi2021online, yoon2021adversarial, nie2022diffusion}.

In contrast, another class of defense methods, called \textit{adversarial detection}, could be achieved by detecting and rejecting adversarial examples, which are friendly to existing machine learning systems 
{due to the  {almost} lossless natural accuracy}, 
and can help to identify security-compromised input sources \citep{abusnaina2021adversarial}. Adversarial detection aims to tell whether  {a test} sample is an adversarial sample, for which the key is to find the discrepancy between the adversarial and natural distributions. However, existing adversarial detection approaches primarily train a tailored detector for specific attacks \citep{feinman2017detecting, ma2018characterizing, lee2018simple} or for a specific classifier \citep{deng2021libre}, 
{which largely overlook modeling the adversarial and natural distributions, resulting in their limited performance against unseen attacks or transferable attacks.}

\begin{figure}[t]
\centering
\includegraphics[width=0.42\textwidth]{ 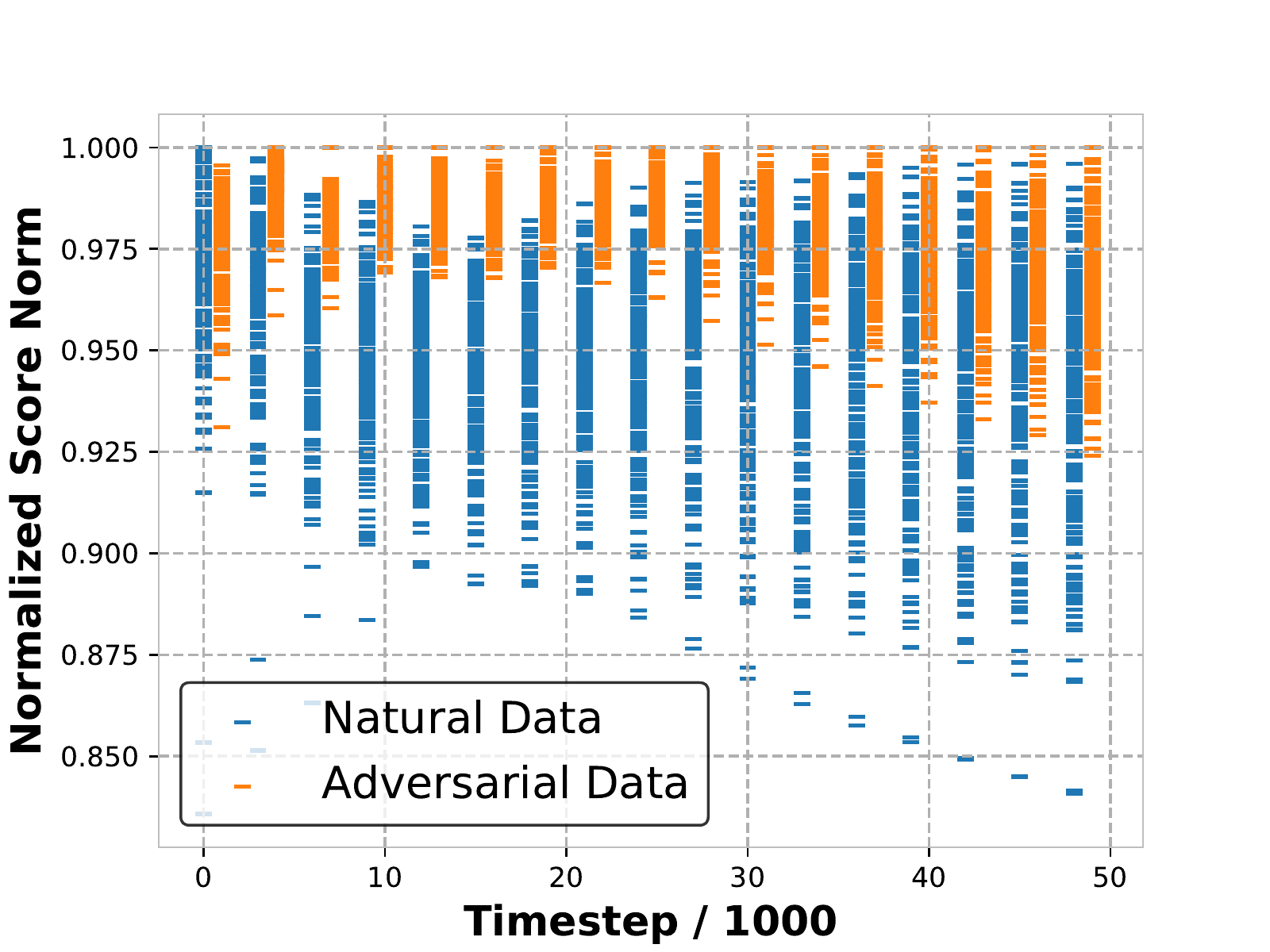}
    \vspace{-14.5pt}
    \caption{Illustration of score norms of $200$ random sampled natural and adversarial images on ImageNet,
    where most natural images have lower norms than adversarial ones at the same timestep but are sensitive to the timesteps due to the significant overlap.}    
    \label{fig: motivation}
    \vspace{-12pt}
\end{figure}

Unfortunately, it is non-trivial to estimate or compare two data distributions,  {especially in high-dimension spaces}. %
One alternative approach is to estimate the \textit{gradient} of log probability density with respect to  {the} sample, \ie, score. This statistic has emerged as a powerful means for adversarial defense \citep{yoon2021adversarial,nie2022diffusion} and diffusion models \citep{song2019generative,song2020score, huang2021variational}.
However, how to effectively exploit the score function for adversarial detection is not well studied.

Recently, \citet{yoon2021adversarial} 
purify adversarial samples by gradually removing the adversarial noise from the (attacked) samples with the score function for adversarial defense. During the purification process, they employ the \textit{norm} of scores (between being-purified adversarial samples and natural samples) to set a threshold for determining which timestep to stop purifying.
They empirically find that natural samples usually have lower score norms than  adversarial samples across purification timesteps.
{Intuitively, the score  could represent the momentum of the sample towards the high-density areas of natural data \citep{song2019generative}. 
This means that
a lower score norm indicates 
the sample is closer to the high-density areas of natural data, \ie, 
a higher probability of the sample following the natural distribution.}
To further understand this, we demonstrate the score norms of natural and adversarial samples at different purification timesteps.
According to Figure \ref{fig: motivation}, most natural samples have lower score norms than adversarial samples at the same timestep, but they are very sensitive to the timesteps due to the significant overlap across all timesteps. This suggests that the score of one sample is useful but not effective enough in identifying the adversarial samples. 

\begin{figure*}[t]
	\centering
	\includegraphics[width=0.86\linewidth]{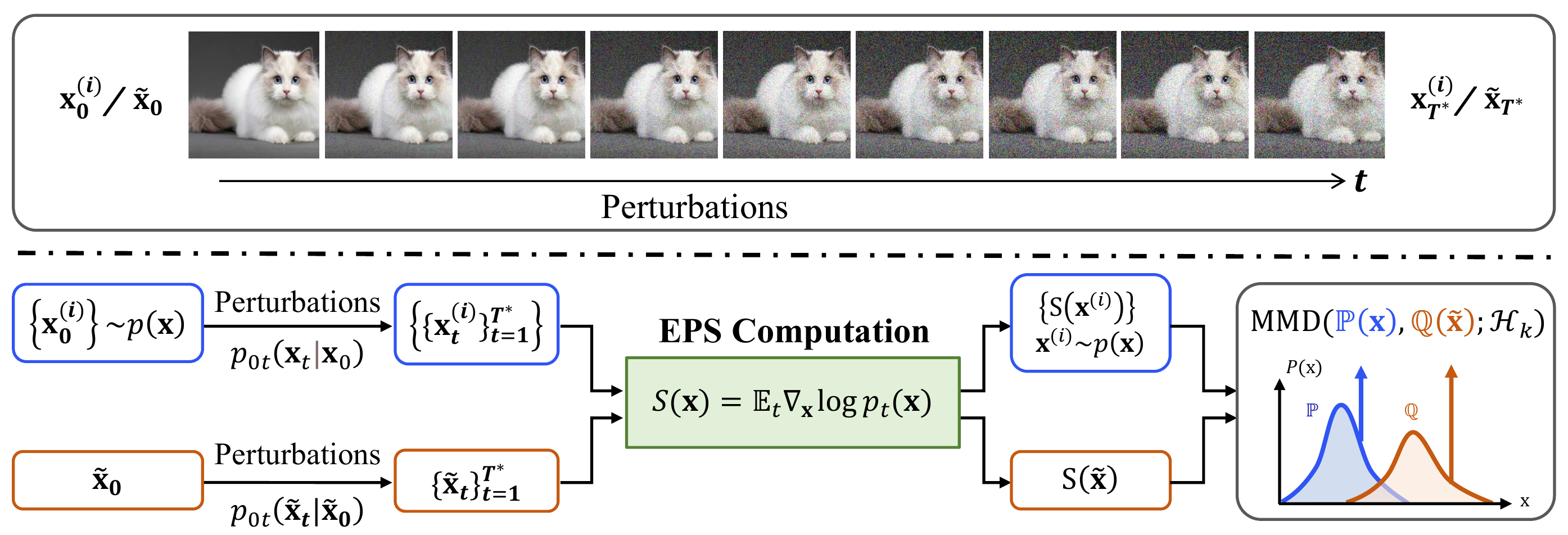}
	\vspace{-7.5pt}
	\caption{Overview of the proposed  \mymethod. EPS denotes the expected score after multiple perturbations of a sample using a pre-trained score model. Specifically, we simultaneously add perturbations to a set of natural images \yyf{$\{\bx^{(i)}_0\}$}~and  {a test} image \yyf{$\tilde{\bx}_0$} following the diffusion process with a  time step $T^*$ to get perturbed images, from which we obtain their EPSs $S(\bx)$ via the score model and calculate the MMD %
	between EPS of the  {test} sample and EPSs  of natural samples.}
	\label{overview}
	\vspace{-5.5pt}
\end{figure*}

To address the above issue,
we propose a new statistic called \textit{expected perturbation score} (EPS), which represents the expected score after multiple perturbations of a given sample. 
In EPS, we consider multiple levels of noise perturbations to diversify one sample, allowing us to capture multi-view observations of one sample and thus extract adequate information from the data. Our theoretical analysis shows 
EPS is a valid statistic in distinguishing between natural and adversarial samples under mild conditions. Thus, we propose an \textit{EPS-based adversarial detection} method (\mymethod) for adversarial detection, 
as illustrated in Figure \ref{overview}. 
Specifically, given a pre-trained score-based diffusion model, \mymethod~consists of three steps: 1) we simultaneously add multiple perturbations to a set of natural images and a  {test} image following the forward diffusion process with a  time step $T^*$; 2) we obtain their EPSs via the score model; 3) we adopt the \textit{maximum mean discrepancy} (MMD) between 
any test sample and natural samples relying on EPS. 
Theoretical and empirical evidences show the superiority of our method.

We summarize our main contributions as follows:
\begin{itemize}[leftmargin=*]
    
    \item  {A novel statistic. We find that the traditional score of one sample is sensitive in identifying adversarial samples due to insufficient information from a single sample only. However, we can perturb the sample by adding various noises and thus can obtain various information from its multi-view observations or perturbations. We propose to calculate the expected score of the multiple perturbations of the sample, namely \textit{expected perturbation score} (EPS). This new statistic can be much more reliable and effective than traditional score (\eg, \citet{yoon2021adversarial}) in measuring the distributional discrepancy. }

    \item  {A novel adversarial detection method with extensive empirical justifications. Based on the novel statistic EPS, we further develop a novel single-sample adversarial detection method called EPS-AD, in which we propose an EPS-based \textit{maximum mean discrepancy} (MMD) and then use it as a metric to measure the discrepancy between the test sample and natural samples. 
    We conduct extensive experiments on both CIFAR-10 and ImageNet across various network architectures, including ResNet, WideResNet and ViT.  Our method consistently outperforms existing methods against $12$ different attacks.}

    \item  {Theoretical justifications. First, we provide theoretical analysis that EPS is a proper statistic to distinguish between natural and adversarial data well under mild conditions (Theorem \ref{thm: adv vs natural}). Specifically, we show that the EPS of the natural sample is closer to those of other natural samples compared to adversarial samples. 
    Second, we theoretically show that the EPS-based MMD between natural and adversarial samples is larger than that among natural samples (Corollary \ref{probability of K}).}

\end{itemize}

\section{Preliminaries}
\label{Preliminaries}

\textbf{Adversarial data generation.} Given a well-trained classifier $\hat{f}$ on a data set $D{=}\{(\bx_i, l_i)\}_{i=1}^n$ with $\bx_i$ being a sample from the input space $\mX$ and $l_i$ being its ground-truth label defined in a label set $\mC = \{1,\ldots, C\}$, adversarial data $\hat{\bx}$ regarding $\bx$ with perturbation $\epsilon$ is given by 
\begin{equation}
    \hat{\bx} = \mathop{\arg\max}\limits_{\hat{\bx} \in \mathcal{B}(\bx, {\epsilon})} \ell\left(\hat{f}(\hat{\bx}), l\right),
\end{equation}
where $\mathcal{B}(\bx, {\epsilon})=\left\{\bx^{\prime} \in \mathcal{X} \mid d\left(\bx, \bx^{\prime}\right) \leq \epsilon\right\}$, $d$ is some distance (\eg, $\ell_2$ or $\ell_\infty$ distance), and $\ell$ is some loss function. For simplicity, we denote $\hat{\bx} = \bx + \boldsymbol{\varepsilon}$ as the adversarial data.

\textbf{Continuous-time diffusion models.}
Following \citet{song2020score}, let $p(\bx)$ be the unknown data distribution. Diffusion models firstly construct a forward diffusion process $\{\bx_t\}_{t=0}^{T_{\text{diff}}}$ indexed by a continuous time variable $t \in[0, T_{\text{diff}}]$, which can be modeled by a stochastic differential equation (SDE) with positive
time increments:
\begin{equation}
\label{forward-SDE}
    \mathrm{d} \bx=\bff(\bx, t) \mathrm{d} t+g(t) \mathrm{d} \bw,
\end{equation}
where $\bx_0 := \bx \sim p(\bx)$, $\mathbf{f}(\cdot, t): \mathbb{R}^{d} \rightarrow \mathbb{R}^{d}$ is a vector-valued function, $g(\cdot): \mathbb{R} \rightarrow \mathbb{R}$ is a scalar function that is independent  {of} $\bx$, and $\bw$ is a standard Wiener process.

Let $p_t(\bx)$ be the the marginal distribution of $\bx_t$ with $p_0(\bx) = p(\bx)$, if $\bff(\bx,t)$ and $g(t)$ are designed well such that $p_T(\bx) \approx \mathcal{N}\left(\mathbf{0}, \boldsymbol{I}_{d}\right)$, by reversing the diffusion process from $t =T_{\text{diff}}$ to $t =0$, we can reconstruct samples $\bx_0 \sim p_0(\bx)$. The reverse process is given by the reverse-time SDE:
\begin{equation}
\label{reverse-SDE}
    \mathrm{d} \mathbf{x}=\left[\mathbf{f}(\mathbf{x}, t)-g(t)^{2} \nabla_{\mathbf{x}} \log p_{t}(\mathbf{x})\right] \mathrm{d} t+g(t) \mathrm{d} \bar{\mathbf{w}},
\end{equation}
where $\bar{\bw}$ is a standard reverse-time Wiener process and $\mathrm{d} t$ is an infinitesimal negative time step. Throughout the paper, we consider the VP-SDE following the setting of  \citet{song2020score}, where $\bff(\bx, t):=-\frac{1}{2} \beta(t) \bx$ and $ g(t):=\sqrt{\beta(t)}$ with $\beta(t)$ being the linear noise schedule, \ie, $\beta(t):=\beta_{\text {min }}+\left(\beta_{\text {max }}-\beta_{\text {min }}\right) t/T_{\text{diff}}$ for $t\in[0,T_{\text{diff}}]$.

Reconstructing samples from the Gaussian distribution requires the score of the marginal distribution, \ie, $\nabla_{\mathbf{x}} \log p_{t}(\mathbf{x})$ in the reverse process Eq. (\ref{reverse-SDE}). To estimate the score function $\nabla_{\mathbf{x}} \log p_{t}(\mathbf{x})$, one effective solution is to train a score model $\mathbf{s}_{\boldsymbol{\theta}}(\mathbf{x}, t)$ on samples with score matching \citep{hyvarinen2005estimation, song2019generative, vincent2011connection}. The training objective function is minimizing:
\begin{equation}%
\label{objective}
    \mathbb{E}_{t }\!\!\left[ \lambda(t)\mathbb{E}_{p_{0}\left(\mathbf{x}_{0}\right) p_{0t}\left(\bx_{t} \mid \bx_0\right)} \!\!\left\|s_{\boldsymbol{\theta}}\left(\mathbf{x}_{t}, t\right)\!{-}\!\nabla_{\mathbf{x}_{t}} \!\log p_{0 t}\!\left(\mathbf{x}_{t} {\mid} \mathbf{x}_{0}\right)\right\|_{2}^{2}\right]\!,
\end{equation}
where $\lambda(t)$ is a weight \wrt $t$ and $p_{0 t}(\mathbf{x}_{t} \mid \mathbf{x}_{0})$ is a transition kernel from $\bx_0$ to $\bx_t$ that can be derived  by Eq. (\ref{forward-SDE}).
 
\textbf{Maximum mean discrepancy.}
Following \citet{gretton2012kernel, borgwardt2006integrating}, let $\mX \subset \mmR^d$ be a separable metric space and $p$, $q$ be Borel probability measures on $\mX$. Given two independent identically distributed (\textit{IID}) observations $S_X = \{\bx^{(i)}\}_{i=1}^n $ and $S_Y = \{\by^{(i)}\}_{i=1}^m $ from distributions $p$ and $q$, respectively, \textit{maximum mean discrepancy} (MMD) aims to measure
the closeness between these two distributions, which is defined as:
\begin{equation}
\operatorname{MMD}(p, q ; \mathcal{F}):=\sup _{f \in \mathcal{F}}|\mathbb{E}[f(X)]-\mathbb{E}[f(Y)]|,
\end{equation}
where $\mF$ is a class of functions $f: \mathcal{X} \rightarrow \mathbb{R}$, and $X \sim p, Y \sim q$ are two random variables. To better study the richness of the MMD function class $\mF$, \citet{borgwardt2006integrating} propose to choose $\mF$ to be the unit ball in a universal reproducing kernel Hilbert space and obtain the following kernel-based MMD,
\begin{equation}
\begin{aligned}
& \operatorname{MMD}\left(p, q ; \mathcal{H}_k\right):=\sup _{f \in \mathcal{H},\|f\|_{\mathcal{H}_k} \leq 1}|\mathbb{E}[f(X)]-\mathbb{E}[f(Y)]| \\
=&\left\|\mu_{p}-\mu_{q}\right\|_{\mathcal{H}_k}{=}\sqrt{\mathbb{E}\left[k\left(X, X^{\prime}\right){+}k\left(Y, Y^{\prime}\right){-}2 k(X, Y)\right]},
\end{aligned}
\end{equation}
where $k: \mathcal{X} \times \mathcal{X} \rightarrow \mathbb{R}$ is the kernel of a reproducing kernel Hilbert space $\mH_k$, $\mu_{p}:=\mathbb{E}[k(\cdot, X)]$ and $\mu_{q}:=\mathbb{E}[k(\cdot, Y)]$ are the kernel mean embeddings of $p$ and $q$, respectively \citep{gretton2012kernel, borgwardt2006integrating,jitkrittum2017linear, liu2020learning, gao2021maximum}. 
In addition, \citet{liu2020learning} propose deep-kernel-based MMD to select the adaptive kennel for two-sample tests.

\textbf{Adversarial detection.}
In this paper, we aim to address the following adversarial detection problem. %
\begin{Prob}
\label{Def: Adversarial Detection}
Let $\mX \subset \mmR^d$ be a separable metric space and $p$ be a Borel probability measure on $\mX$, and \textit{IID} observations $S_X = \{\bx^{(i)}\}_{i=1}^n $  from the distribution $p$ and a ground-truth labeling mapping $f{:}~ \mathbb{R}^{d} {\rightarrow} \mathcal{C}$ with $\mathcal{C}{=}\{1, \ldots, C\}$ being a label set. Assuming that the attacker has access to some well-trained classifier $\hat{f}$ on $S_X$ and \textit{IID} observations $S'_X$ from the distribution $p$, we wish to know whether each new sample in $S_{Y}{=}\left\{\by^{(i)}\right\}_{i=1}^{m}$ that are crafted with $S'_X$ is following the distribution $p$.
\end{Prob}

Note that the definition of \textit{adversarial detection} in Problem \ref{Def: Adversarial Detection} is different from that in two-sample test \citep{grosse2017statistical}. Particularly, Problem \ref{Def: Adversarial Detection} aims to determine whether each example in $S_{Y}$ is sampled from the distribution $p$, while two-sample test aims to tell if two populations $S_{X}$ and $S_{Y}$ come from the same distribution, which focuses on the closeness between two populations.

\section{Expected Perturbation Score for Adversarial Detection}
\label{sec: method}
 {Existing methods (\eg, \citet{yoon2021adversarial}) seek to use the score (the gradient of log probability density \wrt the sample) of one sample to measure the discrepancy between natural and adversarial distributions for adversarial detection.  {However, this measure is not accurate enough for detecting a single test sample, as the insufficient information available for a single sample can result in highly fluctuating distributional discrepancies}. Moreover, 
although they have empirically demonstrated some results about the norm of score in distinguishing adversarial samples from natural samples, yet do not provide deep investigations or theoretical analysis on it.}

 {To address the above issues, we consider obtaining more information of a sample using its multi-view perturbations and propose a new statistic called \textbf{expected perturbation score} (EPS) to measure the distributional discrepancy (Section \ref{sec: Expected Perturbation Score}). Based on this new statistic, we develop an \textbf{EPS-based adversarial detection} method, called \mymethod, as shown in Figure \ref{overview}. Particularly,  we estimate the EPS for each sample using a pre-trained diffusion model, then adopt the \textit{maximum mean discrepancy} (MMD) between the test sample and a set of natural samples relying on EPS
as  {a characteristic of the test sample} for adversarial detection (Section \ref{sec: EPGD}). Last, we provide theoretical analyses that the MMD between EPSs of the natural samples is smaller than that between natural and adversarial samples (Section \ref{sec: thm for EPGD}).}

\subsection{Expected Perturbation Score}
\label{sec: Expected Perturbation Score}

To begin with, we provide the definition of \textit{expected perturbation score} (EPS) to develop our proposed method.

\begin{deftn}
\label{Def: expected perturbation score}
\textbf{(\emph{Expected perturbation score)}}
Let $\mX \subset \mmR^d$ be a separable metric space and $p$ be Borel probability measure on $\mX$. Given a perturbation process transition distribution $p_{0 t}\left(\mathbf{x}_{t} \mid \mathbf{x}_{0}\right)$  {from $\bx_0$ to $\bx_t$}, the expected perturbation score (EPS) of a sample $\bx  {\sim p}$ is given by:
\begin{equation}
\label{Eq: expected perturbation score}
    S\left(\mathbf{x}\right)=\mathbb{E}_{t \sim U(0, T)} \nabla_{\mathbf{x}} \log p_t\left(\mathbf{x}\right), 
\end{equation}
 {where $p_t(\bx)$ is the marginal probability distribution of $\bx_t$ with $p_0(\bx):= p(\bx)$}
and $T$ is the last perturbation timestep.
\end{deftn}

\begin{remark}
Note that the perturbation process transition distribution $p_{0 t}\left(\mathbf{x}_{t}\mid \mathbf{x}_{0}\right)$ can be any distribution, such as Gaussian distribution or uniform distribution.  {Moreover, {unlike only computing a single score of one sample proposed by \citet{song2020score}},
we incorporate multiple levels of noise  {through $p_{0 t}\left(\mathbf{x}_{t}\mid \mathbf{x}_{0}\right)$ at different timesteps $t$} in the definition of EPS, with the aim of diversifying the single sample, which enables {us} to capture multi-view observation data and thus fully exploit more information from the data.}
\end{remark}

Built upon Definition \ref{Def: Adversarial Detection}, we derive a following theorem to give a closer look on EPS $S(\bx)$ for the natural and adversarial data when the perturbation process transition distribution $p_{0 t}\left(\mathbf{x}_{t}\mid \mathbf{x}_{0}\right)$ and $p(\bx)$ are from Gaussian distributions.

\begin{thm}
\label{thm: adv vs natural}
    Assuming that the distribution of natural data $p(\bx){=} \mN({\boldsymbol{\mu}}_{\bx}, \sigma_{\bx}^2\mathbf{I})$,  {where $\mathbf{I}$ is an identity matrix}, given a perturbation transition kernel $p_{0 t}\left(\mathbf{x}_{t} \mid \mathbf{x}_{0}\right)=\mathcal{N}\left( \gamma_{t} \mathbf{x}_{0}, \sigma_{t}^{2} \mathbf{I}\right)$
    with $\gamma_{t}$ and $\sigma_{t}$ being the time-dependent noise schedule, then  the following three conclusions hold: \\
    1) For $\forall{~}\bx \sim p(\bx)$, $S(\bx) \sim \mN({\bf{0}}, \sigma_S^2{\bI})$;\\
    2) For $\forall{~}\by \sim p(\bx)$ and adversarial sample $\hat{\by} {=} \by {+} \boldsymbol{\varepsilon}$, $S(\hat{\by}) \sim \mN(-{\boldsymbol{\mu}}_{S}, \sigma_S^2{\bI})$;\\
    3) For $\forall~\bx, \by \sim p(\bx)$ and adversarial sample $\hat{\by} {=} \by {+} \boldsymbol{\varepsilon}$, 
    \begin{equation}
    \label{natural diff}
        S(\bx) - S(\by) \sim \mN({\bf{0}}, 2\sigma_S^2{\bI});
    \end{equation}
    \begin{equation}
    \label{adver diff}
        S(\bx) - S(\hat{\by}) \sim \mN({\boldsymbol{\mu}}_{S}, 2\sigma_S^2{\bI}),
    \end{equation}
    where ${\boldsymbol{\mu}}_{S}{=}\mmE_{t \sim U(0, T)}\boldsymbol{\mu}_{t}$ with $\boldsymbol{\mu}_{t} = \frac{ \boldsymbol{\varepsilon}}{\gamma_t^2\sigma_{\bx}^2{+}\sigma_t^2}$ and $\sigma_{S}^2=\mmE_{t \sim U(0, T)} \zeta_{t}^{2} $ with $\zeta_{t}^{2} = \frac{1}{\gamma_t^2\sigma_{\bx}^2+\sigma_t^2} $.
\end{thm}

\begin{remark}
     {Note that analyzing scores (\ie, gradients of log probability) under complex practical distributions can be very difficult and infeasible. Nevertheless, we still hope to deeply investigate our EPS from a theoretical perspective and see whether it has some advantages over existing ones in theory. To this end,  we adopt a common practice by assuming that the data follows some Gaussian distribution.}
\end{remark}
   
Theorem \ref{thm: adv vs natural} tells us: 1) the first two findings show that the mean of EPS for the adversarial sample differs from that of the natural sample due to the additional term ${\boldsymbol{\mu}}_{S}$;
2) the third finding indicates that the EPS of the natural sample is closer to that of other natural samples compared to adversarial samples, and this discrepancy becomes more pronounced when the  perturbation transitions $\gamma_t$ and $\sigma_t$ are small.
This motivates us to employ $S(\bx)$ for adversarial detection. 

  {\textbf{Necessity of multiple scores in EPS.} 
 Note that $\zeta_{t}^{2}$ and   {$\|\boldsymbol{\mu}_{t}\|^2$} decrease as the timestep $t$ increases due to the increase of $\gamma_t$ and $\sigma_t$. However, smaller variance $\sigma_{S}^{2}$ and larger mean  {$\|{\boldsymbol{\mu}}_{S}\|^2$} are required for good adversarial detection. If we only consider one score of some unique timestep $t$ (\ie, removing the expectation from the definition of EPS), the variance $\sigma_{S}^{2}$ and mean $\boldsymbol{\mu}_{S}$ of the discrepancy will be so fluctuant that performing detecting adversarial samples will be very sensitive to the timestep $t$ (as validated in Figure \ref{fig: motivation} and Section \ref{Ablation}). To alleviate this issue, we consider taking expectation w.r.t. the timestep on multiple scores. In this way, the distribution of the discrepancy between the natural sample and the adversarial sample will be more stable to the timestep, which makes it easier to obtain a superior solution.}

\textbf{Estimation for EPS.}
Note that EPS (\ie, $S(\bx)$) in Eq.~(\ref{Eq: expected perturbation score}) requires knowing the score function $ \nabla_{\mathbf{x}} \log p_t\left(\mathbf{x}\right) $, which can be estimated by training a neural network with score matching \citep{song2019generative, vincent2011connection}. To this end, we model the perturbation process transition as a Gaussian distribution $p_{0 t}\left(\mathbf{x}_{t} \mid \mathbf{x}_{0}\right)=\mathcal{N}\left( \gamma_{t} \mathbf{x}_{0}, \sigma_{t}^{2} \mathbf{I}\right)$, where $\gamma_{t}=e^{-\frac{1}{2} \int_{0}^{t} \beta(s) d s}$ and $ \sigma_{t}^{2}=1-e^{-\int_{0}^{t} \beta(s) d s} $ with $\beta(t)$ for $t \in [0,1000]$ being the time-dependent noise schedule.  {By optimizing Eq. (\ref{objective}), with sufficient data and model capacity, score matching ensures that the optimal solution to Eq. (\ref{objective}) equals $ \nabla_{\mathbf{x}} \log p_t\left(\mathbf{x}\right) $ for almost $\bx$ and $t$ \citep{song2020score}. As a result, the score $ \nabla_{\mathbf{x}} \log p_t\left(\mathbf{x}\right) $ can be approximated by $s_{\theta}(\bx_t, t)$. 
In practice, we use a pre-trained diffusion model to achieve the estimation for the score $s_{\theta}(\bx_t, t)$. }

\subsection{Exploring EPS for Adversarial Detection}
\label{sec: EPGD}

Using the norm of estimated EPS $S(\bx)$ as a characterization for adversarial detection is straightforward. Nevertheless, 
this manner
can only describe the magnitude of the EPS vector, so it ignores the rich information that can be derived from its direction. It is critical that we design a distance metric to measure the distance between the EPS of a  {test} sample and the EPSs of natural samples in order to derive more useful information from $S(\bx)$. To this end, we propose  an EPS-based adversarial detection method (EPS-AD) with \textit{maximum mean discrepancy} (MMD). 

Benefiting from the superior performance of MMD in measuring two given distributions \citep{long2015learning,zhu2019aligning}, we resort to it for characterizing EPS. The basic idea of MMD is that two distributions would be identical if two random variables  {have identical moments} for any order, and the moment that makes the largest distance between the two distributions should be the measure of the two distributions when the two distributions are not the same \citep{ gong2022born}. 
 {To formalize our approach, we consider two sample sets: $\mathbb{P}_{X}=\{\bx^{(i)}\}_{i=1}^n$ for natural samples and $\mathbb{Q}_{Y}=\{\tilde{\bx}\}$ for a test sample. Employing the MMD, we estimate the distance between $\mathbb{P}_{X}$ and $\mathbb{Q}_{Y}$ as follows:}
\begin{equation}
\begin{aligned}   
\label{MMD}    \widehat{\operatorname{MMD}}_{b}^{2}\left[\mathbb{P}_{X}, \mathbb{Q}_{Y} ; \mathcal{H}_{k}\right]=\!\frac{1}{n^{2}}&\! \sum_{i, j=1}^{n} \!\!k\left(S(\mathbf{x}^{(i)}), S(\mathbf{x}^{(j)})\right)\\
    {-}\frac{2}{n}  \sum_{i=1}^{n}  k\!\left(S(\mathbf{x}^{(i)}), S(\mathbf{y})\right)&{+}k \left(S(\mathbf{y}), S(\mathbf{y})\right),
\end{aligned} 
\end{equation}

where $k$ is the kernel of a reproducing kernel Hilbert space $\mH_k$ such as the Gaussian kernel $k \left(\ba, \bb\right)=\exp \left(-\left\|\ba-\bb\right\|^{2} /\left(2 \sigma^{2}\right)\right)$. In practice, we consider deep kernel MMD for our method following \citet{liu2020learning}  {and refer readers to Supplementary \ref{sec:implementation of deep-kernel-MMD} for more details}. In this way, we can obtain a distance between the test sample and a set of natural samples for adversarial detection.

\subsection{Theoretical Analysis for  \mymethod}
\label{sec: thm for EPGD}
Note that after performing the same perturbation process, the first and the third terms in Eq. (\ref{MMD}) are the same for each test sample  {in most cases such as $k$ is the Gaussian kernel}, thus we only focus on the cross-term $J = \frac{2}{n} \sum_{i=1}^{n} k\left(S(\mathbf{x}^{(i)}), S(\mathbf{y})\right)$ for the test sample.
\begin{coll}
\label{probability of K}
    Considering the Gaussian kernel $k \left(\ba, \bb\right){=}\exp \left(-\left\|\ba-\bb\right\|^{2} /\left(2 \sigma^{2}\right)\right)$ and the assumption in Theorem \ref{thm: adv vs natural}, for $\forall 0 {<} \eta {<}1$, the probability of $P\{k\left(S(\bx), S(\hat{\by})\right){>}\eta\}$ is given by 
    \begin{equation}
        P\{k\left(S(\bx), S(\hat{\by})\right){>}\eta\} = \int_0^C \chi^2_d(z)dz,
    \end{equation}
    where $z = \|{\boldsymbol{\mu}_S}\|^2$ with ${\boldsymbol{\mu}_S} $ being the mean of $S(\bx)-S(\hat{\by})$, C is a constant for given $\eta$ and $\sigma$, $\chi^2$ is the probability density function of noncentral chi-squared distribution with $d$ degrees of freedom \citep{abdel1954approximate}.
\end{coll}

Corollary \ref{probability of K} indicates that the cross-term $J$ will be larger if $\left\| {\boldsymbol{\mu}_S}\right\|^2$ is close to zero given an $\eta$. Combining  Eq. (\ref{natural diff}) and Eq. (\ref{adver diff}), we conclude that the natural data have larger $J$ than the adversarial data with higher probability due to the additional term $\mmE_t \frac{\boldsymbol{\varepsilon}}{\gamma_t^2\sigma_{\bx}^2{+}\sigma_t^2}$, suggesting that  the MMD between EPSs of the natural samples is smaller than that between natural and adversarial samples.  %

\begin{figure*}[h]

    \begin{center}
        \subfigure[Attack Method: PGD]
        {\includegraphics[width=0.32\textwidth]{ 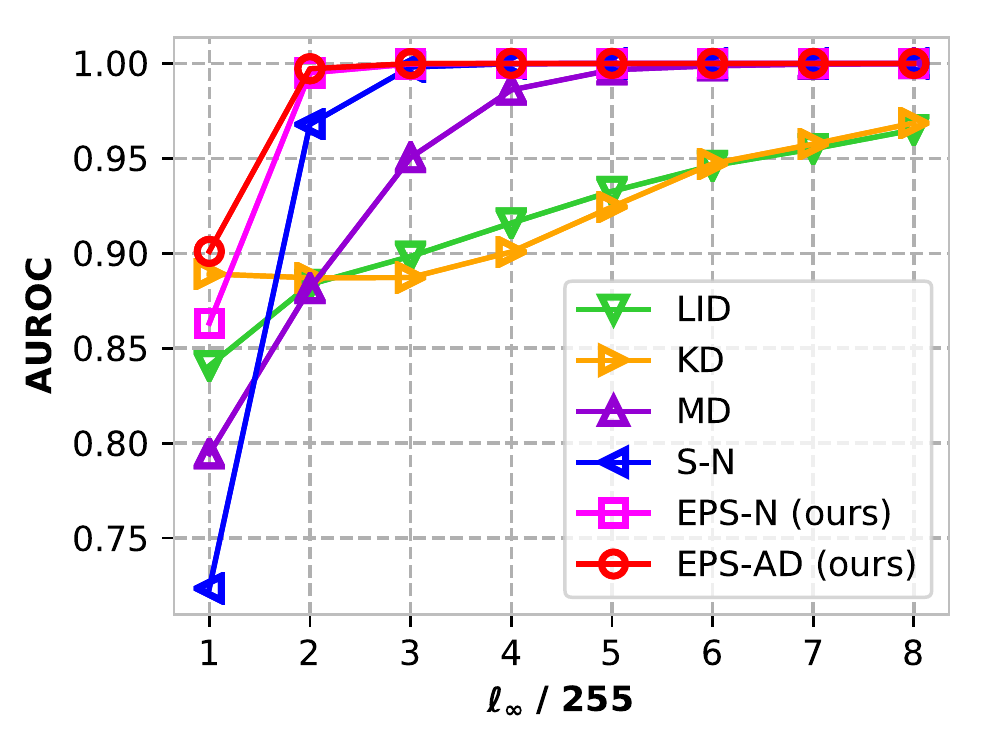}}
        \subfigure[Attack Method: FGSM]
        {\includegraphics[width=0.32\textwidth]{ 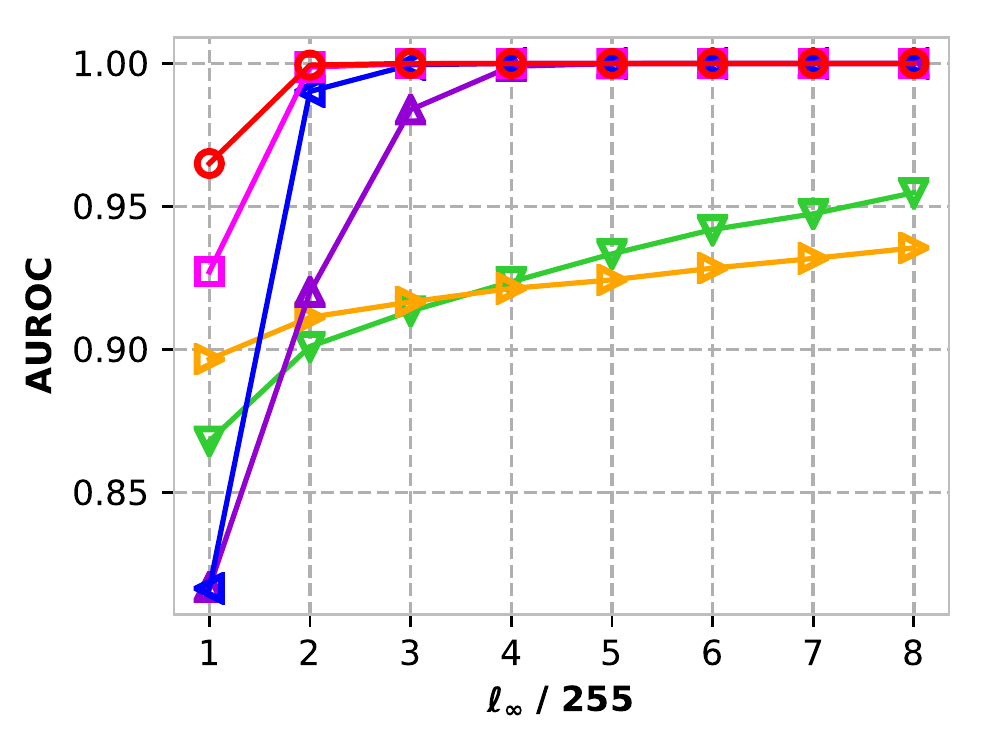}}
        \subfigure[Attack Method: CW]
        {\includegraphics[width=0.32\textwidth]{ 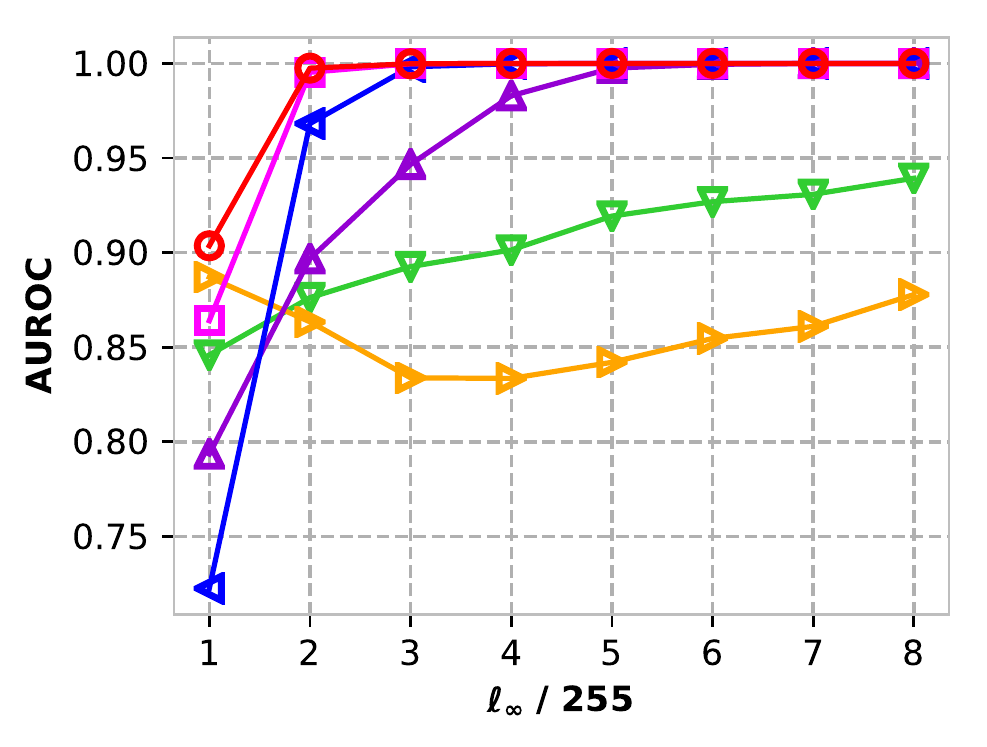}}
        \subfigure[Attack Method: BIM]
        {\includegraphics[width=0.32\textwidth]{ 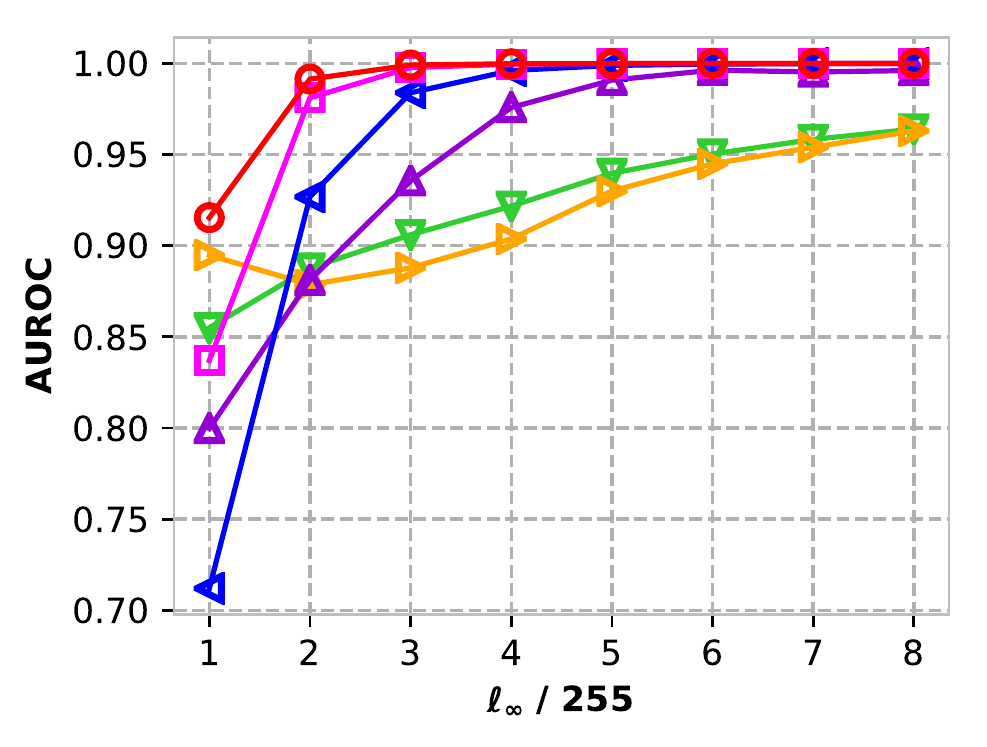}}
        \subfigure[Attack Method: FGSM-$\ell_2$]
        {\includegraphics[width=0.32\textwidth]{ 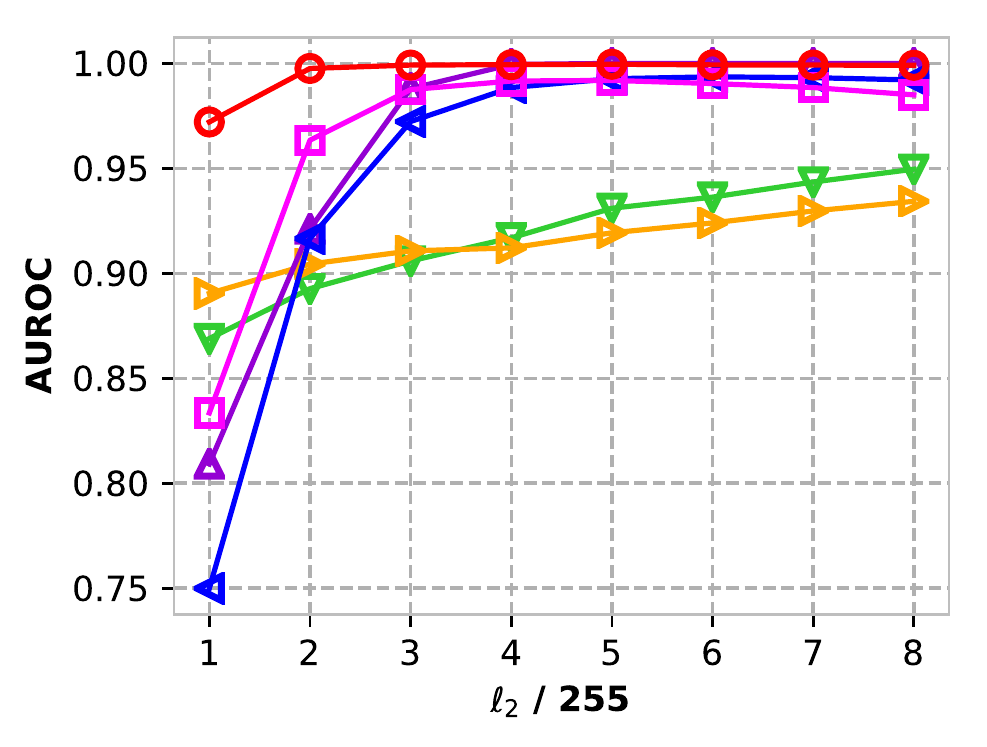}}
        \subfigure[Attack Method: AA Attack]
        {\includegraphics[width=0.32\textwidth]{ 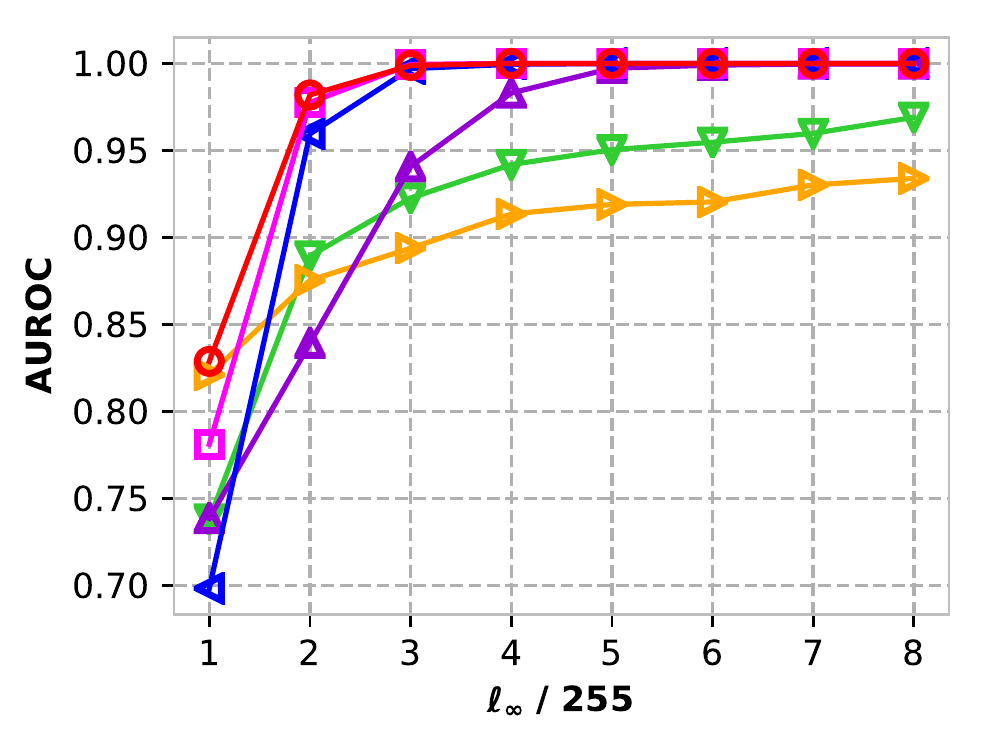}}

        \vspace{-0.9em}
        \caption{
        Comparison with adversarial detection methods on CIFAR-10 in terms of AUROC under  {$\epsilon {\in}\{1/255,{\ldots},8/255\}$}  
        against PGD, FGSM, CW, BIM, FGSM-$\ell_2$, AA. 
        Sub-figures (a) - (f) share the same legend presented in sub-figure (a).
     }
        \label{fig:comparison_cifar10}
    \end{center}
    \vspace{-1.6em}
\end{figure*}

\begin{table*}[t]
    \caption{Comparison of \yyf{different adversarial detection} methods on CIFAR-10 in terms of AUROC under $\epsilon = 4/255$. The bold number indicates the best results.}
    \centering
    \small
    \begin{tabular*}{14cm}{@{}@{\extracolsep{\fill}}c|ccccccc@{}}
    \toprule
        AUROC & FGSM & PGD & BIM & CW & FGSM-$\ell_{2}$ & BIM-$\ell_{2}$ & AA \\ \midrule
        KD & 0.9213 & 0.9007 & 0.9037 & 0.8335 & 0.9121 & 0.9107 & 0.9135 \\ 
         LID & 0.9236 & 0.9159 & 0.9217 & 0.9014 & 0.9169 & 0.9320 & 0.9419 \\ 
         MD & 0.9990 & 0.9860 & 0.9758 & 0.9829 & \textbf{0.9995} & 0.9586 & 0.9830 \\ 
         \yoonwithnorm & \textbf{1.0000} & 0.9998 & 0.9961 & 0.9998 & 0.9885 & 0.9674 & 0.9995 \\ 
         \mymethodwithnorm~(Ours) & \textbf{1.0000} & \textbf{1.0000} & 0.9996 & \textbf{1.0000} & 0.9916 & 0.9883 & \textbf{1.0000} \\ 
         \mymethod~(Ours) & \textbf{1.0000} & \textbf{1.0000} & \textbf{0.9998} & \textbf{1.0000} & \textbf{0.9995} & \textbf{0.9991} & \textbf{1.0000} \\ \bottomrule
    \end{tabular*}
    \vspace{-0.6em}
    \label{AUROC:cifar}
\end{table*}

\section{Experiments}
 {In this section, we start by providing experimental settings (Section \ref{exp_settings}). Based on various strong attack benchmarks, we then compare our method with state-of-the-art adversarial detection approaches (Sections \ref{exp_knowattack} and \ref{unknown_attack}). Additionally, we conduct various ablation studies for further understanding of our approach (Section \ref{Ablation}). Our code is available at \href{https://github.com/ZSHsh98/EPS-AD.git}{https://github.com/ZSHsh98/EPS-AD.git}.}

\subsection{Experimental Settings}
\label{exp_settings}
\textbf{Datasets and network architectures.}
We \yyf{evaluate our method on} CIFAR-10~\citep{krizhevsky2009learning} and ImageNet~\citep{deng2009imagenet}. We implement three widely used architectures \yyf{as classifiers}: WideResNet~\citep{zagoruyko2016wide,gowal2021improving} for CIFAR-10, ResNet~\citep{he2016deep} and ViT~\citep{dosovitskiy2021image} for ImageNet. For diffusion models, we consider the pre-trained diffusion models of Score SDE \citep{song2020score} for CIFAR-10 and Guided Diffusion \citep{dhariwal2021diffusion} for ImageNet, respectively.

\textbf{Attack methods.}
Following \citet{deng2021libre}, we evaluate our adversarial detection method under \yyf{ various attack methods. We consider} the commonly used  $\ell_2$ and $\ell_\infty$ threat models, including PGD \citep{madry2017towards}, FGSM \citep{goodfellow2014explaining}, BIM \citep{kurakin2018adversarial}, MIM \citep{dong2018boosting}, TIM \citep{dong2019evading}, CW \citep{carlini2017towards}, DI\_MIM\citep{xie2019improving}. 
Moreover, we apply two adaptive attack methods such as AutoAttack (AA) \citep{croce2020reliable} and Minimum-Margin Attack (MM) \citep{gao2022fast}.  To show the superiority of our method, we consider the relatively low attack intensities, \ie, $\ell_2$-ball and $\ell_\infty$-ball  with $\epsilon = 4/255$, and iterative attacks run for $5$ steps using step size $\epsilon / 5$, unless stated otherwise. 

\textbf{Baselines.}
We compare our method with several state-of-the-art adversarial detection methods, including kernel density (KD) \citep{feinman2017detecting}, local intrinsic dimensionality (LID) \citep{ma2018characterizing}, mahalanobis distance (MD) \citep{lee2018simple} and LiBRe \citep{deng2021libre}. 
Besides, we construct two new adversarial detection methods based on diffusion models: 1) \yoonwithnorm: using the score norm of raw images, \ie, $||s_{\theta}(\bx, t)||^2$.
2) {\mymethodwithnorm}: using the norm of the EPS, \ie, $||S(\bx)||^2$. \yyf{Differently, our proposed \mymethod~further {calculates} the \textit{maximum mean discrepancy} of EPSs.}

\begin{figure*}[tp]
\label{fig: imagenet}
    \begin{center}
        \subfigure[Attack Method: PGD]
        {\includegraphics[width=0.32\textwidth]{ 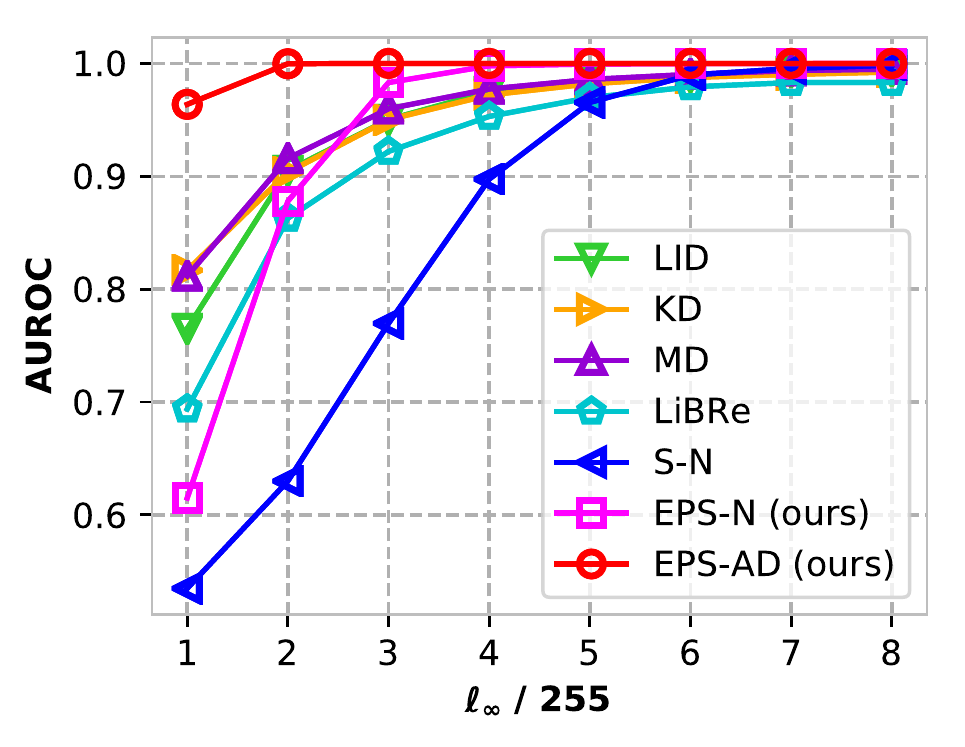}}
        \subfigure[Attack Method: FGSM]
        {\includegraphics[width=0.32\textwidth]{ 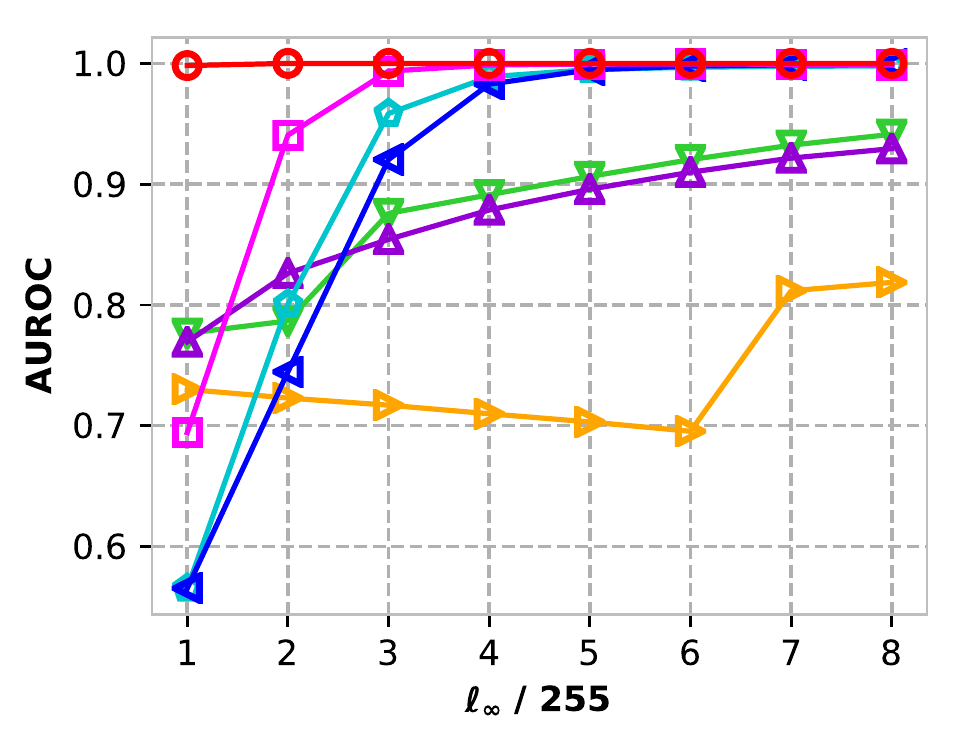}}
        \subfigure[Attack Method: CW]
        {\includegraphics[width=0.32\textwidth]{ 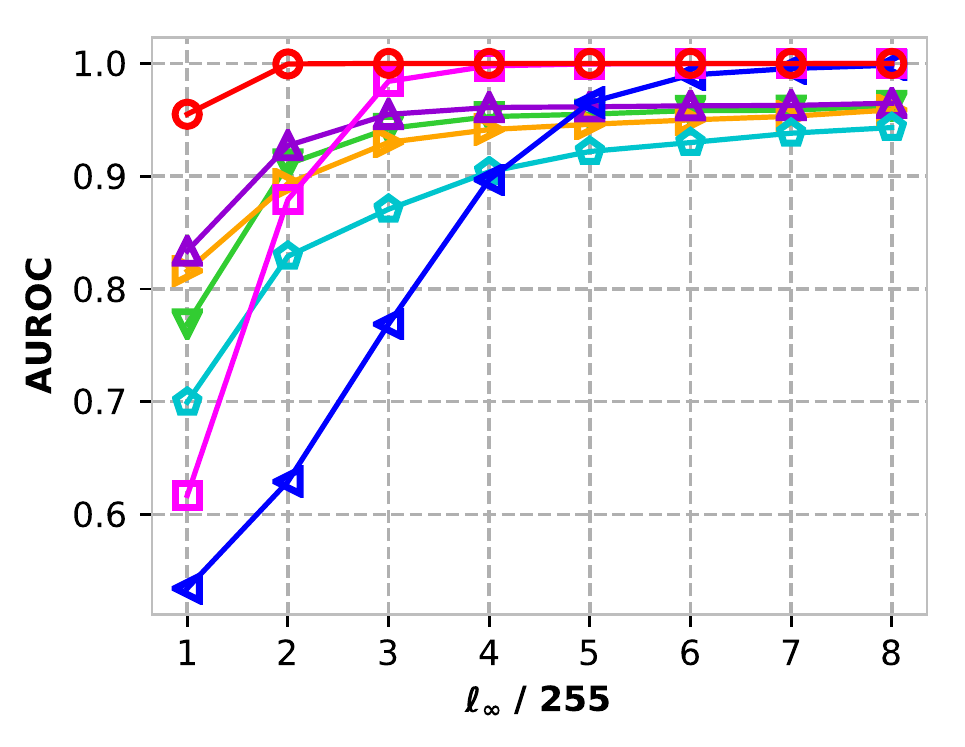}}
        \subfigure[Attack Method: BIM]
        {\includegraphics[width=0.32\textwidth]{ 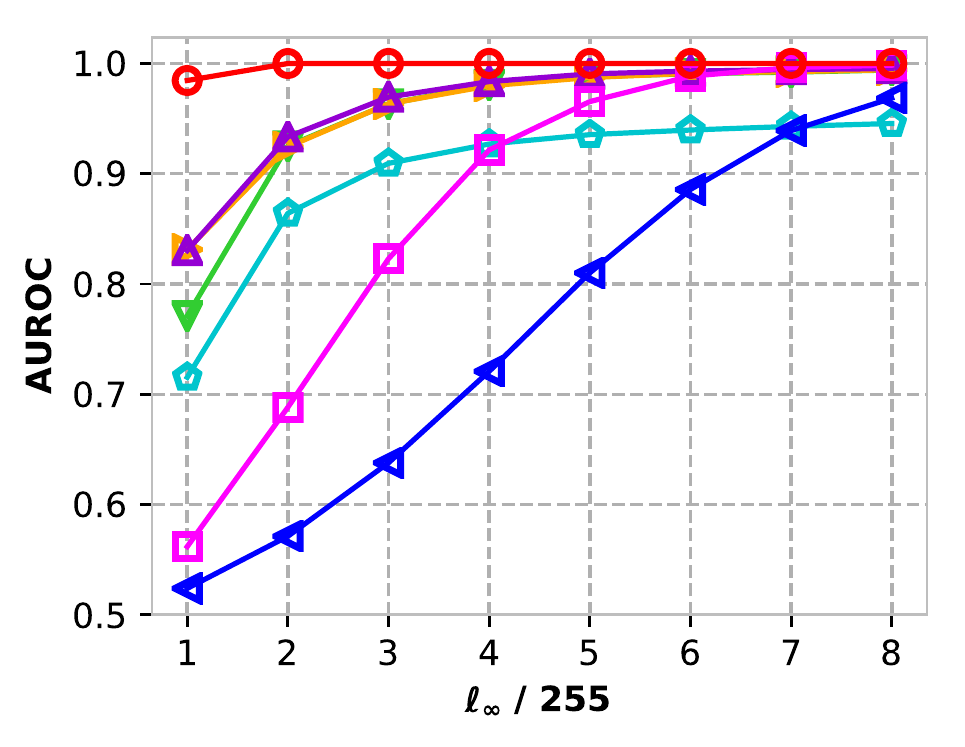}}
        \subfigure[Attack Method: FGSM-$\ell_2$]
        {\includegraphics[width=0.32\textwidth]{ 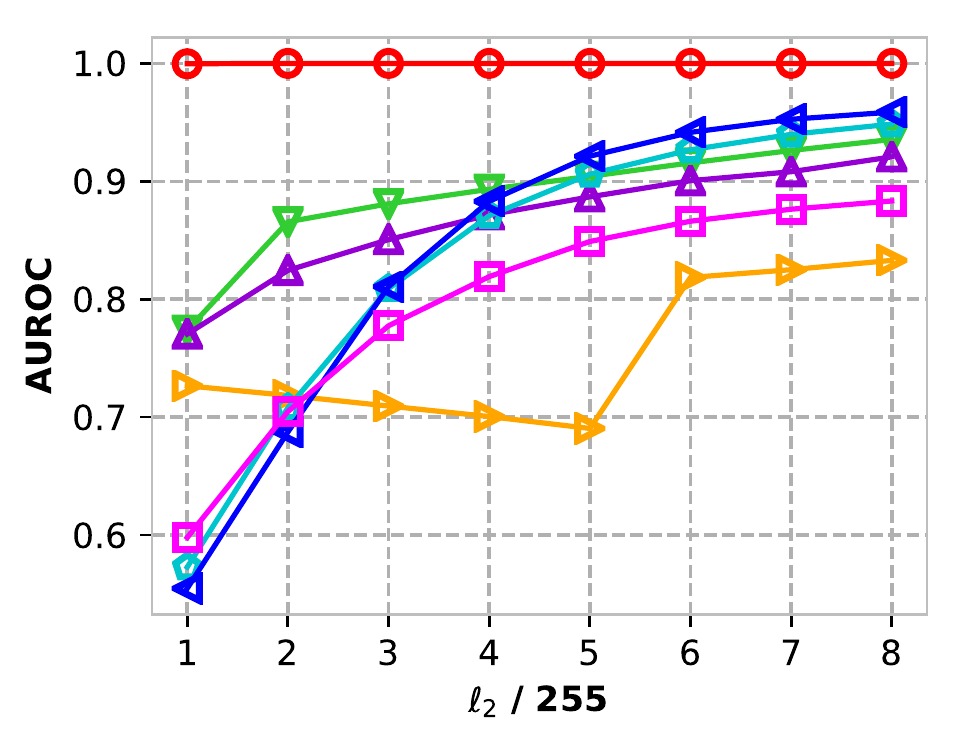}}
        \subfigure[Attack Method: AA Attack]
        {\includegraphics[width=0.32\textwidth]{ 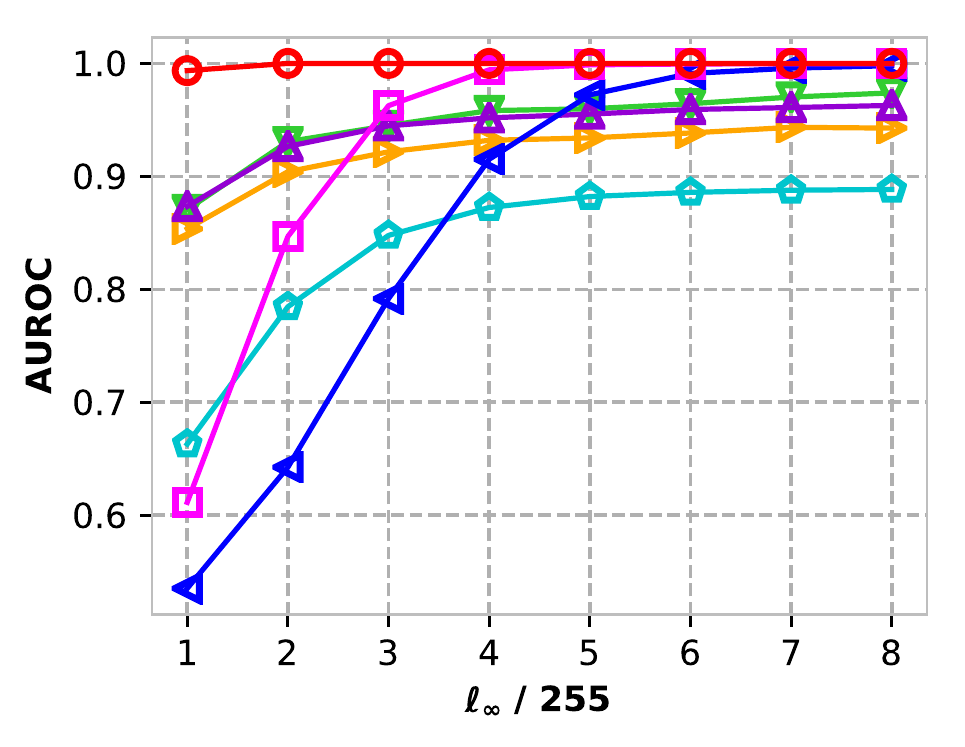}}

        \vspace{-1em}
        \caption{  Comparison with adversarial detection methods on ImageNet in terms of AUROC under  {$\epsilon {\in}\{1/255,{\ldots},8/255\}$}  
        against PGD, FGSM, CW, BIM, FGSM-$\ell_2$, AA. 
        Sub-figures (a) - (f) share the same legend presented in sub-figure (a).
      }
        \label{fig:comparison_ImageNet}
    \end{center}
    \vspace{-1.6em}
\end{figure*}

\begin{table*}[ht]

    \caption{Comparison of \yyf{different adversarial detection} methods on ImageNet in terms of AUROC under $\epsilon = 4/255$. The bold number indicates the best results.}
    \centering
    \small
    \begin{tabular*}{14cm}{@{}@{\extracolsep{\fill}}c|ccccccc@{}}
    \toprule
        AUROC & FGSM & PGD & BIM & CW & FGSM-$\ell_{2}$ & BIM-$\ell_{2}$ & AA \\ \midrule
        KD & 0.7099 & 0.9720 & 0.9797 & 0.9413 & 0.7004 & 0.9775 & 0.9319 \\
        LID & 0.8912 & 0.9750 & 0.9808 & 0.9528 & 0.8932 & 0.9816 & 0.9582 \\
        MD & 0.8786 & 0.9773 & 0.9835 & 0.9609 & 0.8715 & 0.9830 & 0.9518 \\
        LiBRe & 0.9889 & 0.9530 & 0.9269 & 0.9039 & 0.8708 & 0.9211 & 0.8724 \\
        \yoonwithnorm & 0.9828 & 0.8974 & 0.7208 & 0.8969 & 0.8830 & 0.6762 & 0.9151 \\
        \mymethodwithnorm~(Ours) & 0.9987 & 0.9978 & 0.9215 & 0.9978 & 0.8191 & 0.7172 & 0.9943 \\
        \mymethod~(Ours) & \textbf{1.0000} & \textbf{1.0000} & \textbf{1.0000} & \textbf{1.0000} & \textbf{1.0000} & \textbf{1.0000} & \textbf{1.0000} \\ \bottomrule
    \end{tabular*}
    \vspace{-0.5em}
    \label{AUROC:image}
\end{table*}

\textbf{Evaluation metric.}
We evaluate the performance of adversarial detection approaches with the area under the receiver operating characteristic (AUROC),  {which is a widely used statistic for assessing the discriminatory capacity of distribution models \citep{jimenez2012insights}.} 
Considering the computational cost of applying 12 attacks to the classifier, especially for the ImageNet, following \citet{gao2021maximum}, 
we randomly select two disjoint subsets as adversarial and natural samples (each containing 500 samples) and compute the AUROC value over these two subsets.
\yyf{Notably, our method is applicable to different data set sizes, which is verified in Supplementary.} Moreover, we set $T^*=20$ in $S(x)$ on CIFAR-10 and $T^*=50$  on ImageNet for both our \mymethod~and \mymethodwithnorm, and set $t^*=5$ in $s_{\theta}(\bx, t)$ on CIFAR-10 and $t^*=20$ on ImageNet for \yoonwithnorm.

\subsection{Detecting on Known Attacks}
\label{exp_knowattack}
We start by comparing \mymethod~with the state-of-the-art adversarial detection methods that are trained with seen adversarial examples, against $\ell_2$ and $\ell_\infty$ threat models. \yyf{Considering adversarial detection becomes more challenging when the attack intensity of adversarial samples  is low},
to broadly evaluate the adversarial detection performance, we compare our method with other baselines on different attacks under different attack intensities. Moreover, to show the best performance of KD, LID and MD, we test the detection performance on corresponding adversarial examples.

\begin{table*}[t]
    \vspace{-2pt}
    \caption{Comparison of AUROC for detecting unseen attacks on CIFAR-10, where ``FGSM (seen)''  denotes the seen adversarial attack used for the training of KD, LID and MD.}
    \label{unseen}
    \centering
    \small
    \begin{tabular*}{14cm}{@{}@{\extracolsep{\fill}}c|ccccccc@{}}
    \toprule
         {AUROC} & FGSM(seen) & PGD & BIM & CW & FGSM-$\ell_{2}$ (seen)& BIM-$\ell_{2}$ & AA \\ \midrule
        KD & 0.9213 & 0.9007 & 0.9082 & 0.8339 & 0.9146 & 0.9146 & 0.9135 \\
        LID & 0.9236 & 0.8964 & 0.9028 & 0.8828 & 0.9160 & 0.8984 & 0.9253 \\
        MD & 0.9990 & 0.9855 & 0.9742 & 0.9835 & 0.9992 & 0.9503 & 0.9820 \\
        \yoonwithnorm & \textbf{1.0000} & 0.9998 & 0.9961 & 0.9998 & 0.9885 & 0.9674 & 0.9995 \\
        \mymethodwithnorm~(Ours) & \textbf{1.0000} & \textbf{1.0000} & 0.9996 & \textbf{1.0000} & 0.9916 & 0.9883 & \textbf{1.0000} \\
        \mymethod~(Ours) & \textbf{1.0000} & \textbf{1.0000} & \textbf{0.9998} & \textbf{1.0000} & \textbf{0.9995} & \textbf{0.9991} & \textbf{1.0000} \\ \bottomrule
    \end{tabular*}
    \vspace{-1.8em}
\end{table*}

\begin{table*}[t]
    \caption{Comparison of AUROC for detecting transferable attacks on ImageNet, where  KD, LID,  MD and LiBRe are trained with adversarial examples with ResNet-50 but detect the adversarial examples  crafted with ResNet-101.}
    \label{transferred}
    \centering
    \small
    \begin{tabular*}{14cm}{@{}@{\extracolsep{\fill}}c|ccccccc@{}}
    \toprule
         {AUROC} & FGSM & PGD & BIM & CW & FGSM-$\ell_{2}$ & BIM-$\ell_{2}$ & AA \\ \midrule
        KD & 0.7754 & 0.5999 & 0.5847 & 0.7632 & 0.7906 & 0.7756 & 0.7698 \\
        LID & 0.8467 & 0.7627 & 0.7663 & 0.7704 & 0.8520 & 0.7925 & 0.7967 \\
        MD & 0.8467 & 0.7698 & 0.7684 & 0.7665 & 0.8067 & 0.7759 & 0.7880 \\
        \textit LiBRe & 0.9849 & 0.8414 & 0.7161 & 0.8286 & 0.8489 & 0.7250 & 0.8485 \\
        \yoonwithnorm & 0.9816 & 0.8965 & 0.7166 & 0.8963 & 0.8764 & 0.6705 & 0.9106 \\
        \mymethodwithnorm~(Ours) & 0.9983 & 0.9975 & 0.9178 & 0.9979 & 0.8235 & 0.7215 & 0.9930 \\
        \mymethod~(Ours) & \textbf{1.0000} & \textbf{1.0000} & \textbf{1.0000} & \textbf{1.0000} & \textbf{1.0000} & \textbf{0.9998} & \textbf{1.0000} \\ \bottomrule
    \end{tabular*}
    \vspace{-0.2em}
\end{table*}

\textbf{Results on CIFAR-10.}
Figure \ref{fig:comparison_cifar10} shows our adversarial detection performance against $6$ attacks under different attack intensities  {$\epsilon {\in}\{1/255,{\ldots},8/255\}$} on CIFAR-10 over WideResNet-28-10 \citep{zagoruyko2016wide} compared to other baselines. We demonstrate other attacks and more results on WideResNet-70-16 \citep{gowal2021improving} in Supplementary. Obviously, our \mymethod~have much higher AUROC performance than other methods. \yyf{Critically, we observe} that our \mymethod~preserves almost non-degraded AUROC  {when the attack intensity $\epsilon$ surpasses $2/255$ against $\ell_2$ and $\ell_\infty$ attacks,}
\yyf{which shows the stability of \mymethod~when detecting challenge adversarial samples.}

In addition, we report quantitative results for adversarial detection under the attack intensity $\epsilon = 4/255$ in Tables \ref{AUROC:cifar} and \ref{AUROC:image}. \yyf{The results show that by employing EPS  and measuring their MMD, \mymethod~consistently outperforms existing methods against various attacks in terms of AUROC. We also see that by simply applying the norm of EPS, \mymethodwithnorm~achieves superior adversarial detection performance, which demonstrates the effectiveness of EPS.}

\textbf{Results on ImageNet.} 
\yyf{We report the adversarial detection performance against $\ell_2$ and $\ell_\infty$ attacks on ImageNet over ResNet-50 \citep{he2016deep} in  Tables~\ref{AUROC:cifar}, \ref{AUROC:image} and Figure \ref{fig:comparison_ImageNet} . We defer the results of one widely-used ViT architecture, DeiT-S~\citep{dosovitskiy2021image}, in Supplementary. We observe that our approach consistently outperforms baselines under various attacks,
especially for detecting PGD, FGSM-$\ell_2$ and AA attacks. 
These results reveal that our proposed 
\mymethod~is effective even on a large-scale data set.} Moreover, we observe that \mymethodwithnorm~exhibits poor results compared to \mymethod~when detecting $\ell_2$ attacks (\eg, FGSM-$\ell_2$ and BIM-$\ell_2$) since the norm of EPS ignores rich information  contained in the EPS vector, which is not effective enough to detect
on a large-scale data set.

\subsection{Detecting on Unseen and  Transferable Attacks}
\label{unknown_attack}

\yyf{
In light of  poor performance for adversarial detection baselines against unseen attacks and transferable attacks, we evaluate our method in the context of these kinds of attacks. }

\textbf{Detecting on unseen attacks.}
To detect unseen attacks, we train KD, LID, and MD detectors on CIFAR-10 using only FGSM and FGSM-$\ell_2$ adversarial examples and evaluate their performance under other attacks. Combining Tables \ref{unseen} and \ref{AUROC:cifar}, \ref{AUROC:image}, we find that adversarial detection performance for MD and LID worsens. An explanation is that their detectors are trained with vectorized features extracted from the seen samples through logistic regression, resulting in their limited generalization on unseen attacks. \yyf{In contrast, diffusion-based detection methods show superior performance since they directly model the distribution of natural data.}

\textbf{Detecting on transferable attacks.}
To validate the transferability, we train KD, LID, MD and LiBRe detectors with ResNet-50 but detect the adversarial examples from a surrogate ResNet-101 model. Comparing Tables \ref{transferred} and \ref{AUROC:cifar}, \ref{AUROC:image}, the non-{diffusion}-based methods (\eg, KD, LID, MD and LiBRe) drop significantly against transferable attacks. By contrast, our \mymethod~method achieves significantly better transferability, since it does not rely on specific classifiers, but rather models the distribution of natural data, indicating its versatility in various attack scenarios.

\begin{figure}[t]
\centering
\includegraphics[width=0.39\textwidth]{ 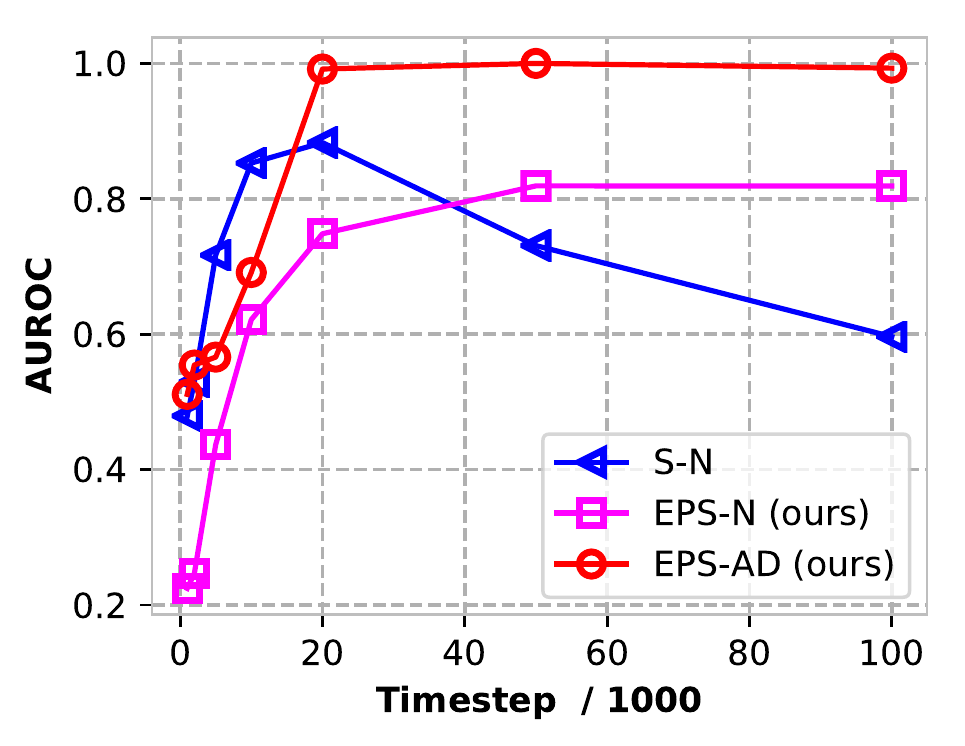}
    \vspace{-12pt}
    \caption{Impact of timestep on ImageNet under different timesteps  $\{1,2,5,10, 20,50,100\}$ against FGSM-$\ell_2$.}    
    \label{fig: ablation_imagenet}
    \vspace{-20pt}
\end{figure}

\subsection{Ablation Study on Impact of Timestep}
\label{Ablation}

We conduct experiments on ImageNet over ResNet-50 to show the effect of the timestep. 
 {To this end, we set the total timestep $T = 100$, which is sufficient for both EPS-AD and EPS-N to achieve a good solution.}
As shown in Figure \ref{fig: ablation_imagenet}, we draw two observations: 1) our \mymethod~and \mymethodwithnorm~are insensitive to the  {total} timestep $T$ while \yoonwithnorm~fluctuates greatly with the timestep $t$; 2) As the  {total} timestep $T$ increases, our \mymethod~and \mymethodwithnorm~exhibit progressively better performance, however, this gain gradually decreases when $T$ exceeds the optimal value.
This is due to the larger diffusion time, the mean ${\boldsymbol{\mu}}_{S}$ in Eq. (\ref{adver diff}) gradually approaches  zeros, resulting in a smaller discrepancy between the natural and adversarial distributions. 
This phenomenon is consistent with the result in \citet{nie2022diffusion} (Theorem 1).

\section{Conclusion}
In this paper, we propose a new statistic \textit{expected perturbation score} (EPS) to capture the information from multi-view observations of one sample, which is able to distinguish between natural and adversarial data well. Relying on EPS, we propose a novel adversarial detection method, \mymethod. We provide theoretical analysis to demonstrate the superiority of \mymethod. Extensive experiments on CIFAR-10 and ImageNet across different network architectures including ResNet, WideResNet and ViT show our \mymethod~successfully detects adversarial samples  in various attack scenarios.

\section*{Acknowledgements}
This work was partially supported by the 
National Natural Science Foundation of China (NSFC) (62072190, 62122013), Key-Area Research and Development Program Guangdong Province 2018B010107001, Program for Guangdong Introducing Innovative and Enterpreneurial Teams 2017ZT07X183, NSFC Young Scientists Fund No. 62006202, and Guangdong Basic and Applied Basic Research Foundation No. 2022A1515011652.

\balance
\bibliography{reference}

\begin{thebibliography}{58}
\providecommand{\natexlab}[1]{#1}
\providecommand{\url}[1]{\texttt{#1}}
\expandafter\ifx\csname urlstyle\endcsname\relax
  \providecommand{\doi}[1]{doi: #1}\else
  \providecommand{\doi}{doi: \begingroup \urlstyle{rm}\Url}\fi

\bibitem[Abdel-Aty(1954)]{abdel1954approximate}
Abdel-Aty, S.
\newblock Approximate formulae for the percentage points and the probability
  integral of the non-central $\chi$ 2 distribution.
\newblock \emph{Biometrika}, 41\penalty0 (3/4):\penalty0 538--540, 1954.

\bibitem[Abusnaina et~al.(2021)Abusnaina, Wu, Arora, Wang, Wang, Yang, and
  Mohaisen]{abusnaina2021adversarial}
Abusnaina, A., Wu, Y., Arora, S., Wang, Y., Wang, F., Yang, H., and Mohaisen,
  D.
\newblock Adversarial example detection using latent neighborhood graph.
\newblock In \emph{Proceedings of the IEEE/CVF International Conference on
  Computer Vision}, pp.\  7687--7696, 2021.

\bibitem[Borgwardt et~al.(2006)Borgwardt, Gretton, Rasch, Kriegel,
  Sch{\"o}lkopf, and Smola]{borgwardt2006integrating}
Borgwardt, K.~M., Gretton, A., Rasch, M.~J., Kriegel, H.-P., Sch{\"o}lkopf, B.,
  and Smola, A.~J.
\newblock Integrating structured biological data by kernel maximum mean
  discrepancy.
\newblock \emph{Bioinformatics}, 22\penalty0 (14):\penalty0 e49--e57, 2006.

\bibitem[Carlini \& Wagner(2017)Carlini and Wagner]{carlini2017towards}
Carlini, N. and Wagner, D.
\newblock Towards evaluating the robustness of neural networks.
\newblock In \emph{2017 IEEE Symposium on Security and Privacy (SP)}, pp.\
  39--57. IEEE, 2017.

\bibitem[Croce \& Hein(2020)Croce and Hein]{croce2020reliable}
Croce, F. and Hein, M.
\newblock Reliable evaluation of adversarial robustness with an ensemble of
  diverse parameter-free attacks.
\newblock In \emph{Proceedings of the 37th International Conference on Machine
  Learning}, pp.\  2206--2216. PMLR, 2020.

\bibitem[Deng et~al.(2009)Deng, Dong, Socher, Li, Li, and
  Fei-Fei]{deng2009imagenet}
Deng, J., Dong, W., Socher, R., Li, L.-J., Li, K., and Fei-Fei, L.
\newblock Imagenet: A large-scale hierarchical image database.
\newblock In \emph{2009 IEEE Conference on Computer Vision and Pattern
  Recognition}, pp.\  248--255. IEEE, 2009.

\bibitem[Deng et~al.(2021)Deng, Yang, Xu, Su, and Zhu]{deng2021libre}
Deng, Z., Yang, X., Xu, S., Su, H., and Zhu, J.
\newblock Libre: A practical bayesian approach to adversarial detection.
\newblock In \emph{Proceedings of the IEEE/CVF Conference on Computer Vision
  and Pattern Recognition}, pp.\  972--982, 2021.

\bibitem[Dhariwal \& Nichol(2021)Dhariwal and Nichol]{dhariwal2021diffusion}
Dhariwal, P. and Nichol, A.
\newblock Diffusion models beat gans on image synthesis.
\newblock \emph{Advances in Neural Information Processing Systems},
  34:\penalty0 8780--8794, 2021.

\bibitem[Dong et~al.(2018)Dong, Liao, Pang, Su, Zhu, Hu, and
  Li]{dong2018boosting}
Dong, Y., Liao, F., Pang, T., Su, H., Zhu, J., Hu, X., and Li, J.
\newblock Boosting adversarial attacks with momentum.
\newblock In \emph{Proceedings of the IEEE Conference on Computer Vision and
  Pattern Recognition}, pp.\  9185--9193, 2018.

\bibitem[Dong et~al.(2019)Dong, Pang, Su, and Zhu]{dong2019evading}
Dong, Y., Pang, T., Su, H., and Zhu, J.
\newblock Evading defenses to transferable adversarial examples by
  translation-invariant attacks.
\newblock In \emph{Proceedings of the IEEE/CVF Conference on Computer Vision
  and Pattern Recognition}, pp.\  4312--4321, 2019.

\bibitem[Dosovitskiy et~al.(2021)Dosovitskiy, Beyer, Kolesnikov, Weissenborn,
  Zhai, Unterthiner, Dehghani, Minderer, Heigold, Gelly,
  et~al.]{dosovitskiy2021image}
Dosovitskiy, A., Beyer, L., Kolesnikov, A., Weissenborn, D., Zhai, X.,
  Unterthiner, T., Dehghani, M., Minderer, M., Heigold, G., Gelly, S., et~al.
\newblock An image is worth 16x16 words: Transformers for image recognition at
  scale.
\newblock In \emph{International Conference on Learning Representations}, 2021.

\bibitem[Feinman et~al.(2017)Feinman, Curtin, Shintre, and
  Gardner]{feinman2017detecting}
Feinman, R., Curtin, R.~R., Shintre, S., and Gardner, A.~B.
\newblock Detecting adversarial samples from artifacts.
\newblock \emph{arXiv preprint arXiv:1703.00410}, 2017.

\bibitem[Gao et~al.(2021)Gao, Liu, Zhang, Han, Liu, Niu, and
  Sugiyama]{gao2021maximum}
Gao, R., Liu, F., Zhang, J., Han, B., Liu, T., Niu, G., and Sugiyama, M.
\newblock Maximum mean discrepancy test is aware of adversarial attacks.
\newblock In \emph{International Conference on Machine Learning}, pp.\
  3564--3575. PMLR, 2021.

\bibitem[Gao et~al.(2022)Gao, Wang, Zhou, Liu, Xie, Niu, Han, and
  Cheng]{gao2022fast}
Gao, R., Wang, J., Zhou, K., Liu, F., Xie, B., Niu, G., Han, B., and Cheng, J.
\newblock Fast and reliable evaluation of adversarial robustness with
  minimum-margin attack.
\newblock In \emph{Proceedings of the 39th International Conference on Machine
  Learning}, pp.\  7144--7163. PMLR, 2022.

\bibitem[Gong et~al.(2022)Gong, Xiang, Liu, and Zhou]{gong2022born}
Gong, L.-H., Xiang, L.-Z., Liu, S.-H., and Zhou, N.-R.
\newblock Born machine model based on matrix product state quantum circuit.
\newblock \emph{Physica A: Statistical Mechanics and its Applications},
  593:\penalty0 126907, 2022.

\bibitem[Goodfellow et~al.(2015)Goodfellow, Shlens, and
  Szegedy]{goodfellow2014explaining}
Goodfellow, I.~J., Shlens, J., and Szegedy, C.
\newblock Explaining and harnessing adversarial examples.
\newblock In \emph{International Conference on Learning Representations}, 2015.

\bibitem[Gowal et~al.(2021)Gowal, Rebuffi, Wiles, Stimberg, Calian, and
  Mann]{gowal2021improving}
Gowal, S., Rebuffi, S.-A., Wiles, O., Stimberg, F., Calian, D.~A., and Mann,
  T.~A.
\newblock Improving robustness using generated data.
\newblock \emph{Advances in Neural Information Processing Systems},
  34:\penalty0 4218--4233, 2021.

\bibitem[Gretton et~al.(2012)Gretton, Borgwardt, Rasch, Sch{\"o}lkopf, and
  Smola]{gretton2012kernel}
Gretton, A., Borgwardt, K.~M., Rasch, M.~J., Sch{\"o}lkopf, B., and Smola, A.
\newblock A kernel two-sample test.
\newblock \emph{Journal of Machine Learning Research}, 13\penalty0
  (1):\penalty0 723--773, 2012.

\bibitem[Grosse et~al.(2017)Grosse, Manoharan, Papernot, Backes, and
  McDaniel]{grosse2017statistical}
Grosse, K., Manoharan, P., Papernot, N., Backes, M., and McDaniel, P.
\newblock On the (statistical) detection of adversarial examples.
\newblock \emph{arXiv preprint arXiv:1702.06280}, 2017.

\bibitem[He et~al.(2016)He, Zhang, Ren, and Sun]{he2016deep}
He, K., Zhang, X., Ren, S., and Sun, J.
\newblock Deep residual learning for image recognition.
\newblock In \emph{Proceedings of the IEEE Conference on Computer Vision and
  Pattern Recognition}, 2016.

\bibitem[Huang et~al.(2021)Huang, Lim, and Courville]{huang2021variational}
Huang, C.-W., Lim, J.~H., and Courville, A.~C.
\newblock A variational perspective on diffusion-based generative models and
  score matching.
\newblock \emph{Advances in Neural Information Processing Systems},
  34:\penalty0 22863--22876, 2021.

\bibitem[Hyv{\"a}rinen \& Dayan(2005)Hyv{\"a}rinen and
  Dayan]{hyvarinen2005estimation}
Hyv{\"a}rinen, A. and Dayan, P.
\newblock Estimation of non-normalized statistical models by score matching.
\newblock \emph{Journal of Machine Learning Research}, 6\penalty0 (4), 2005.

\bibitem[Jim{\'e}nez-Valverde(2012)]{jimenez2012insights}
Jim{\'e}nez-Valverde, A.
\newblock Insights into the area under the receiver operating characteristic
  curve (auc) as a discrimination measure in species distribution modelling.
\newblock \emph{Global Ecology and Biogeography}, 21\penalty0 (4):\penalty0
  498--507, 2012.

\bibitem[Jitkrittum et~al.(2017)Jitkrittum, Xu, Szab{\'o}, Fukumizu, and
  Gretton]{jitkrittum2017linear}
Jitkrittum, W., Xu, W., Szab{\'o}, Z., Fukumizu, K., and Gretton, A.
\newblock A linear-time kernel goodness-of-fit test.
\newblock \emph{Advances in Neural Information Processing Systems}, 30, 2017.

\bibitem[Krizhevsky(2009)]{krizhevsky2009learning}
Krizhevsky, A.
\newblock Learning multiple layers of features from tiny images.
\newblock \emph{(Technical Report) University of Toronto.}, 2009.

\bibitem[Kurakin et~al.(2018)Kurakin, Goodfellow, and
  Bengio]{kurakin2018adversarial}
Kurakin, A., Goodfellow, I.~J., and Bengio, S.
\newblock Adversarial examples in the physical world.
\newblock In \emph{Artificial Intelligence Safety and Security}, pp.\  99--112.
  Chapman and Hall/CRC, 2018.

\bibitem[Laidlaw et~al.(2021)Laidlaw, Singla, and Feizi]{laidlaw2020perceptual}
Laidlaw, C., Singla, S., and Feizi, S.
\newblock Perceptual adversarial robustness: Defense against unseen threat
  models.
\newblock In \emph{International Conference on Learning Representations}, 2021.

\bibitem[Lee et~al.(2018)Lee, Lee, Lee, and Shin]{lee2018simple}
Lee, K., Lee, K., Lee, H., and Shin, J.
\newblock A simple unified framework for detecting out-of-distribution samples
  and adversarial attacks.
\newblock \emph{Advances in Neural Information Processing Systems}, 31, 2018.

\bibitem[Li \& Vorobeychik(2014)Li and Vorobeychik]{li2014feature}
Li, B. and Vorobeychik, Y.
\newblock Feature cross-substitution in adversarial classification.
\newblock \emph{Advances in Neural Information Processing Systems}, 27, 2014.

\bibitem[Li et~al.(2023)Li, Zhang, Cao, and Tan]{li2023learning}
Li, J., Zhang, S., Cao, J., and Tan, M.
\newblock Learning defense transformations for counterattacking adversarial
  examples.
\newblock \emph{Neural Networks}, 2023.

\bibitem[Liu et~al.(2020)Liu, Xu, Lu, Zhang, Gretton, and
  Sutherland]{liu2020learning}
Liu, F., Xu, W., Lu, J., Zhang, G., Gretton, A., and Sutherland, D.~J.
\newblock Learning deep kernels for non-parametric two-sample tests.
\newblock In \emph{International Conference on Machine Learning}, pp.\
  6316--6326. PMLR, 2020.

\bibitem[Long et~al.(2015)Long, Cao, Wang, and Jordan]{long2015learning}
Long, M., Cao, Y., Wang, J., and Jordan, M.
\newblock Learning transferable features with deep adaptation networks.
\newblock In \emph{International Conference on Machine Learning}, pp.\
  97--105. PMLR, 2015.

\bibitem[Lu et~al.(2022)Lu, Zhou, Bao, Chen, Li, and Zhu]{lu2022dpm}
Lu, C., Zhou, Y., Bao, F., Chen, J., Li, C., and Zhu, J.
\newblock Dpm-solver: A fast ode solver for diffusion probabilistic model
  sampling in around 10 steps.
\newblock 2022.

\bibitem[Ma et~al.(2018)Ma, Li, Wang, Erfani, Wijewickrema, Schoenebeck, Song,
  Houle, and Bailey]{ma2018characterizing}
Ma, X., Li, B., Wang, Y., Erfani, S.~M., Wijewickrema, S., Schoenebeck, G.,
  Song, D., Houle, M.~E., and Bailey, J.
\newblock Characterizing adversarial subspaces using local intrinsic
  dimensionality.
\newblock In \emph{International Conference on Learning Representations}, 2018.

\bibitem[Madry et~al.(2018)Madry, Makelov, Schmidt, Tsipras, and
  Vladu]{madry2017towards}
Madry, A., Makelov, A., Schmidt, L., Tsipras, D., and Vladu, A.
\newblock Towards deep learning models resistant to adversarial attacks.
\newblock In \emph{International Conference on Learning Representations}, 2018.

\bibitem[Nie et~al.(2022)Nie, Guo, Huang, Xiao, Vahdat, and
  Anandkumar]{nie2022diffusion}
Nie, W., Guo, B., Huang, Y., Xiao, C., Vahdat, A., and Anandkumar, A.
\newblock Diffusion models for adversarial purification.
\newblock In \emph{International Conference on Machine Learning}, 2022.

\bibitem[Niu et~al.(2022)Niu, Wu, Zhang, Chen, Zheng, Zhao, and
  Tan]{niu2022efficient}
Niu, S., Wu, J., Zhang, Y., Chen, Y., Zheng, S., Zhao, P., and Tan, M.
\newblock Efficient test-time model adaptation without forgetting.
\newblock In \emph{International conference on machine learning}, pp.\
  16888--16905. PMLR, 2022.

\bibitem[Niu et~al.(2023)Niu, Wu, Zhang, Wen, Chen, Zhao, and
  Tan]{niu2023towards}
Niu, S., Wu, J., Zhang, Y., Wen, Z., Chen, Y., Zhao, P., and Tan, M.
\newblock Towards stable test-time adaptation in dynamic wild world.
\newblock In \emph{The Eleventh International Conference on Learning
  Representations}, 2023.

\bibitem[Ozbulak et~al.(2019)Ozbulak, Van~Messem, and
  De~Neve]{ozbulak2019impact}
Ozbulak, U., Van~Messem, A., and De~Neve, W.
\newblock Impact of adversarial examples on deep learning models for biomedical
  image segmentation.
\newblock In \emph{MICCAI}, 2019.

\bibitem[Pang et~al.(2022)Pang, Zhang, He, Dong, Su, Chen, Zhu, and
  Liu]{pang2022two}
Pang, T., Zhang, H., He, D., Dong, Y., Su, H., Chen, W., Zhu, J., and Liu,
  T.-Y.
\newblock Two coupled rejection metrics can tell adversarial examples apart.
\newblock In \emph{Proceedings of the IEEE/CVF Conference on Computer Vision
  and Pattern Recognition}, pp.\  15223--15233, 2022.

\bibitem[Radford et~al.(2016)Radford, Metz, and
  Chintala]{radford2015unsupervised}
Radford, A., Metz, L., and Chintala, S.
\newblock Unsupervised representation learning with deep convolutional
  generative adversarial networks.
\newblock In \emph{International Conference on Learning Representations}, 2016.

\bibitem[Raghuram et~al.(2021)Raghuram, Chandrasekaran, Jha, and
  Banerjee]{raghuram2021general}
Raghuram, J., Chandrasekaran, V., Jha, S., and Banerjee, S.
\newblock A general framework for detecting anomalous inputs to dnn
  classifiers.
\newblock In \emph{International Conference on Machine Learning}, pp.\
  8764--8775. PMLR, 2021.

\bibitem[Ramesh et~al.(2022)Ramesh, Dhariwal, Nichol, Chu, and
  Chen]{ramesh2022hierarchical}
Ramesh, A., Dhariwal, P., Nichol, A., Chu, C., and Chen, M.
\newblock Hierarchical text-conditional image generation with clip latents.
\newblock \emph{arXiv preprint arXiv:2204.06125}, 2022.

\bibitem[Rombach et~al.(2022)Rombach, Blattmann, Lorenz, Esser, and
  Ommer]{rombach2022high}
Rombach, R., Blattmann, A., Lorenz, D., Esser, P., and Ommer, B.
\newblock High-resolution image synthesis with latent diffusion models.
\newblock In \emph{Proceedings of the IEEE/CVF Conference on Computer Vision
  and Pattern Recognition}, pp.\  10684--10695, 2022.

\bibitem[Saharia et~al.(2022)Saharia, Chan, Saxena, Li, Whang, Denton,
  Ghasemipour, Ayan, Mahdavi, Lopes, et~al.]{saharia2022photorealistic}
Saharia, C., Chan, W., Saxena, S., Li, L., Whang, J., Denton, E., Ghasemipour,
  S. K.~S., Ayan, B.~K., Mahdavi, S.~S., Lopes, R.~G., et~al.
\newblock Photorealistic text-to-image diffusion models with deep language
  understanding.
\newblock \emph{arXiv preprint arXiv:2205.11487}, 2022.

\bibitem[Shi et~al.(2021)Shi, Holtz, and Mishne]{shi2021online}
Shi, C., Holtz, C., and Mishne, G.
\newblock Online adversarial purification based on self-supervision.
\newblock In \emph{International Conference on Learning Representations}, 2021.

\bibitem[Song \& Ermon(2019)Song and Ermon]{song2019generative}
Song, Y. and Ermon, S.
\newblock Generative modeling by estimating gradients of the data distribution.
\newblock \emph{Advances in Neural Information Processing Systems}, 32, 2019.

\bibitem[Song et~al.(2021)Song, Sohl-Dickstein, Kingma, Kumar, Ermon, and
  Poole]{song2020score}
Song, Y., Sohl-Dickstein, J., Kingma, D.~P., Kumar, A., Ermon, S., and Poole,
  B.
\newblock Score-based generative modeling through stochastic differential
  equations.
\newblock In \emph{International Conference on Learning Representations}, 2021.

\bibitem[Stutz et~al.(2020)Stutz, Hein, and Schiele]{stutz2020confidence}
Stutz, D., Hein, M., and Schiele, B.
\newblock Confidence-calibrated adversarial training: Generalizing to unseen
  attacks.
\newblock In \emph{International Conference on Machine Learning}, pp.\
  9155--9166. PMLR, 2020.

\bibitem[Szegedy et~al.(2014)Szegedy, Zaremba, Sutskever, Bruna, Erhan,
  Goodfellow, and Fergus]{szegedy2013intriguing}
Szegedy, C., Zaremba, W., Sutskever, I., Bruna, J., Erhan, D., Goodfellow, I.,
  and Fergus, R.
\newblock Intriguing properties of neural networks.
\newblock In \emph{International Conference on Learning Representations}, 2014.

\bibitem[Vincent(2011)]{vincent2011connection}
Vincent, P.
\newblock A connection between score matching and denoising autoencoders.
\newblock \emph{Neural Computation}, 23\penalty0 (7):\penalty0 1661--1674,
  2011.

\bibitem[Wang \& He(2021)Wang and He]{wang2021enhancing}
Wang, X. and He, K.
\newblock Enhancing the transferability of adversarial attacks through variance
  tuning.
\newblock In \emph{Proceedings of the IEEE/CVF Conference on Computer Vision
  and Pattern Recognition}, pp.\  1924--1933, 2021.

\bibitem[Wong et~al.(2020)Wong, Rice, and Kolter]{wong2020fast}
Wong, E., Rice, L., and Kolter, J.~Z.
\newblock Fast is better than free: Revisiting adversarial training.
\newblock In \emph{International Conference on Learning Representations}, 2020.

\bibitem[Xie et~al.(2019)Xie, Zhang, Zhou, Bai, Wang, Ren, and
  Yuille]{xie2019improving}
Xie, C., Zhang, Z., Zhou, Y., Bai, S., Wang, J., Ren, Z., and Yuille, A.~L.
\newblock Improving transferability of adversarial examples with input
  diversity.
\newblock In \emph{Proceedings of the IEEE/CVF Conference on Computer Vision
  and Pattern Recognition}, pp.\  2730--2739, 2019.

\bibitem[Yoon et~al.(2021)Yoon, Hwang, and Lee]{yoon2021adversarial}
Yoon, J., Hwang, S.~J., and Lee, J.
\newblock Adversarial purification with score-based generative models.
\newblock In \emph{International Conference on Machine Learning}, pp.\
  12062--12072. PMLR, 2021.

\bibitem[Zagoruyko \& Komodakis(2016)Zagoruyko and
  Komodakis]{zagoruyko2016wide}
Zagoruyko, S. and Komodakis, N.
\newblock Wide residual networks.
\newblock In \emph{British Machine Vision Conference 2016}, 2016.

\bibitem[Zhang et~al.(2022)Zhang, Wu, Huang, Huang, Wang, Su, and
  Lyu]{zhang2022improving}
Zhang, J., Wu, W., Huang, J.-t., Huang, Y., Wang, W., Su, Y., and Lyu, M.~R.
\newblock Improving adversarial transferability via neuron attribution-based
  attacks.
\newblock In \emph{Proceedings of the IEEE/CVF Conference on Computer Vision
  and Pattern Recognition}, pp.\  14993--15002, 2022.

\bibitem[Zhu et~al.(2019)Zhu, Zhuang, and Wang]{zhu2019aligning}
Zhu, Y., Zhuang, F., and Wang, D.
\newblock Aligning domain-specific distribution and classifier for cross-domain
  classification from multiple sources.
\newblock In \emph{Proceedings of the AAAI Conference on Artificial
  Intelligence}, volume~33, pp.\  5989--5996, 2019.

\end{thebibliography}
\bibliographystyle{icml2023}

\onecolumn
\vskip 0.3in
\icmltitle{Supplementary Materials for \\``\mytitle''}
\yyf{In Supplementary, we provide detailed proofs of the theorem and corollary, descriptions of related works, more details and more experimental results of the proposed \mymethod. We organize the supplementary into the following sections.}
In Section \ref{sec:relatedwork}, we provide descriptions of related works regarding adversarial detection.
In Section \ref{sec:proof}, we derive the proofs of the theorem and corollary. 
In Section \ref{sec;pesudocode}, we demonstrate the pseudo-code of our proposed \mymethod.
In Section \ref{sec:implementation}, we present detailed implementation of our experiments.
In Section \ref{sec: Impact of EPS}, we  show the impact of EPS in our method.
In Section \ref{sec: Impact of Adding Perturbations}, we  study the impact of adding perturbations over samples in our method.
In Section \ref{sec:experimental_results}, we report more comparison results, more ablation studies of our proposed \mymethod. 

\setcounter{section}{0}
\renewcommand\thesection{\Alph{section}}

\section{Related Work}
\label{sec:relatedwork}
\textbf{Diffusion models.}
Diffusion models have emerged as a powerful generative model in many synthesis tasks \citep{song2019generative, dhariwal2021diffusion,saharia2022photorealistic,rombach2022high,ramesh2022hierarchical}. Since then, many {researchers} exploit the diffusion model to adversarial purification for improving the robustness of model \citep{nie2022diffusion, yoon2021adversarial}, where the score becomes a powerful means. Yet, only few researchers apply them to adversarial detection. Recently, \citet{yoon2021adversarial} employ the score model to purify the adversarial examples and use the norm of scores as the stopping condition of the purification. They also demonstrate some results about the norm of score in distinguishing adversarial samples from natural samples. However, this criterion is not effective enough for adversarial detection. In our work, we comprehensively consider multiple scores of perturbed samples, where these perturbed samples are from the same sample, and exploit rich information from these scores for adversarial detection. We empirically compare our approach to these methods and find that our approach outperforms these methods by a large margin.   

\textbf{Adversarial attack.}
Numerous studies have been proposed to attack neural networks by slightly modifying the input data to trigger misclassifications of classifiers. We enumerate a series of such works in what {follows}. Fast Gradient Sign Method (FGSM) \citep{goodfellow2014explaining} simply adds small noise along the gradient of the loss function. To further {adjust} the direction of increment, \citet{kurakin2018adversarial} propose Basic Iterative Method (BIM) that extends FGSM to iteratively take multiple steps. \citet{madry2017towards} propose Projected Gradient Descent (PGD) by combining the iterative method with random initialization for the adversarial example. Meanwhile, \citet{dong2018boosting} propose Momentum Iterative Method (MIM) by adding a momentum term to BIM for achieving a more stable attack. 
Besides, \citet{dong2019evading} propose Translation-Invariant Method (TIM) by optimizing the perturbation over translated images to obtain more transferable attacks, which can be incorporated into the methods FGSM and BIM. 
 {To further improve the transferability of attack, \citet{wang2021enhancing} propose variance tunning to guide the gradient update, namely VMI-FGSM; \citet{zhang2022improving} conduct feature-level attacks with more accurate neuron importance estimations, called Neuron Attribution-based (NAA) attack.}
Besides the non-targeted attack methods mentioned above, there are also many methods {that} perturb data to one target label. For example, \citet{carlini2017towards} perform {an} attack by incorporating the iterative mechanism of BIM, called CW. Besides, there are {approaches} that combine multiple attacks, such as the AutoAttack \citep{croce2020reliable} and the Minimum-margin (MM) attack \citep{gao2022fast}, the faster version of AutoAttack. {The various} attacks may cause serious consequences in security-critical tasks, raising an urgent requirement for advanced techniques to achieve a robust model.

\textbf{Adversarial detection.}
To ensure the safety of machine learning system, a plethora of exploration for adversarial detection has attracted increasing sight. The most common idea is to filter out adversarial samples from test data using a trained binary classifier. Recently, statistics on hidden-layer features of DNNs are widely considered for adversarial detection. \citet{feinman2017detecting} train a logistic regression classifier using Kernel Density (KD) of features in the last hidden layer, as well as Bayesian Uncertainty (BU) as a basis. \citet{ma2018characterizing} consider the local intrinsic dimensionality (LID) of the features of DNNs as the characteristics for detection. \citet{lee2018simple} use a mahalanobis distance-based score to detect adversarial examples.  {\citet{raghuram2021general} extract the intermediate layer representations of DNNs in the form of a meta-algorithm with configurable components for detection.} In addition, \citet{deng2021libre} {train} a Bayesian neural network by adding uniform noises to the samples.  {Another mainstream strategy for handling adversarial examples is to equip the classifier with a rejection option. \citet{stutz2020confidence} propose a confidence-calibrated adversarial training framework by inducing the model towards low confidence predictions on adversarial examples to decide which sample to reject. \citet{pang2022two} suggest  coupling confidence with a proposed R-Con as metrics to separate adversarial samples from normal ones.} However, these methods train a tailored detector for some specific attacks or for a specific classifier, which largely overlooks the modeling for data distribution, leading to their limited performance under unknown attacks.

\section{Proofs in Section \ref{sec: method}}
\label{sec:proof}
\subsection{Proof of Theorem \ref{thm: adv vs natural}}

\textbf{Theorem \ref{thm: adv vs natural}}
\emph{
    Assuming that the distribution of natural data $p(\bx){=} \mN({\boldsymbol{\mu}}_{\bx}, \sigma_{\bx}^2\mathbf{I})$,  {where $\mathbf{I}$ is an identity matrix}, given a perturbation transition kernel $p_{0 t}\left(\mathbf{x}_{t} \mid \mathbf{x}_{0}\right)=\mathcal{N}\left( \gamma_{t} \mathbf{x}_{0}, \sigma_{t}^{2} \mathbf{I}\right)$
    with $\gamma_{t}$ and $\sigma_{t}$ being the time-dependent noise schedule, then  the following three conclusions hold: \\
     1) For $\forall{~}\bx \sim p(\bx)$, $S(\bx) \sim \mN({\bf{0}}, \sigma_S^2{\bI})$;\\
    2) For $\forall{~}\by \sim p(\bx)$ and adversarial sample $\hat{\by} {=} \by {+} \boldsymbol{\varepsilon}$, $S(\hat{\by}) \sim \mN(-{\boldsymbol{\mu}}_{S}, \sigma_S^2{\bI})$;\\
    3) For $\forall~\bx, \by \sim p(\bx)$ and adversarial sample $\hat{\by} {=} \by {+} \boldsymbol{\varepsilon}$, 
    \begin{equation}
    \label{appendix_natural diff}
        S(\bx) - S(\by) \sim \mN({\bf{0}}, 2\sigma_S^2{\bI});
    \end{equation}
    \begin{equation}
    \label{appendix_adver diff}
        S(\bx) - S(\hat{\by}) \sim \mN({\boldsymbol{\mu}}_{S}, 2\sigma_S^2{\bI}),
    \end{equation}
    where ${\boldsymbol{\mu}}_{S}{=}\mmE_t \frac{ \boldsymbol{\varepsilon}}{\gamma_t^2\sigma_{\bx}^2{+}\sigma_t^2}$ and $\sigma_{S}^2=\mmE_t \frac{1}{\gamma_t^2\sigma_{\bx}^2+\sigma_t^2}$.
    }
\begin{proof}
1) Based on the distribution $p(\bx)$, \ie, $\bx_0 = \bx \sim \mN({\boldsymbol{\mu}}_{\bx}, \sigma_{\bx}^2\mathbf{I})$, we obtain $\bx_0 = {\boldsymbol{\mu}}_{\bx} + \sigma_{\bx} \bz $ with $\bz \sim \mN({\bf{0}},\bI)$; based on the perturbation transition kernel $p_{0 t}\left(\mathbf{x}_{t} \mid \mathbf{x}_{0}\right)=\mathcal{N}\left(\gamma_{t} \mathbf{x}_{0}, \sigma_{t}^{2} \mathbf{I}\right)$, we have $\bx_t = \gamma_t \bx_0 +  \sigma_t \bz$. Combining the distribution of $\bx$ and $\bx_0$, we have
\begin{equation}
    \bx_t = \gamma_t \boldsymbol{\mu}_{\bx} + \sqrt{\gamma_t^2 \sigma_{\bx}^2+\sigma_t^2} \bz, ~~~~\ie, \bx_t \sim \mN(\gamma_t \boldsymbol{\mu}_{\bx},\gamma_t^2 \sigma_{\bx}^2+\sigma_t^2)
\end{equation}
For $\bx_t \sim p_t(\bx) = \mN(\gamma_t \boldsymbol{\mu}_{\bx},\gamma_t^2 \sigma_{\bx}^2+\sigma_t^2)$, we calculate the derivation 
\begin{equation}
\label{eq: derivation}
    \nabla_{\mathbf{x}} \log p_t\left(\mathbf{x}\right) = -\frac{\bx_t - \gamma_t \boldsymbol{\mu}_{\bx}}{\gamma_t^2 \sigma_{\bx}^2+\sigma_t^2} = -\frac{1}{\sqrt{\gamma_t^2 \sigma_{\bx}^2+\sigma_t^2}} \mN({\bf{0}},\bI).
\end{equation}
Taking expectation to $t$, we give the distribution of $S(\bx)$
\begin{equation}
\label{derivation: nat}
    S(\bx) \sim \mN({\bf{0}}, \sigma_S^2{\bI}),~~\text{where}~ \sigma_{S}^2=\mmE_t \frac{1}{\gamma_t^2\sigma_{\bx}^2+\sigma_t^2}.
\end{equation}

2) Based on $\by \sim p(\bx)$ and $\hat{\by} = \by +\boldsymbol{\varepsilon}$, we obtain $\hat{\by}_0 = \hat{\by} \sim \mN({\boldsymbol{\mu}}_{\bx}+\boldsymbol{\varepsilon}, \sigma_{\bx}^2\mathbf{I}) $. Then, we have 
\begin{equation}
    \nabla_{\hat{\by}} \log p_t\left(\hat{\by}\right) = -\frac{{\by}_t + \boldsymbol{\varepsilon} - \gamma_t \boldsymbol{\mu}_{\bx}}{\gamma_t^2 \sigma_{\bx}^2+\sigma_t^2} = -\frac{1}{\sqrt{\gamma_t^2 \sigma_{\bx}^2+\sigma_t^2}} \mN({\bf{0}},\bI) - \frac{ \boldsymbol{\varepsilon}}{\gamma_t^2\sigma_{\bx}^2{+}\sigma_t^2},
\end{equation}
where the last equation is based on $p_t({\by}) = p_t(\bx) = \mN(\gamma_t \boldsymbol{\mu}_{\bx},\gamma_t^2 \sigma_{\bx}^2+\sigma_t^2)$.

Taking expectation to $t$, we give the distribution of $S(\hat{\by})$
\begin{equation}
\label{derivation: adv}
    S(\hat{\by}) \sim \mN({-\boldsymbol{\mu}}_{S}, \sigma_S^2{\bI}),~~\text{where}~{\boldsymbol{\mu}}_{S}{=}\mmE_t \frac{ \boldsymbol{\varepsilon}}{\gamma_t^2\sigma_{\bx}^2{+}\sigma_t^2} ~~\text{and}~ \sigma_{S}^2=\mmE_t \frac{1}{\gamma_t^2\sigma_{\bx}^2+\sigma_t^2}.
\end{equation}

3) According to the additive property of the Gaussian distribution, combing Eq. (\ref{derivation: nat}) and Eq. (\ref{derivation: adv}), we obtain the third conclusion.

\end{proof}

\newpage
\subsection{Proof of Corollary \ref{probability of K}}
\textbf{Corollary \ref{probability of K}}
\emph{
    Considering the Gaussian kernel $k \left(\ba, \bb\right){=}\exp \left(-\left\|\ba-\bb\right\|^{2} /\left(2 \sigma^{2}\right)\right)$ and the assumption in Theorem \ref{thm: adv vs natural}, for $\forall 0 {<} \eta {<}1$, the probability of $P\{k\left(S(\bx), S(\hat{\by})\right){>}\eta\}$ is given by 
    \begin{equation}
        P\{k\left(S(\bx), S(\hat{\by})\right){>}\eta\} = \int_0^C \chi^2_d(z)dz,
    \end{equation}
    where $z = \|{\boldsymbol{\mu}_S}\|^2$ with ${\boldsymbol{\mu}_S} $ being the mean of $S(\bx)-S(\hat{\by})$, C is a constant for given $\eta$ and $\sigma$, $\chi^2$ is the probability density function of noncentral chi-squared distribution with $d$ degrees of freedom \citep{abdel1954approximate}.
}
\begin{proof}
Based on the Gaussian kernel $k \left(\ba, \bb\right)$, we have
\begin{align}
    P\left\{\kappa(S(\bx), S(\hat{\by})){>}\eta\right\} = P\left\{\exp \left(-\left\|S(\bx)-S(\hat{\by})\right\|^{2} /\left(2 \sigma^{2}\right)\right){>}\eta\right\}
    =  P\left\{\left\|S(\bx)-S(\hat{\by})\right\|^{2}<-2 \sigma^2 \ln \eta \right\}.
\end{align}
Let $\boldsymbol{\xi} \sim S(\bx)-S(\hat{\by})$, then  $\boldsymbol{\xi_i} \sim \mN((\boldsymbol{\mu}_{S})_i, 2\sigma_{S}^2)$, thus we have 
\begin{equation}
     P\left\{\kappa(S(\bx), S(\hat{\by})){>}\eta\right\}  = P\left\{ \sum_{i=1}^d \boldsymbol{\xi_i}^2< -2 \sigma^2 \ln \eta \right\} = P\left\{ \sum_{i=1}^d (\frac{\boldsymbol{\xi_i}}{\sqrt{2}\sigma_S})^2< \frac{- \sigma^2 \ln \eta}{  \sigma_S^2}\right\}
\end{equation}
Note that $\frac{\boldsymbol{\xi_i}}{\sqrt{2}\sigma_S} \sim \mN((\boldsymbol{\mu}_{S})_i,1)$,
based on the definition of noncentral chi-squared distribution \citep{abdel1954approximate}, we have 
\begin{equation}
    \sum_{i=1}^d (\frac{\boldsymbol{\xi_i}}{\sqrt{2}\sigma_S})^2 \sim \chi^2_d(\|{\boldsymbol{\mu}_S}\|^2)
\end{equation}
Let $C = \frac{-\sigma^2 \ln \eta}{  \sigma_S^2}$ and $z = \|{\boldsymbol{\mu}_S}\|^2$, we obtain the conclusion
\begin{equation}
    P\left\{k(\bx, \hat{\by}){>}\eta\right\} = \int_0^C \chi^2_d(z)dz.
 \end{equation}

\end{proof}

\section{Pseudo-code of \mymethod}
\label{sec;pesudocode}
\begin{algorithm}[h]
\caption{ Expected Perturbation Score for Adversarial Detection (\mymethod).}
\begin{algorithmic}[1]
    \INPUT  A natural sample set $\{\bx_{0}^{(i)}\}_{i=1}^n$, a test sample $\tilde{\bx}_{0}$, a pre-trained score model $s_{\boldsymbol{\theta}}\left(\mathbf{x}_{t}, t\right)$ and a diffusion timestep $T^{*}$.
    \OUTPUT 
    MMD between EPS of the test sample $\tilde{\bx}_{0}$ and EPSs of natural samples $\{\bx_{0}^{(i)}\}$.

    \STATE Set initial time step $t=1$~;
    \FOR{$ t =1, \ldots, T^{*}$}
    \STATE Obtain perturbed natural samples $\{\bx_{t}^{(i)}\}_{i=1}^n$ according to  $p_{0 t}\left(\mathbf{x}_{t} \mid \mathbf{x}_{0}\right)$.
    
    \STATE Obtain perturbed test sample $\tilde{\bx}_{t}^{(i)}$ according to  $p_{0 t}\left(\hat{\mathbf{x}}_{t} \mid \tilde{\mathbf{x}}_{0}\right)$. 
    \ENDFOR
    \STATE Compute EPSs of natural samples $\{S(\bx^{(i)})\}_{i=1}^n$ using Eq. (\ref{Eq: expected perturbation score}).

    \STATE Compute EPS of the test sample $S(\tilde{\bx})$ using Eq. (\ref{Eq: expected perturbation score}).
           
    \STATE Compute the deep kernel MMD between $\{S(\bx^{(i)})\}_{i=1}^n$ and $S(\tilde{\bx})$ using Eq. (\ref{MMD}).
\end{algorithmic}
\end{algorithm}

\newpage

\section{More Details for Experiment Settings}
\label{sec:implementation}
\subsection{Implementation Details of Our Method}
\label{sec:implementation of deep-kernel-MMD}
Our adversarial detection method is built upon diffusion models. Specifically, we consider the
pre-trained diffusion model of Score SDE for CIFAR-10 
following \citet{song2020score} and choose the \textit{vp/cifar10\_ddpmpp\_deep\_continuous} checkpoint from the \textit{score\_sde} library 
\footnote{\scriptsize{\url{https://github.com/yang-song/score_sde}}};
for ImageNet, we consider the
pre-trained diffusion model of Guided Diffusion following \citet{dhariwal2021diffusion} and use the \textit{$256 \times 256$ diffusion (unconditional)} checkpoint from the \textit{guided-diffusion} library \footnote{\scriptsize\url{https:// github.com/openai/guided-diffusion}}.

For classifiers, we use pre-trained WiderResNet-28-10 \citep{zagoruyko2016wide} and  WiderResNet-70-16 \citep{gowal2021improving} for CIFAR-10, and ResNet-50, ResNet-101 \citep{he2016deep}, and DeiT-S \citep{dosovitskiy2021image} for ImageNet. For attacks, we consider $8$ attack intensities $\epsilon {\in}\{1/255,{\ldots},8/255\}$ with  iterative attacks run for $5$ steps using step size $\epsilon/5$ under $12$ different $\ell_2$ and $\ell_\infty$ attack methods to generate adversarial examples, including PGD, PGD-$\ell_2$ \citep{madry2017towards}, FGSM, FGSM-$\ell_2$ \citep{goodfellow2014explaining}, BIM, BIM-$\ell_2$ \citep{kurakin2018adversarial}, MIM \citep{dong2018boosting}, TIM \citep{dong2019evading}, CW \citep{carlini2017towards}, DI\_MIM\citep{xie2019improving} and two adaptive {attacks} AutoAttack (AA) \citep{croce2020reliable} and Minimum-Margin Attack (MM) \citep{gao2022fast} that is a faster version of AA. For evaluation, we choose the area under the receiver operating {characteristic} curve (AUROC) as metric for adversarial detection.  

 {In our proposed \mymethod, we employ the deep kernel MMD algorithm as described by \citet{liu2020learning}. The deep kernel is defined as $k_{\omega}(S(\mathbf{x}), S(\mathbf{y}))=\left[(1-\epsilon_0) \kappa\left(\phi_{\omega}(S(\mathbf{x})), \phi_{\omega}(S(\mathbf{y}))\right)+\epsilon_0\right] q(S(\mathbf{x}), S(\mathbf{y}))$, where  $\epsilon_0 \in (0,1)$, $\phi_{\omega}$ is a deep neural network, $\kappa$ is a Gaussian kernel with bandwidth $\sigma_{\phi_{\omega}}$ and $q(S(\mathbf{x}), S(\mathbf{y}))$ is a Gaussian kernel with bandwidth $\sigma_q$. 
We use the deep network $\phi_{\omega}$ based on the discriminator architecture in DCGAN \cite{radford2015unsupervised}.
Through all our experiments, we use only FGSM and FGSM-$\ell_{2}$ adversarial samples ($\epsilon=1/255$), $10,000$ each, along with $10,000$ nature samples to calculate their EPSs and train the parameters $\{\epsilon_0, \phi_{\omega}, \sigma_{\phi_{\omega}}, \sigma_q\}$  to detect all the other types of attacks with $8$ varying attack intensities following Algorithm 1 in \citet{liu2020learning}\footnote{\scriptsize\url{https://github.com/fengliu90/DK-for-TST}}, in which we set the learning rate to $0.00002$ for CIFAR-10 and $0.002$ for ImageNet.}
The deep kernel can  also be trained on a general public dataset which we leave for our future work. Note that our method is suitable for detecting all the $\ell_2$ and $\ell_\infty$ adversarial samples.
We conduct our experiments based on Python 3.7 and Pytorch 1.7.1 on a server with 1× RTX 3090 GPU.

\subsection{Implementation Details of Baselines}
We choose three standard adversarial detection approaches, KD \citep{feinman2017detecting}, LID \citep{ma2018characterizing} and MD \citep{lee2018simple} as baselines for both CIFAR-10 and ImageNet, as well as  LiBRe \citep{deng2021libre} individually on the ImageNet, which trains a Bayesian neural network by adding the uniform noises into the samples. 

\textbf{KD \& LID \& MD.}
We implement KD following  the codebase\footnote{\scriptsize\url{https://github.com/rfeinman/detecting-adversarial-samples}}, LID following the codebase\footnote{\scriptsize\url{https://github.com/xingjunm/lid_adversarial_subspace_detection}}, and  MD following the codebase\footnote{\scriptsize\url{https://github.com/pokaxpoka/deep_Mahalanobis_detector}}. These three methods \yyf{train a logistic regressor} to distinguish natural, noisy and adversarial samples. To show their best performance for adversarial detection, we choose the noise scale of the $\ell_2$ distance in KD, LID and MD as 40, 1, 10 on CIFAR-10 and 20, 1, 10 on ImageNet. Besides, for KD, we choose bandwidth as 10 on CIFAR-10 and 20 on ImageNet.

\textbf{LiBRe.}
We implement LiBRe on ImageNet  following their codebase \footnote{\scriptsize\url{https://github.com/thudzj/ScalableBDL}}. 
We evaluate its performance on ResNet-50 and ResNet-101 to make a comparison.

\subsection{Implementation Details of Figure \ref{fig: motivation}}
For the results in Figure \ref{fig: motivation}, we calculate the norms of scores of natural samples and adversarial samples at different purification timesteps using a  score model pre-trained on ImageNet, where these adversarial samples are crafted by FGSM with $\epsilon=1/255$. Before feeding these samples into the score model, we do not perturb them again. To better demonstrate these results, we normalize score norm with the maximum of scores in each timestep.

\newpage
\section{Impact of EPS for \mymethod}
\label{sec: Impact of EPS}
In our method \mymethod, we use diffusion-based score model to calculate the characteristics of samples, \ie, EPS, which has the same dimension as this sample. In this experiment, we investigate the impact of EPS in our method. As a result, we remove the calculation of EPS from our method, instead using the raw sample as their characteristic. Table  \ref{appendix_impact_of_diffusion} shows  adversarial detection performance of our method against $6$ attacks under attack intensities $\epsilon \in \{2/255, 4/255\}$ on ImageNet  over ResNet-50 compared to that without EPS. Obviously, \mymethod~without employing EPS demonstrates  significant  performance drop ($\approx 28\% \downarrow$), suggesting the superiority of our proposed EPS in distinguishing between adversarial and natural samples.

\begin{table}[ht]
    \caption{Impact of EPS with ResNet-50 on ImageNet under $\epsilon=2/255$ and $\epsilon=4/255$.}
    \centering
    \small
    \begin{tabular*}{15.5cm}{@{}@{\extracolsep{\fill}}cc|cccccccc@{}}
    \toprule
         Perturbation & Method & FGSM & PGD & BIM & MIM & TIM & CW \\ \midrule
        \multirow{2}{*}{$\epsilon = 2/255$} & \mymethod~ (w/o EPS) & 0.7132 & 0.7108 & 0.7101 & 0.7119 & 0.7120 & 0.7107 \\
        & \mymethod~ (w/ EPS) & \textbf{1.0000} & \textbf{0.9997} & \textbf{0.9999} & \textbf{1.0000} & \textbf{0.9983} & \textbf{0.9995} \\ \midrule
        \multirow{2}{*}{$\epsilon = 4/255$} & \mymethod~ (w/o EPS) & 0.7176 & 0.7144 & 0.7131 & 0.7156 & 0.7156 & 0.7142 \\
        & \mymethod~ (w/ EPS) & \textbf{1.0000} & \textbf{1.0000} & \textbf{1.0000} & \textbf{1.0000} & \textbf{1.0000} & \textbf{1.0000} \\ \midrule
         Perturbation & Method & DI\_MIM & PGD-$\ell_{2}$ & FGSM-$\ell_{2}$ & BIM-$\ell_{2}$ & MM & AA  \\ \midrule
        \multirow{2}{*}{$\epsilon = 2/255$} & \mymethod~ (w/o EPS) & 0.7117 & 0.7118 & 0.7133 & 0.7097 & 0.7066 & 0.7067 \\
        \rule{0pt}{10pt}
        & \mymethod~ (w/ EPS) & \textbf{0.9999} & \textbf{1.0000} & \textbf{1.0000} & \textbf{0.9999} & \textbf{1.0000} & \textbf{1.0000} \\  \midrule
        \multirow{2}{*}{$\epsilon = 4/255$} & \mymethod~ (w/o EPS) & 0.7156 & 0.7154 & 0.7175 & 0.7128 & 0.7101 & 0.7104 \\
        & \mymethod~ (w/ EPS) & \textbf{1.0000} & \textbf{1.0000} & \textbf{1.0000} & \textbf{1.0000} & \textbf{1.0000} & \textbf{1.0000} \\  \bottomrule
    \end{tabular*}
    \label{appendix_impact_of_diffusion}
\end{table}

\section{Impact of Adding Perturbations over Samples for \mymethod }
\label{sec: Impact of Adding Perturbations}
Adding  perturbations into the samples is \yyf{critical} for our proposed \mymethod. \yyf{To investigate  the impact of this operation, we conduct ablation studies against 6 adversarial attack methods on ImageNet}. Table \ref{appendix_impact_of_perturb} demonstrates adversarial detection performance of \mymethod~against 6 attacks under $\epsilon=1/255$ on ImageNet over ResNet-50 compared to that without adding perturbations. We observe that  the adversarial detection performance is significantly improved with adding perturbations. Specifically, our method obtains about $1.86\% \uparrow$ on average of 12 attacks, in which the maximum is $4.84\% \uparrow$ against BIM-$\ell_{2}$. This coincides {with} the conclusion in Theorem \ref{thm: adv vs natural} that adding the perturbations helps distinguish between adversarial and natural samples.

\begin{table}[ht]
    \caption{Impact of adding perturbations with ResNet-50 on ImageNet under $\epsilon=1/255$.}
    \centering
    \small
    \begin{tabular*}{12.5cm}{@{}@{\extracolsep{\fill}}c|cccccccc@{}}
    \toprule
        Method & PGD & BIM & CW & FGSM-$\ell_{2}$ & BIM-$\ell_{2}$ & AA \\ \midrule
        \mymethod~(w/o perturbation) & 0.9274 & 0.9405 & 0.9257 & 0.9960 & 0.9443 & 0.9835 \\
        \mymethod~(w/ perturbation) & \textbf{0.9637} & \textbf{0.9845} & \textbf{0.9549} & \textbf{0.9997} & \textbf{0.9927} & \textbf{0.9936} \\ \midrule
        Method & FGSM & MIM & TIM & DI\_MIM & PGD-$\ell_{2}$ & MM \\ \midrule
        \mymethod~(w/o perturbation) & 0.9954 & 0.9885 & 0.9397 & 0.9791 & 0.9858 & 0.9818 \\
        \mymethod~(w/ perturbation) & \textbf{0.9982} & \textbf{0.9972} & \textbf{0.9561} & \textbf{0.9817} & \textbf{0.9961} & \textbf{0.9928} \\ \bottomrule
    \end{tabular*}
    \label{appendix_impact_of_perturb}
\end{table}

\newpage

\section{More Results of Adversarial Detection}
\label{sec:experimental_results}

\yyf{To further evaluate the effectiveness of our proposed \mymethod, in subsection \ref{subsec:comparison}, we conduct more comparison experiments on detecting more adversarial {attacks} on ImageNet and CIFAR-10 datasets. 
To demonstrate the generalization of \mymethod, we conduct more experiments on detecting unseen attacks and transferable attacks.
In subsection \ref{subsec:ablation}, we additionally study the mechanism of \mymethod~by demonstrating the impact of time steps, set size, low attack intensity,  transferability across datasets,  {robustness against adaptive attacks} and inference efficiency.} We provide extra results of all compared experiments in Tables \ref{appendix_additional AUROC:cifar and image}, \ref{appendix_attitional_unseen_results}, \ref{appendix_attitional_transform_results} and Figures \ref{appendix_fig:comparison_CIFAR}, \ref{appendix_fig:comparison_ImageNet},  \ref{appendix_fig:ablation_on_setsize}, as supplements of additional attack methods corresponding to the main body.

\subsection{More Comparison Experiments}
\label{subsec:comparison}
\begin{figure*}[th]
    \begin{center}
        \subfigure[Attack Method: BIM-L2]
        {\includegraphics[width=0.32\textwidth]{ 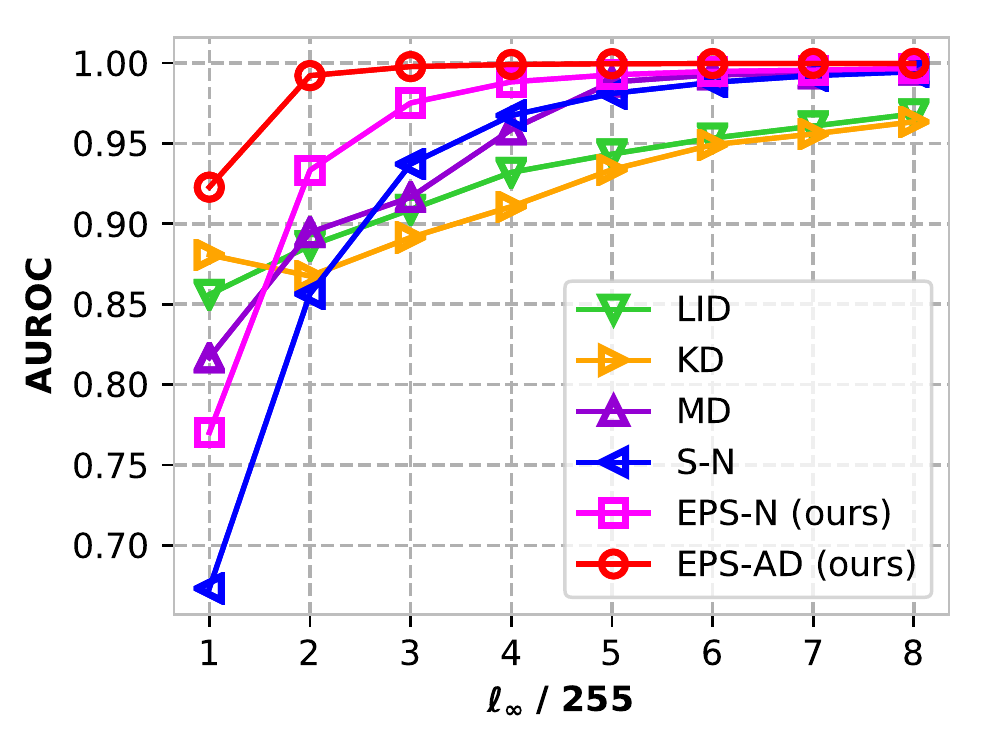}}
        \subfigure[Attack Method: DI\_MIM]
        {\includegraphics[width=0.32\textwidth]{ 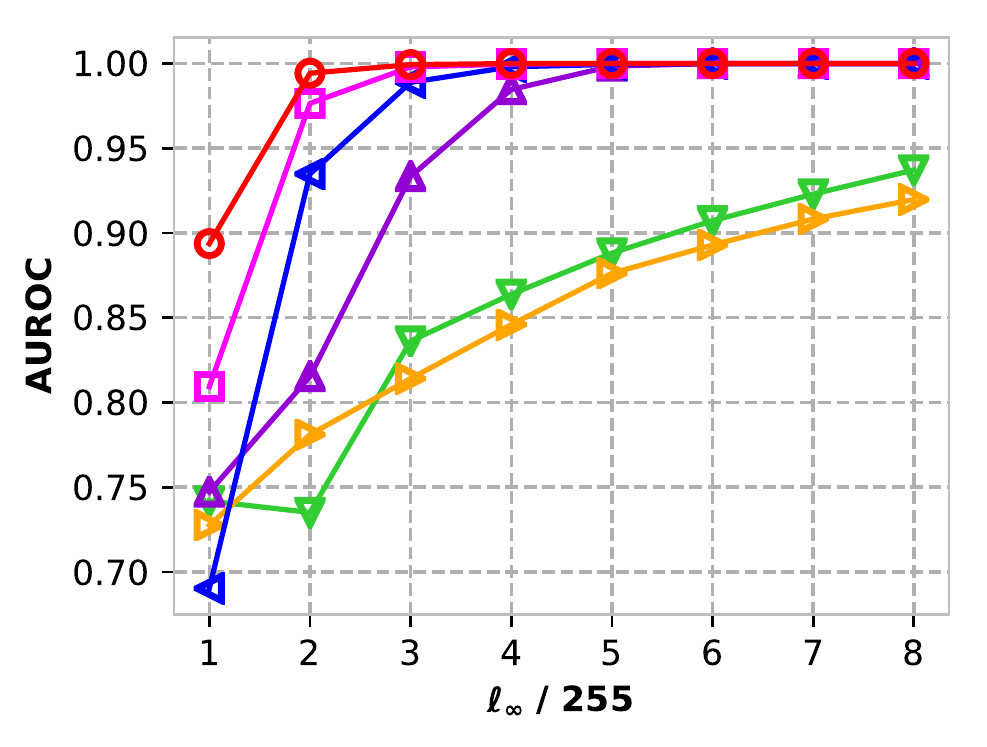}}
        \subfigure[Attack Method: MIM]
        {\includegraphics[width=0.32\textwidth]{ 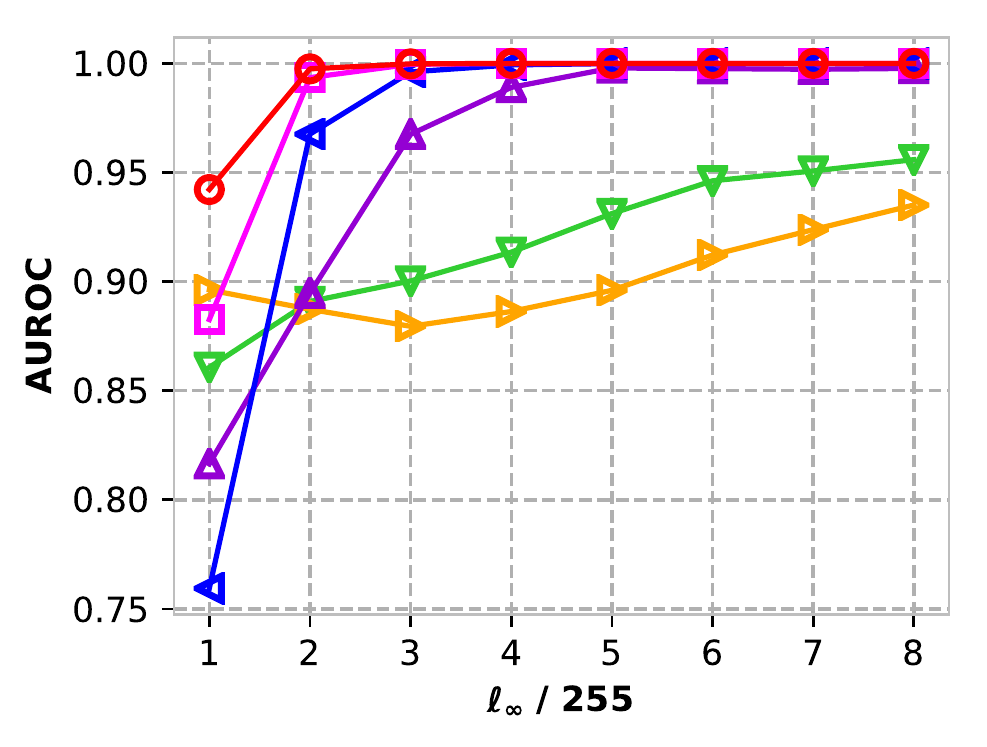}}
        \subfigure[Attack Method: MM Attack]
        {\includegraphics[width=0.32\textwidth]{ 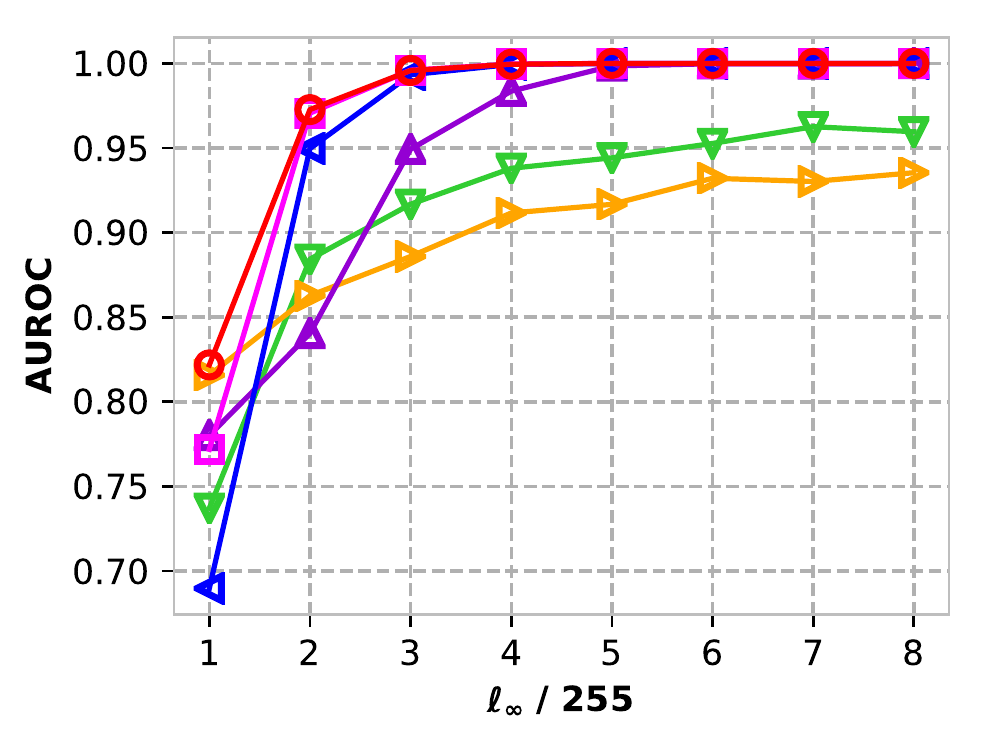}}
        \subfigure[Attack Method: PGD-$\ell_2$]
        {\includegraphics[width=0.32\textwidth]{ 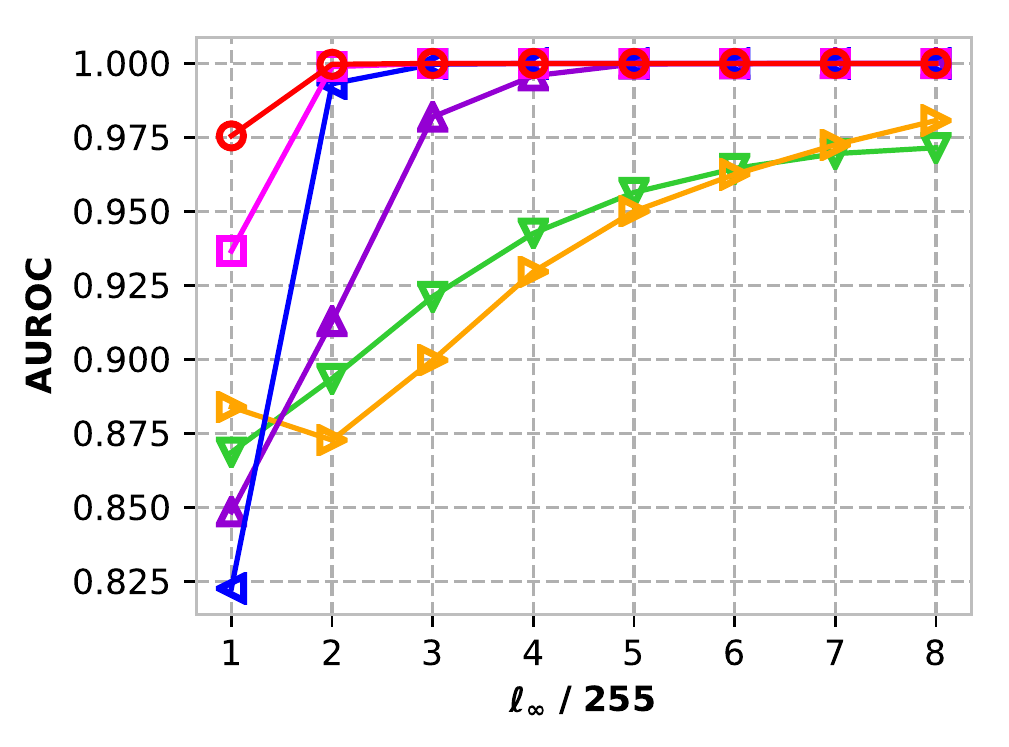}}
        \subfigure[Attack Method: TIM]
        {\includegraphics[width=0.32\textwidth]{ 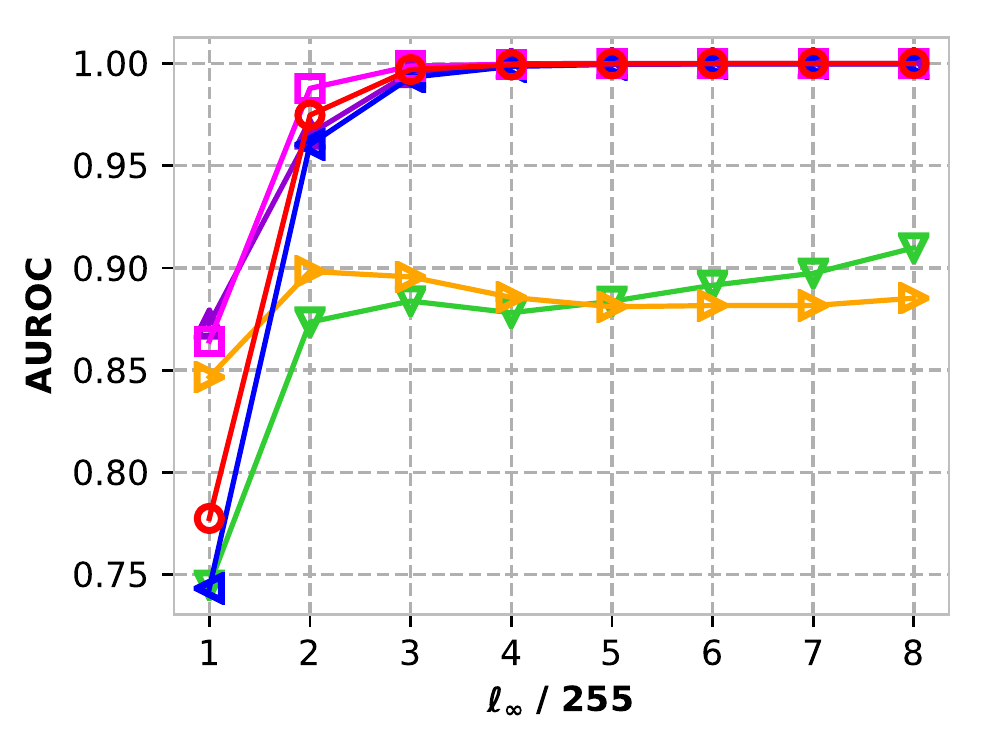}}

        \vspace{-1em}
        \caption{ More Results of adversarial detection on CIFAR-10. Sub-figures (a) - (f) report the AUROC on different attacks under  {$\epsilon {\in}\{1/255,{\ldots},8/255\}$} and share the same legend in sub-figure (a). 
      }
        \label{appendix_fig:comparison_CIFAR}
    \end{center}
    \vspace{-2em}
\end{figure*}

\begin{figure*}[th]
\label{fig: imagenet_APPENDIX}
    \begin{center}
        \subfigure[Attack Method: BIM-L2]
        {\includegraphics[width=0.32\textwidth]{ 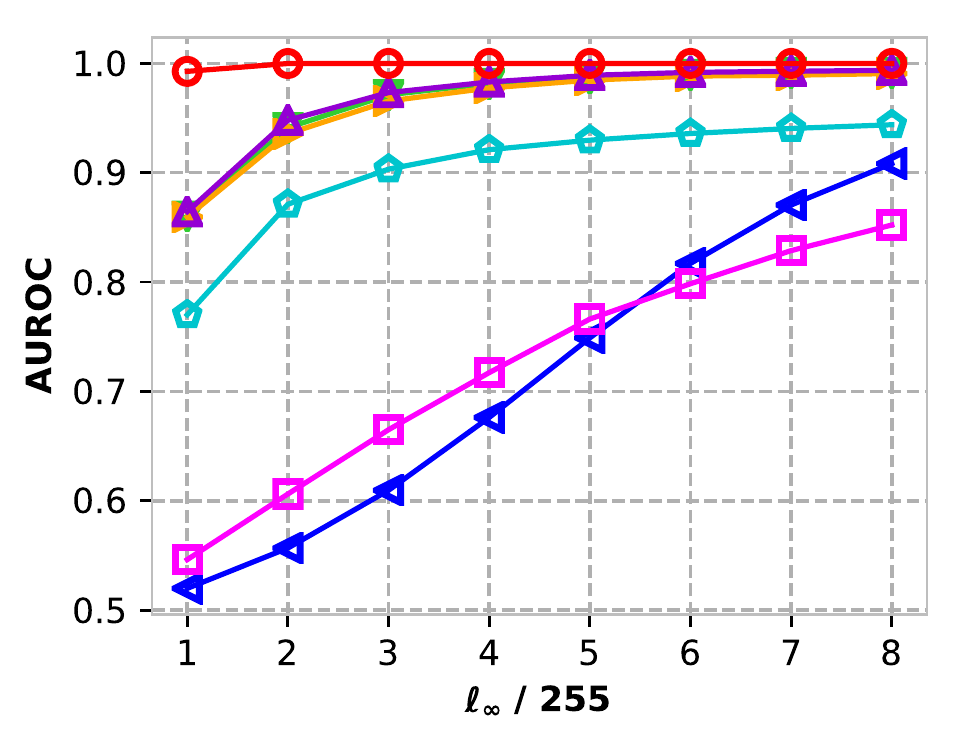}}
        \subfigure[Attack Method: DI\_MIM]
        {\includegraphics[width=0.32\textwidth]{ 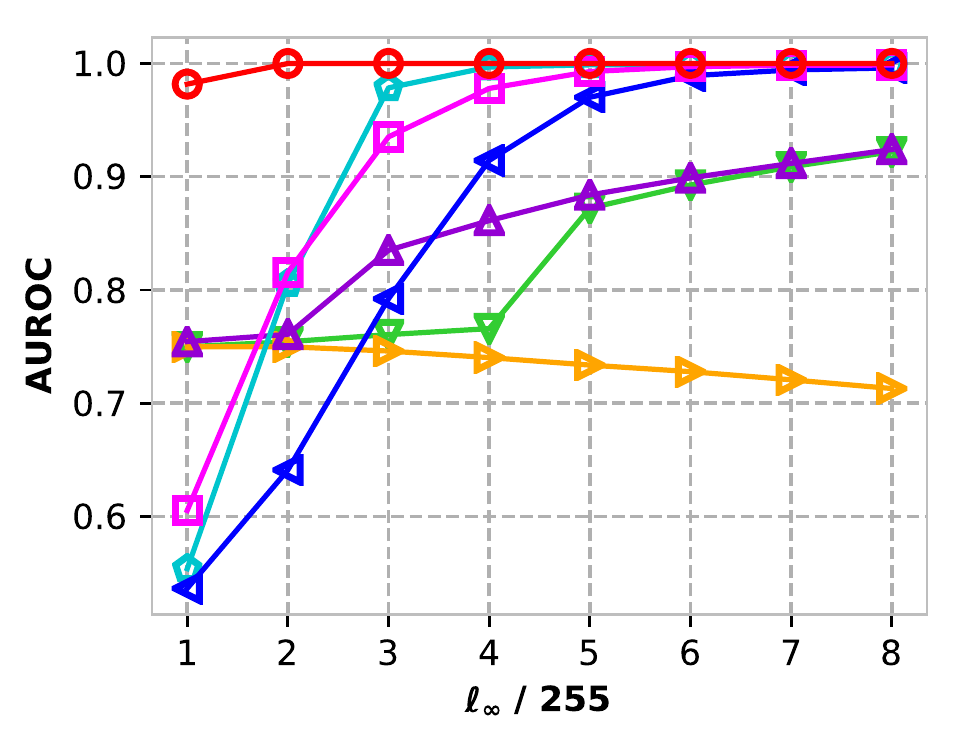}}
        \subfigure[Attack Method: MIM]
        {\includegraphics[width=0.32\textwidth]{ 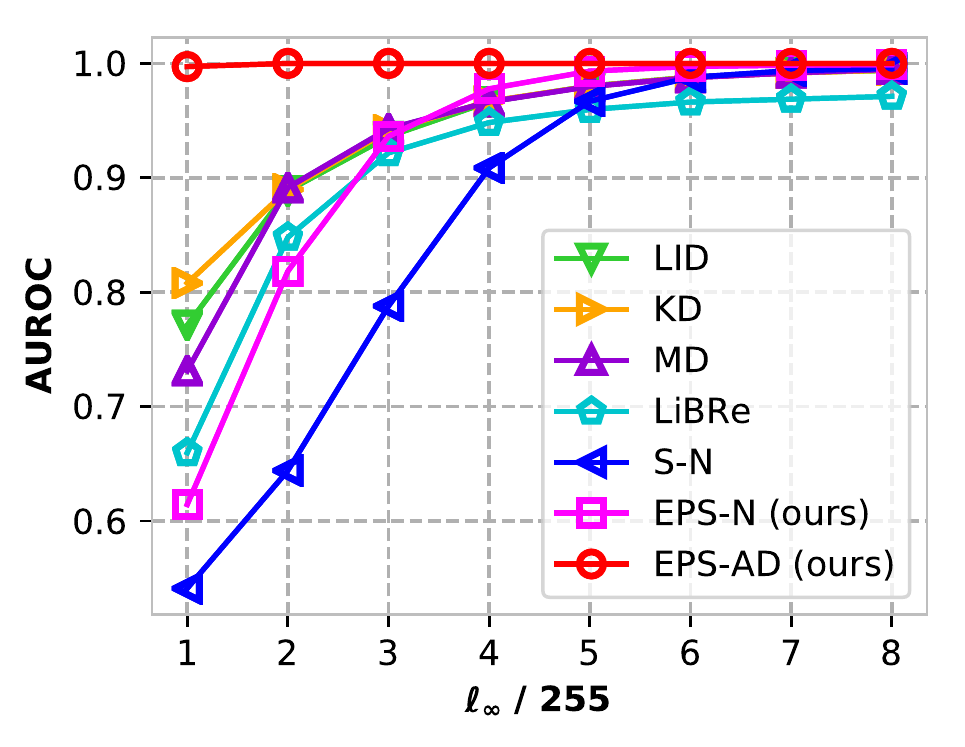}}
        \subfigure[Attack Method: MM Attack]
        {\includegraphics[width=0.32\textwidth]{ 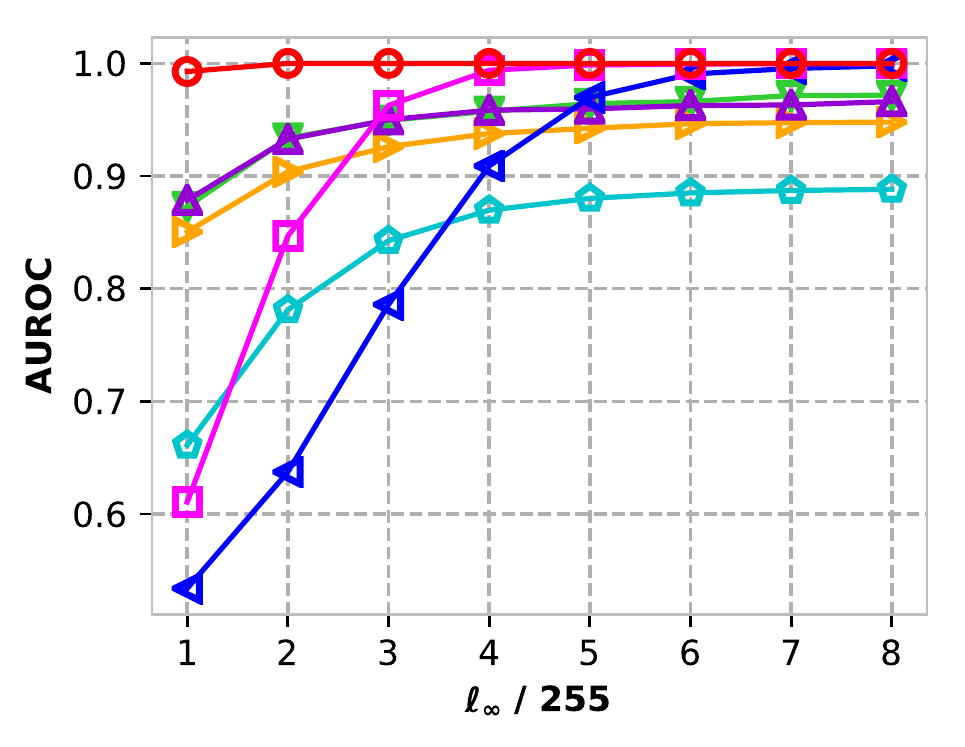}}
        \subfigure[Attack Method: PGD-$\ell_2$]
        {\includegraphics[width=0.32\textwidth]{ 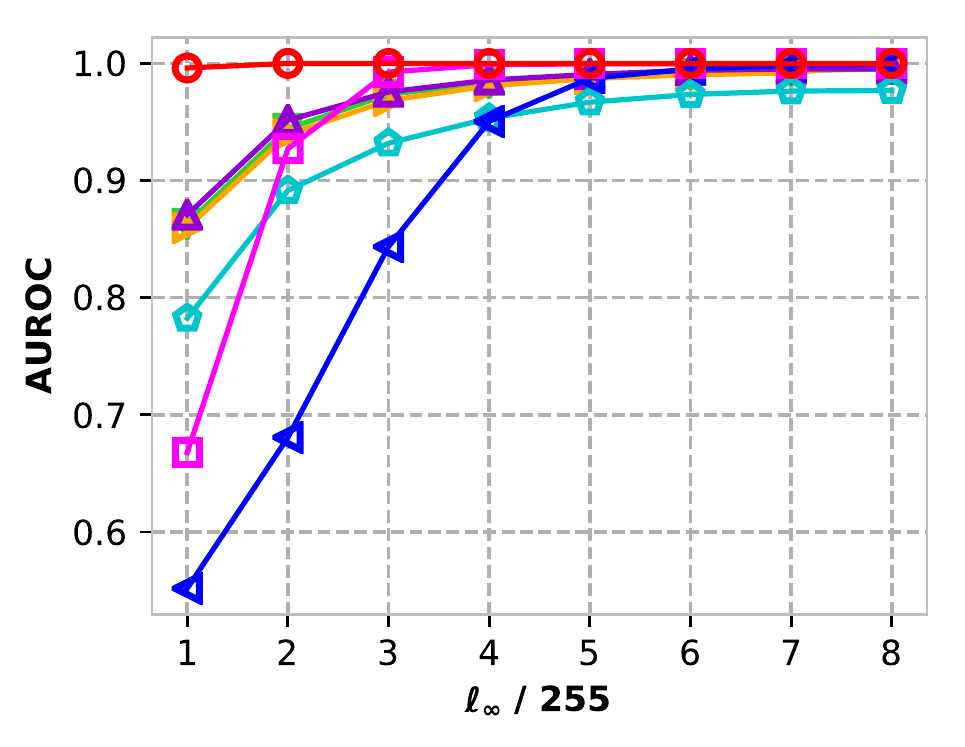}}
        \subfigure[Attack Method: TIM]
        {\includegraphics[width=0.32\textwidth]{ 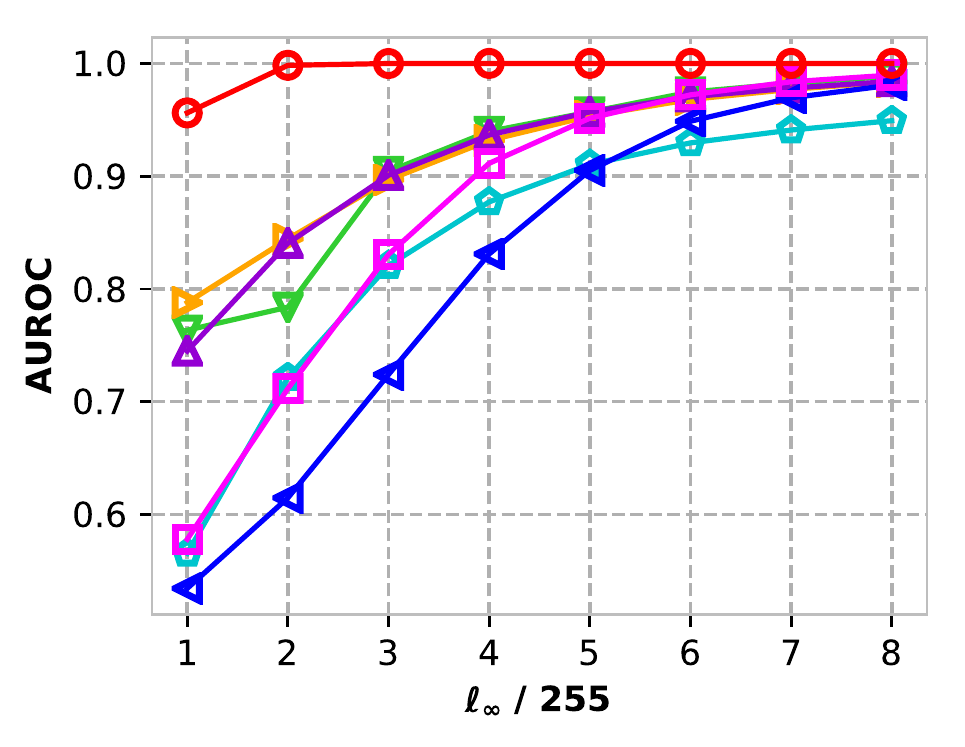}}

        \vspace{-1em}
        \caption{ More Results of adversarial detection on ImageNet. Sub-figures (a) - (f) report the AUROC on different attacks under  {$\epsilon {\in}\{1/255,{\ldots},8/255\}$} and share the same legend in sub-figure (c). 
      }
        \label{appendix_fig:comparison_ImageNet}
    \end{center}
    \vspace{-1.5em}
\end{figure*}

\paragraph{More comparison results on basic setup.}
In Table \ref{appendix_additional AUROC:cifar and image}, we provide additional attack results including MIM, TIM, DI\_MIM, PGD-$\ell_2$, MM  on CIFAR-10 and ImageNet.
We observe that \mymethod~keeps dominant position under these attacks,  \mymethodwithnorm~and \yoonwithnorm~exhibit poor performance on ImageNet due to the fact that these two methods use the norm of vectors and thus overlook the rich information that can be deviated from the vector. Moreover, we find that in Figure \ref{appendix_fig:comparison_CIFAR} and Figure \ref{appendix_fig:comparison_ImageNet} , most adversarial detection methods suffer from the extremely low attack intensity (\eg, $\epsilon=1/255$). In contrast, our method \mymethod~still has promising detection performance (\yyf{refer to Table \ref{appendix_low_attack_indentsity_test} for more quantitative results}).

\begin{table}[ht]
    \caption{More results of \yyf{different adversarial detection} methods on CIFAR-10 and ImageNet in terms of AUROC under $\epsilon = 4/255$ (MIM, TIM, DI, MIM, PGD-$\ell_2$, MM). }
    \centering
    \small
    \begin{tabular*}{12.5cm}{@{}@{\extracolsep{\fill}}cc|ccccccc@{}}
    \toprule
        Dataset & Method & MIM & TIM & DI\_MIM & PGD-$\ell_{2}$ & MM \\ \midrule
        \multirow{6}{*}{CIFAR-10} & KD & 0.8863 & 0.8856 & 0.8459 & 0.9296 & 0.9116 \\
         & LID & 0.9135 & 0.8781 & 0.8638 & 0.9426 & 0.9380 \\
         & MD & 0.9890 & \textbf{0.9998} & 0.9846 & 0.9958 & 0.9836 \\
         & \yoonwithnorm & 0.9993 & 0.9985 & 0.9980 & \textbf{1.0000} & 0.9994 \\
         & \mymethodwithnorm~(Ours) & 0.9999 & 0.9996 & 0.9996 & \textbf{1.0000} & \textbf{0.9996} \\
         & \mymethod~(Ours) & \textbf{1.0000} & 0.9993 & \textbf{0.9999} & \textbf{1.0000} & 0.9995 \\ \midrule
        \multirow{7}{*}{ImageNet} & KD & 0.9669 & 0.9317 & 0.7402 & 0.9809 & 0.9377 \\
        & LID & 0.9665 & 0.9397 & 0.7658 & 0.9833 & 0.9575 \\
        & MD & 0.9664 & 0.9365 & 0.8612 & 0.9858 & 0.9585 \\
        & LiBRe & 0.9483 & 0.8772 & 0.9968 & 0.9532 & 0.8697 \\
        & \yoonwithnorm & 0.9087 & 0.8310 & 0.9143 & 0.9505 & 0.9090 \\
        & \mymethodwithnorm~(Ours) & 0.9778 & 0.9114 & 0.9779& 0.9989 & 0.9939 \\
        & \mymethod~(Ours) & \textbf{1.0000} & \textbf{1.0000} & \textbf{1.0000} & \textbf{1.0000} & \textbf{1.0000} \\ \bottomrule
    \end{tabular*}
    \label{appendix_additional AUROC:cifar and image}
\end{table}

\paragraph{More comparison results on unseen and transferable attacks.}
We also compare our \mymethod~with KD, LID  MD and LiBRe under $6$ additional unseen or transferable attacks (MIM, TIM, DI\_MIM, PGD-$\ell_2$, MM,  {VMI-FGSM \citep{wang2021enhancing})} to further evaluate the effectiveness of our method. In Table \ref{appendix_attitional_unseen_results} and Table \ref{appendix_attitional_transform_results}, Our approach consistently exhibits superior generalization compared to other baselines. 

\begin{table}[ht]
    \caption{More results of AUROC for detecting the unseen attacks (MIM, TIM, DI\_MIM, PGD-$\ell_2$, MM) on CIFAR-10. ``FGSM (seen)''  denotes the seen adversarial attack used for the training of KD, LID and MD.}
    \centering
    \small
    \begin{tabular*}{12cm}{@{}@{\extracolsep{\fill}}c|ccccccc@{}}
    \toprule
        Method & FGSM(seen) & MIM & TIM & DI\_MIM & PGD-$\ell_{2}$ & MM \\ \midrule
        KD & 0.9213 & 0.8867 & 0.8876 & 0.8549 & 0.9303 & 0.9114 \\
        LID & 0.9236 & 0.9131 & 0.8479 & 0.8631 & 0.9090 & 0.9244 \\
        MD & 0.9990 & 0.9858 & \textbf{0.9998} & 0.9791 & 0.9958 & 0.9829 \\
        \yoonwithnorm & \textbf{1.0000} & 0.9993 & 0.9985 & 0.9980 & \textbf{1.0000} & 0.9994 \\
        \mymethodwithnorm~(Ours) & \textbf{1.0000} & 0.9999 & {0.9996} & 0.9996 & \textbf{1.0000} & \textbf{0.9996} \\
        \mymethod~(Ours) & \textbf{1.0000} & \textbf{1.0000} & {0.9993} & \textbf{0.9999} & \textbf{1.0000} & {0.9995} \\ \bottomrule
    \end{tabular*}
    \label{appendix_attitional_unseen_results}
\end{table}

\begin{table}[ht]
    \caption{More results of AUROC for detecting the  transferable attacks (MIM, TIM, DI\_MIM, PGD-$\ell_2$, MM, VMI-FGSM) on ImageNet, where  KD, LID,  MD and LiBRe are trained with adversarial examples with ResNet-50 but detect the adversarial examples  crafted with ResNet-101.}
    \centering
    \small
    \begin{tabular*}{11cm}{@{}@{\extracolsep{\fill}}c|ccccccc@{}}
    \toprule
        Method & MIM & TIM & DI\_MIM & PGD-$\ell_{2}$ & MM & VMI-FGSM \\ \midrule
        KD & 0.6355 & 0.7006 & 0.7558 & 0.7028 & 0.7381 & 0.7669 \\
        LID & 0.7780 & 0.7978 & 0.7654 & 0.7734 & 0.7913 & 0.7386 \\
        MD & 0.7756 & 0.7612 & 0.8395 & 0.7827 & 0.7864 & 0.8039 \\
        \textit LiBRe & 0.8966 & 0.7317 & 0.9722 & 0.8749 & 0.8388 & 0.9944 \\
        \yoonwithnorm & 0.9049 & 0.8216 & 0.9117 & 0.9490 & 0.9043 & 0.8638 \\
        \mymethodwithnorm~(Ours) & 0.9771 & 0.9049 & 0.9765 & 0.9987 & 0.9930 & 0.9609 \\
        \mymethod~(Ours) & \textbf{1.0000} & \textbf{1.0000} & \textbf{1.0000} & \textbf{1.0000} & \textbf{1.0000} & \textbf{1.0000} \\ \bottomrule
    \end{tabular*}
    \label{appendix_attitional_transform_results}
\end{table}

\paragraph{More comparison results on CIFAR-10 over robust WideResNet-70-16 .}
We further compare our method with baselines on {an} adversarial trained classifier on CIFAR-10,  \eg, WideResNet-70-16 \citep{gowal2021improving} against various attacks. In Tables \ref{appendix_wrn_70_16_2pixel}, \ref{appendix_wrn_70_16_4pixel}, \yyf{We observe that diffusion-based detection methods are much better than other baselines trained with specific adversarial samples. One possible reason is that adversarial samples are difficult to deceive robust classifiers, which means that such adversarial samples are ineffective for training effective detectors.}

\begin{table}[th]
    \caption{Comparison of AUROC for using adversarial trained WideResNet-70-16 as classifier on CIFAR-10 under $\epsilon=2/255$. Due to the constraint of memory and resources, we omit the detection results on AutoAttack for KD, LID and MD.}
    \centering
    \small
    \begin{tabular*}{12.5cm}{@{}@{\extracolsep{\fill}}c|cccccc@{}}
    \toprule
        Method & KD & LID & MD & \yoonwithnorm & \mymethodwithnorm~ (Ours) & \mymethod~(Ours) \\ \midrule
        FGSM & 0.5852 & 0.7551 & 0.6924 & 0.9797 & 0.9976 & 0.9978 \\
        PGD & 0.5672 & 0.7517 & 0.6846 & 0.9625 & 0.9954 & 0.9961 \\
        BIM & 0.5786 & 0.7543 & 0.6787 & 0.9568 & 0.9935 & 0.9930 \\
        MIM & 0.5795 & 0.7544 & 0.6804 & 0.9679 & 0.9963 & 0.9957 \\
        TIM & 0.5812 & 0.7533 & 0.6783 & 0.8468 & 0.9506 & 0.9407 \\
        CW & 0.5559 & 0.7511 & 0.6830 & 0.9600 & 0.9949 & 0.9953 \\
        DI\_MIM & 0.5763 & 0.7505 & 0.6712 & 0.8877 & 0.9728 & 0.9700 \\
        PGD-$\ell_{2}$ & 0.6116 & 0.7632 & 0.7049 & 0.9942 & 0.9994 & 0.9998 \\
        FGSM-$\ell_{2}$  & 0.6114 & 0.7619 & 0.7032 & 0.9471 & 0.9766 & 0.9861 \\
        BIM-$\ell_{2}$  & 0.7550 & 0.7629 & 0.7060 & 0.9396 & 0.9756 & 0.9828 \\ \bottomrule
    \end{tabular*}
    \label{appendix_wrn_70_16_2pixel}
\end{table}

\begin{table}[th]
    \caption{Comparison of AUROC for using adversarial trained WideResNet-70-16 as classifier on CIFAR-10 under  $\epsilon=4/255$.}
    \centering
    \small
    \begin{tabular*}{12.5cm}{@{}@{\extracolsep{\fill}}c|cccccc@{}}
    \toprule
        Method & KD & LID & MD & \yoonwithnorm & \mymethodwithnorm~ (Ours) & \mymethod~(Ours) \\ \midrule
        FGSM & 0.6020 & 0.7628 & 0.7668 & 0.9999 & \textbf{1.0000} & \textbf{1.0000} \\
        PGD & 0.5913 & 0.7598 & 0.7535 & 0.9998 & \textbf{1.0000} & \textbf{1.0000} \\
        BIM & 0.6076 & 0.7617 & 0.7588 & 0.9994 & \textbf{1.0000} & \textbf{1.0000} \\
        MIM & 0.7683 & 0.7625 & 0.7601 & 0.9997 & \textbf{1.0000} & \textbf{1.0000} \\
        TIM & 0.6029 & 0.7605 & 0.7563 & 0.9917 & 0.9984 & \textbf{0.9987} \\
        CW & 0.5919 & 0.7581 & 0.7524 & 0.9998 & \textbf{1.0000} & \textbf{1.0000} \\
        DI\_MIM & 0.6015 & 0.7562 & 0.7551 & 0.9983 & 0.9998 & \textbf{0.9999} \\
        PGD-$\ell_{2}$ & 0.8314 & 0.7774 & 0.7544 & \textbf{1.0000} & \textbf{1.0000} & \textbf{1.0000} \\
        FGSM-$\ell_{2}$ & 0.7805 & 0.7709 & 0.7489 & 0.9803 & 0.9902 & \textbf{0.9967} \\
        BIM-$\ell_{2}$ & 0.8351 & 0.7739 & 0.8141 & 0.9835 & 0.9927 & \textbf{0.9955} \\ \bottomrule
    \end{tabular*}
    \label{appendix_wrn_70_16_4pixel}
\end{table}

\paragraph{More comparison results on ImageNet over DeiT-S.}
We further make an attempt on Vision-transformer-based architecture (\ie~DeiT-S \citep{dosovitskiy2021image}) on ImageNet. \yyf{Considering the specificity of the vision-transformer-based structure}, we compare our method to three baselines (LID, \yoonwithnorm~{and} \mymethodwithnorm). From Table \ref{appendix_deits_4pixel}, our method exhibits consistent superiority when compared to other baselines, suggesting the versatility of \mymethod~with different architectures.

\begin{table}[h]
    \caption{Comparison of AUROC for using DeiT-S as classifier on ImageNet under $\epsilon=4/255$.}
    \centering
    \small
    \begin{tabular*}{9cm}{@{}@{\extracolsep{\fill}}c|cccc@{}}
    \toprule
        Method & LID & \yoonwithnorm & \mymethodwithnorm~(Ours) & \mymethod~(Ours) \\ \midrule
        FGSM & 0.8846 & 0.9789 & 0.9984 & \textbf{1.0000} \\
        PGD & 0.9162 & 0.8935 & 0.9969 & \textbf{1.0000} \\
        BIM & 0.9191 & 0.7331 & 0.9215 & \textbf{1.0000} \\
        MIM & 0.9102 & 0.9025 & 0.9780 & \textbf{1.0000} \\
        TIM & 0.9019 & 0.8091 & 0.8765 & \textbf{0.9606} \\
        CW & 0.8742 & 0.8934 & 0.9975 & \textbf{0.9999} \\
        DI\_MIM & 0.7246 & 0.9074 & 0.9752 & \textbf{1.0000} \\
        PGD-$\ell_{2}$ & 0.9041 & 0.9451 & 0.9987 & \textbf{1.0000} \\
        FGSM-$\ell_{2}$ & 0.8698 & 0.7665 & 0.7023 & \textbf{1.0000} \\
        BIM-$\ell_{2}$ & 0.9002 & 0.6564 & 0.6544 & \textbf{1.0000} \\
        MM & 0.9164 & 0.8902 & 0.9886 & \textbf{0.9993} \\
        AA  & 0.9191  & 0.9023  & 0.9915  & \textbf{1.0000} \\ \bottomrule
    \end{tabular*}
    \label{appendix_deits_4pixel}
\end{table}

 {\paragraph{More comparison results with rejection option method.}
We further compare our method with R-Rejection \citep{pang2022two} against FGSM attack ($\epsilon=4/255$) on CIFAR-10. From Table \ref{appendix_comparison_with_rejection_method}, we find that our method \mymethod~outperforms the baseline by a large margin, demonstrating the superiority of our method.}

\begin{table}[!ht]
    \caption{ {Comparison with R-Rejection \citep{pang2022two} on CIFAR-10 in terms of AUROC against FGSM under $\epsilon = 4/255$.}}
    \centering
    \small
    \begin{tabular*}{14.7cm}{@{}@{\extracolsep{\fill}}c|cccccccc@{}}
    \toprule
        Method & PGD & PGD-$\ell_{2}$ & FGSM & FGSM-$\ell_{2}$ & BIM & BIM-$\ell_{2}$ & AA & MM \\ \midrule
        R-Rejection & 0.9260 & 0.9420 & 0.8490 & 0.8690 & 0.7610 & 0.8390 & 0.8970 & 0.9080 \\
        \mymethodwithnorm~(Ours) & \textbf{1.0000} & \textbf{1.0000} & \textbf{1.0000} & 0.9916 & 0.9996 & 0.9883 & \textbf{1.0000} & \textbf{0.9996} \\
        \mymethod~(Ours) & \textbf{1.0000} & \textbf{1.0000} & \textbf{1.0000} & \textbf{0.9995} & \textbf{0.9998} & \textbf{0.9991} & \textbf{1.0000} & 0.9995 \\ \bottomrule
    \end{tabular*}
    \label{appendix_comparison_with_rejection_method}
\end{table}

\newpage
\subsection{More Discussions of \mymethod}
\label{subsec:ablation}

\begin{figure*}[tp]
    \begin{center}
        \subfigure[Impact of set size on ImageNet]
        {\includegraphics[width=0.32\textwidth]{ 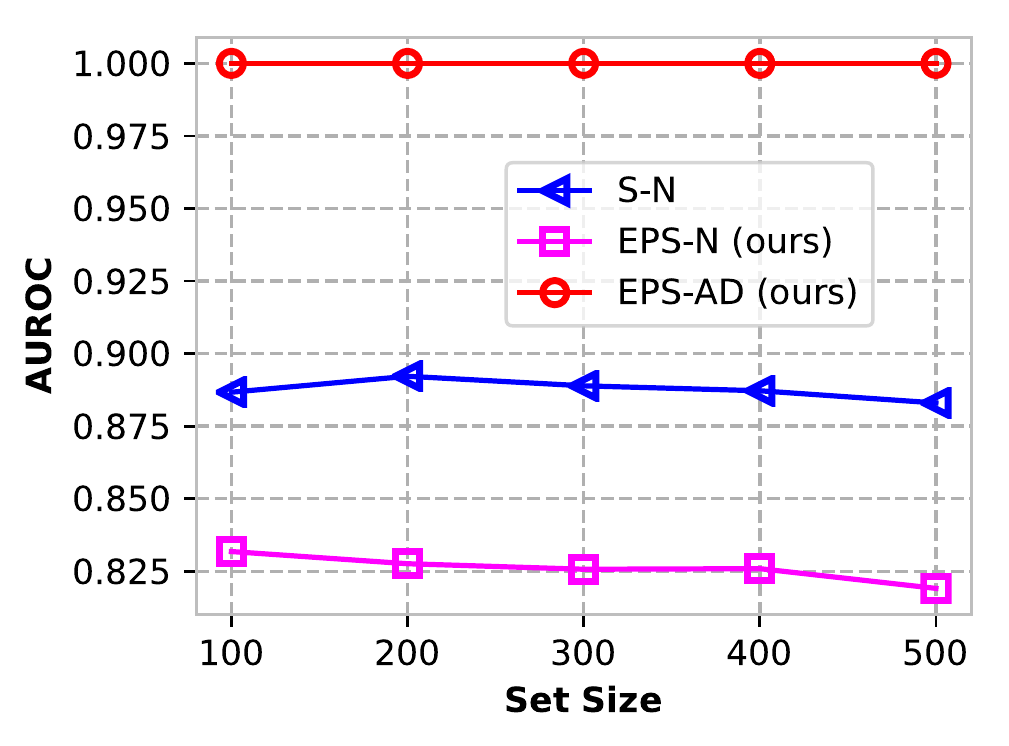}}
        \subfigure[Impact of set size on CIFAR-10]
        {\includegraphics[width=0.32\textwidth]{ 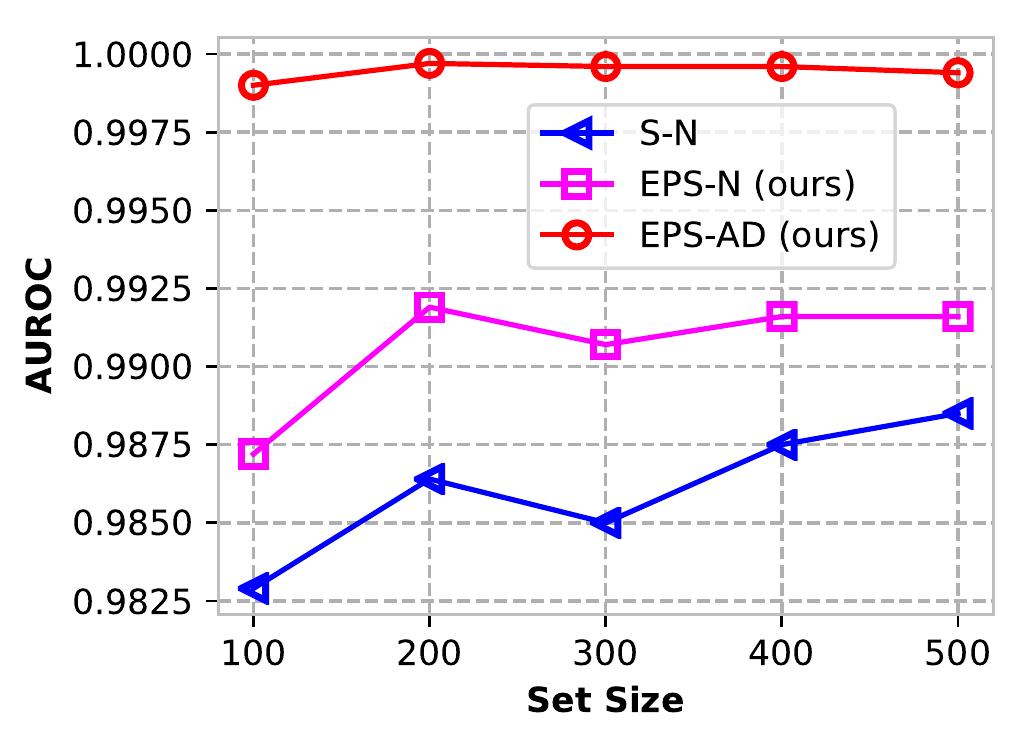}}
        \subfigure[Impact of timestep]
        {\includegraphics[width=0.32\textwidth]{ 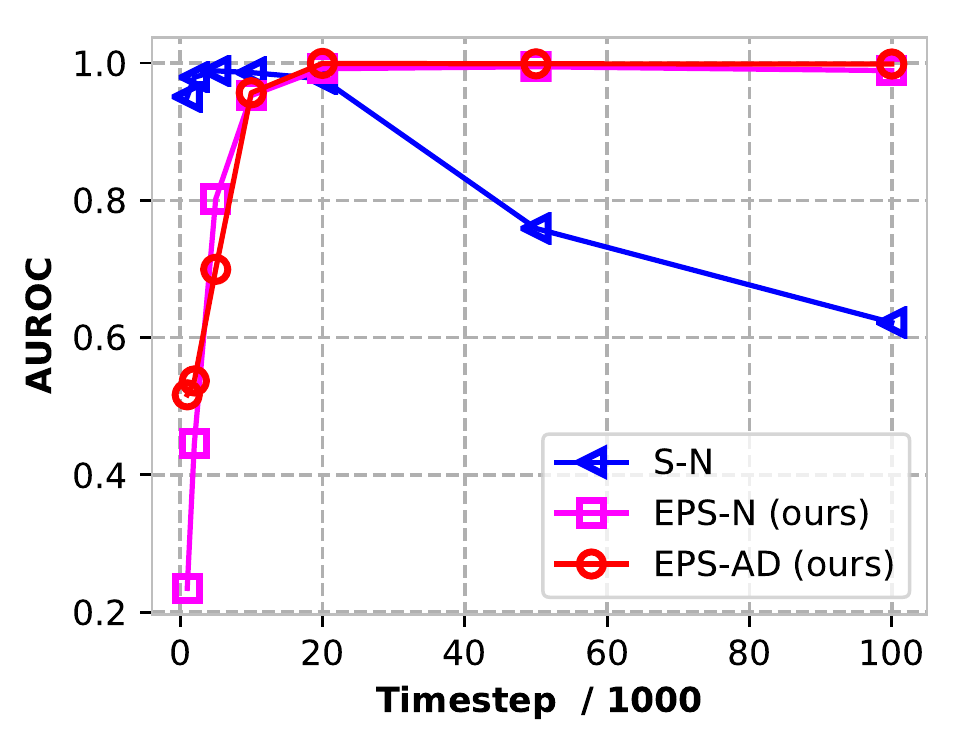}}
        \vspace{-1em}
        \caption{ Impact of different set sizes and diffusion time step. Sub-figures (a) and (b) report the AUROC on FGSM-$\ell_2$ attack under \yyf{$\epsilon =4/255$}. Sub-figure (c) reports the AUROCs of different diffusion time in \{1, 2, 5, 10, 20, 50, 100\} on CIFAR-10 dataset.
      }
        \label{appendix_fig:ablation_on_setsize}
    \end{center}
    \vspace{-1.5em}
\end{figure*}

\paragraph{Ablation of timestep.}
The ablation study of timestep is provided in Table \ref{appendix_impact_of_diff_t_cifar}  {and Table \ref{appendix_impact_of_diff_t_imagenet}}. We observe that when the timestep $T>10$, both \mymethod~and  \mymethodwithnorm~obtain usable AUROC in detecting FGSM-$\ell_2$ attack, which verifies that our method is insensitive to the timestep. 
 {Meanwhile, our methods consistently outperform the baseline S-N when $T>20$, demonstrating the superiority of our methods.}
Moreover, Our approaches \mymethod~and \mymethodwithnorm~achieve optimum performance when $T^*=20$ on CIFAR-10 and $T^*=50$ on ImageNet, while the method \yoonwithnorm~achieves optimum when $T^*=5$ on CIFAR-10 and $T^*=20$ on ImageNet.

\begin{table}[!ht]
    \caption{ {Impact of timestep with WideResNet-28-10 on CIFAR-10 against FGSM-$\ell_2$ under $\epsilon = 4/255$.}}
    \centering
    \small
    \begin{tabular*}{14cm}{@{}@{\extracolsep{\fill}}c|ccccccccc@{}}
    \toprule
    \rule{0pt}{12pt}
         \diagbox{Method}{timestep} & 1 & 2 & 5 & 10 & 20 & 50 & 100 & 200 \\ \hline
        \yoonwithnorm & 0.9508 & 0.9792 & \textbf{0.9885} & 0.9860 & 0.9768 & 0.7590 & 0.6218 & 0.5597 \\
        \mymethodwithnorm~(Ours) & 0.2346 & 0.4454 & 0.8018 & 0.9526 & \textbf{0.9916} & 0.9915 & 0.9889 & 0.9814 \\
        \mymethod~(Ours) & 0.5166 & 0.5368 & 0.6994 & 0.9566 & \textbf{0.9994} & 0.9988 & 0.9985 & 0.9943 \\ \midrule
        \diagbox{Method}{timestep} & 300 & 400 & 500 & 600 & 700 & 800 & 900 & 1000 \\ \hline
        \yoonwithnorm & 0.5383 & 0.5274 & 0.5210 & 0.5172 & 0.5151 & 0.5147 & 0.5146 & 0.5144 \\
        \mymethodwithnorm~(Ours) & 0.9766 & 0.9729 & 0.9705 & 0.9689 & 0.9680 & 0.9675 & 0.9672 & 0.9671 \\
        \mymethod~(Ours) & 0.9921 & 0.9918 & 0.9910 & 0.9904 & 0.9896 & 0.9887 & 0.9878 & 0.9870 \\ \bottomrule
    \end{tabular*}
    \label{appendix_impact_of_diff_t_cifar}
\end{table}

\begin{table}[!ht]
    \caption{ {Impact of timestep with ResNet-50 on ImageNet against FGSM-$\ell_2$ under $\epsilon = 4/255$.}}
    \centering
    \small
    \begin{tabular*}{12cm}{@{}@{\extracolsep{\fill}}c|cccccccc@{}}
    \toprule
    \rule{0pt}{12pt}
        \diagbox{Method}{timestep} & 1 & 2 & 5 & 10 & 20 & 50 & 100 \\ \hline
        \yoonwithnorm & 0.4794 & 0.5298 & 0.7168 & 0.8528 & \textbf{0.8830} & 0.7309 & 0.5964 \\
        \mymethodwithnorm~(Ours) & 0.2246 & 0.2466 & 0.4369 & 0.6215 & 0.7484 & \textbf{0.8191} & 0.8190 \\
        \mymethod~(Ours) & 0.5112 & 0.5545 & 0.5662 & 0.6912 & 0.9917 & \textbf{1.0000} & 0.9930 \\ \bottomrule
    \end{tabular*}
    \label{appendix_impact_of_diff_t_imagenet}
\end{table}

\paragraph{Impact of set size.}
Previous adversarial detection methods usually measure the discrepancy well only with large amount of data \citep{gao2021maximum}. To show the effectiveness of our proposed \mymethod, in this experiment, we further ablate the effect of set size by conducting experiments on 100 to 500 samples subset of CIFAR-10 with WideResNet-28-10 and ImageNet with ResNet-50. Performance of our three methods, \yoonwithnorm, \mymethodwithnorm~and \mymethod, is shown in Table \ref{appendix_impact_of_set_size} and Figure \ref{appendix_fig:ablation_on_setsize}. It is obvious that \mymethod~consistently outperforms \mymethodwithnorm~and \yoonwithnorm~with small set size and large set size. Moreover, \mymethod~is robust to the changes of set size while \mymethodwithnorm~and \yoonwithnorm~fluctuate with set size, especially on CIFAR-10 dataset.

\begin{table}[ht]
    \caption{Impact of data set size with WideResNet-28-10 on CIFAR-10 and ResNet-50 on ImageNet against FGSM-$\ell_2$ under $\epsilon = 4/255$.}
    \centering
    \small
    \begin{tabular*}{11.7cm}{@{}@{\extracolsep{\fill}}c|c|cccccccc@{}}
    \toprule
         Dataset & \diagbox{Method}{size} & 100 & 200 & 300 & 400 & 500 \\ \midrule
        \multirow{3}{*}{CIFAR-10} & \yoonwithnorm & 0.9829 & 0.9864 & 0.9850 & 0.9875 & 0.9885 \\
        & \mymethodwithnorm~(Ours) & 0.9872 & 0.9919 & 0.9907 & 0.9916 & 0.9916 \\
        & \mymethod~(Ours) & 0.9990 & 0.9997 & 0.9996 & 0.9996 & 0.9994 \\ \midrule
        \rule{0pt}{10pt}
        \multirow{3}{*}{ImageNet} & \yoonwithnorm & 0.8867 & 0.8922 & 0.8889 & 0.8872 & 0.8830 \\
        & \mymethodwithnorm~(Ours) & 0.8318 & 0.8276 & 0.8257 & 0.8259 & 0.8191 \\
        & \mymethod~(Ours) & 1.0000 & 1.0000 & 1.0000 & 1.0000 & 1.0000 \\ \bottomrule
    \end{tabular*}
    \label{appendix_impact_of_set_size}
\end{table}

\paragraph{Detecting on low attack intensity.}
To further reveal the superiority of our \mymethod, we conduct an experiment under an extremely low attack intensity (\eg, $\epsilon=1/255$) on ImageNet. In Table \ref{appendix_low_attack_indentsity_test}, our \mymethod~achieves a significant advantage in detecting adversarial samples crafted with extremely low attack intensity, demonstrating its significant effectiveness.

\paragraph{Detecting on adversarial samples across datasets.}
We further exploit the transferability across different {datasets}. To this end, we utilize a pre-trained score-based diffusion model on ImageNet to perform detecting adversarial samples from CIFAR-10. Specifically, we randomly select two disjoint subsets as adversarial and natural samples (each containing 500 samples) from CIFAR-10 and use a score model pre-trained on ImageNet to calculate the AUROC, which is named \mymethod$^{*}$. Table \ref{appendix_cross_dataset_test} demonstrates detection performance of $6$ methods against $12$ attacks under $\epsilon = 2/255$ on CIFAR-10 over WideResNet-28-10. We observe that \mymethod$^{*}$ still exhibits superior detection performance compared to KD, LID, MD baselines, and achieves a comparable performance compared to other diffusion-based methods that use the score model pre-trained on CIFAR-10.

 {\paragraph{Results of \mymethod~with $5$ different random seeds.}
We run our \mymethod~on CIFAR-10 and ImageNet against FGSM-$\ell_2$ under $\epsilon = 4/255$ with $5$ different random seeds and report the standard deviation of AUROC in Table \ref{multiple_runn_standard_deviation}. From the results, our \mymethod~has a very small standard error, indicating the consistency and repeatability of our methods.}

\begin{table}[ht]
    \caption{ {Standard deviation of AUROC with $5$ different random seeds over WideResNet-28-10 on CIFAR-10 and ResNet-50 on ImageNet against FGSM-$\ell_2$ under $\epsilon = 4/255$.}}
    \centering
    \small
    \begin{tabular*}{8.5cm}{@{}@{\extracolsep{\fill}}c|cc@{}}
    \toprule
        Dataset & CIFAR-10 & ImageNet \\ \midrule
        PGD & $1.0000 \pm 0$ & $0.9999 \pm 4 \times 10^{-5}$ \\ %
        \rule{0pt}{10pt}
        FGSM & $1.0000 \pm 0$ & $1.0000 \pm 0$ \\
        BIM & $0.9999 \pm 8.94 \times 10^{-5}$ & $1.0000 \pm 0$ \\
        MIM & $1.0000 \pm 0$ & $1.0000 \pm 0$ \\
        TIM & $0.9996 \pm 1.67 \times 10^{-4}$ & $0.9999 \pm 1.26 \times 10^{-4}$ \\
        DI\_MIM & $0.9999 \pm 4 \times 10^{-5}$ & $1.0000 \pm 0$ \\
        CW & $1.0000 \pm 0$ & $0.9999 \pm 4 \times 10^{-5}$ \\
        PGD-$\ell_{2}$ & $1.0000 \pm 0$ & $1.0000 \pm 0$ \\
        FGSM-$\ell_{2}$ & $0.9996 \pm 2.45 \times 10^{-4}$ & $1.0000 \pm 0$ \\
        BIM-$\ell_{2}$ & $0.9992 \pm 2.97 \times 10^{-4}$ & $1.0000 \pm 0$ \\
        MM & $0.9967 \pm 3.08 \times 10^{-3}$ & $1.0000 \pm 0$ \\
        AA & $0.9997 \pm 6.4 \times 10^{-4}$ & $1.0000 \pm 0$ \\ \bottomrule
    \end{tabular*}
    \label{multiple_runn_standard_deviation}
\end{table}

 {\paragraph{Robustness of \mymethod~against adaptive attacks.}
To verify the robustness of our approach to adaptive attacks, we conduct the experiment on CIFAR-10 under PGD attack in a white-box setting, \ie, both the classifier and the detector are available to the attacker. Specifically, we follow the strategy in \citet{ma2018characterizing} to construct a modified attack objective: $\operatorname{max} \ell(\hat{f}(\hat{\mathbf{x}}), l) - \alpha {\widehat{\mathrm{MMD}}_{b}}^{2}\left(\left\{\mathbf{x}^{(i)}\right\}_{i=1}^{n}, \left\{\hat{\mathbf{x}}\right\} \right)$,
where $\alpha$ is a constant balancing between the attack of the classifier and detector. When $\alpha$ increases, the attack successful rate (ASR) of PGD drops severely (when $\alpha=0$, $ASR=0.96$; when $\alpha=1$, $ASR=0.04$), indicating that the difficulty of attack increases significantly. Conversely, from Table \ref{appendix_against_adaptive_attack}, our \mymethod~achieves a high AUROC of $0.9154$ even with a large $\alpha=1$. The results demonstrate the robustness of our approach to adaptive attacks. The reason is that the minimization of the MMD between adversarial samples and natural samples gradually narrows the distributional discrepancy between natural and crafted adversarial samples, making it harder for the adversarial samples to mislead a trained classifier. This phenomenon is consistent with the discussion in \citet{ma2018characterizing}.}

\begin{table}[ht]
    \caption{ {AUROC of EPS-AD under different $\alpha$ against an adaptive attack (PGD) under $\epsilon = 4/255$ on CIFAR-10.}}
        \centering
    \small
    \begin{tabular*}{14.7cm}{@{}@{\extracolsep{\fill}}c|cccccccccc@{}}
    \toprule
    \rule{0pt}{12pt}
        \diagbox{Method}{$\alpha$} & $0$ & $10^{-7}$ & $10^{-6}$ & $10^{-5}$ & $10^{-4}$ & $10^{-3}$ & $10^{-2}$ & $10^{-1}$ & $1$ \\ \midrule
        \mymethod~(Ours) & 1.0000 & 0.9859 & 0.9681 & 0.9353 & 0.9303 & 0.9284 & 0.9168 & 0.9145 & 0.9154 \\ \bottomrule
    \end{tabular*}
    \label{appendix_against_adaptive_attack}
\end{table}

\paragraph{Inference efficiency of \mymethod.}
 {Given a fixed-size model, the computational cost of our method (\mymethod) mainly depends on two factors: the resolution of input images and total diffusion timestep $T$. Actually, our \mymethod~performs adversarial detection efficiently, especially on low-resolution images, and yields promising performance compared to existing methods. To evaluate the efficiency of \mymethod, we randomly choose $500$ images from CIFAR-10 and ImageNet respectively in detecting FGSM-$\ell_2$ adversarial samples on a single RTX3090 GPU. The average time costs per image for CIFAR-10 ($T=20$) and ImageNet ($T=50$) are $0.043s$ and $2.369s$, respectively. }

 {To further demonstrate the effect of the resolution of input images and total diffusion timestep $T$  on the efficiency, we provide the results of different resolutions of input images under different diffusion timesteps against FGSM-$\ell_2$ attack. As shown in Table \ref{appendix_time_imagenet}, our EPS-AD with resolution $256 \times 256$ shows superior adversarial detection performance when $20 \le T \le 100$ on ImageNet and takes 0.954s when $T = 20$, which is much more efficient than that with $T = 50$.  When equipped with lower-resolution input size $128 \times 128$, our EPS-AD stills keeps superior performance on ImageNet when $T = 50$ and only takes $0.780s$.  
 Moreover, in Table \ref{appendix_time_cifar}, the average time costs per image of our \mymethod~on CIFAR-10 with $T=10$ and $T=20$ are superior or comparable to existing methods (i.e., KD, LID, and MD), as well as the detection performance.
}

 {Note that the computational efficiency of our method can be further improved by applying an efficient sampling strategy \citep{lu2022dpm}, a low-resolution diffusion model \citep{dhariwal2021diffusion} and a sparse diffusion timestep (e.g., sampling with a time interval of 2/1000 during the diffusion process). We leave these techniques for our future work.}

\begin{table}[!ht]
    \caption{Results of adversarial detection performance (AUROC \& time costs per image) against FGSM-$\ell_{2}$ attack ($\epsilon=4/255$) under different total diffusion timesteps on CIFAR-10. 
    }
    \centering
    \small
    \begin{tabular*}{11.3cm}{@{}@{\extracolsep{\fill}}c|ccc|c|c@{}}
    \toprule
         {Dataset} & KD & LID & MD & Timestep & \mymethod \\ \midrule
        \multirow{5}{*}{CIFAR-10} & \multirow{5}{*}{\makecell[c]{0.9121 \\ (0.019s)}} & \multirow{5}{*}{\makecell[c]{0.9169 \\ (0.012s)}} & \multirow{5}{*}{\makecell[c]{\textbf{0.9995} \\ (0.029s)}} & $T=5$ & 0.6994 (0.021s) \\
        &  &  &  & $T=10$ & 0.9566 (0.028s) \\
        &  &  &  & $T=20$ & \textbf{0.9995} (0.043s) \\
        &  &  &  & $T=50$ & 0.9988 (0.090s) \\
        &  &  &  & $T=100$ & 0.9985 (0.168s) \\ \bottomrule
    \end{tabular*}
    \label{appendix_time_cifar}
    \vspace{-10pt}
\end{table}

\begin{table}[!ht]
    \caption{Results of adversarial detection performance (AUROC \& time costs per image) against FGSM-L2 attack ($\epsilon=4/255$) under different total diffusion timesteps on ImageNet, where ``\mymethod~($128 \times 128$)'' and  ``\mymethod~( $256 \times 256$)'' denote \mymethod~with different resolution inputs. 
    }
    \centering
    \small
    \begin{tabular*}{15.5cm}{@{}@{\extracolsep{\fill}}c|cccc|c|cc@{}}
    \toprule
         {Dataset} & KD & LID & MD & LiBRe & Timestep & \mymethod~($128 \times 128$) & \mymethod~( $224 \times 224$) \\ \midrule
        \multirow{5}{*}{ImageNet} & \multirow{5}{*}{\makecell[c]{0.7004 \\ (0.029s)}} & \multirow{5}{*}{\makecell[c]{0.8932 \\ (0.020s)}} & \multirow{5}{*}{\makecell[c]{0.8715 \\ (0.069s)}} & \multirow{5}{*}{\makecell[c]{0.8708 \\ (0.029s)}} & $T=5$ & 0.4652 (0.112s) & 0.5662 (0.268s) \\
        &  &  &  &  & $T=10$ & 0.6385 (0.187s) & 0.6912 (0.498s) \\
        &  &  &  &  & $T=20$ & 0.7501 (0.335s) & 0.9917 (0.992s) \\
        &  &  &  &  & $T=50$ & 0.9991 (0.780s) & \textbf{1.0000} (2.369s) \\
        &  &  &  &  & $T=100$ & 0.9392 (1.519s) & 0.9930 (4.696s) \\ \bottomrule
    \end{tabular*}
    \label{appendix_time_imagenet}
    \vspace{-10pt}
\end{table}

\begin{table}[ht]
    \caption{Comparision of \yyf{different adversarial detection} methods with attack intensity $\epsilon = 1/255$ over ResNet-50 on ImageNet.}
    \centering
    \small
    \begin{tabular*}{13.5cm}{@{}@{\extracolsep{\fill}}c|ccccccc@{}}
    \toprule
        Method & KD & LID & MD & LiBRe & \yoonwithnorm & \mymethodwithnorm~ (Ours) & \mymethod~(Ours) \\ \midrule
        FGSM & 0.7301 & 0.7765 & 0.7694 & 0.5653 & 0.5654 & 0.6942 & \textbf{0.9982} \\
        PGD & 0.8167 & 0.7651 & 0.8116 & 0.6938 & 0.5348 & 0.6149 & \textbf{0.9637} \\
        BIM & 0.8313 & 0.7711 & 0.8303 & 0.7154 & 0.5240 & 0.5619 & \textbf{0.9845} \\
        MIM & 0.8079 & 0.7711 & 0.7313 & 0.6597 & 0.5411 & 0.6144 & \textbf{0.9972} \\
        TIM & 0.7879 & 0.7634 & 0.7449 & 0.5655 & 0.5344 & 0.5780 & \textbf{0.9561} \\
        CW & 0.8161 & 0.7690 & 0.8336 & 0.6996 & 0.5345 & 0.6169 & \textbf{0.9549} \\
        DI\_MIM & 0.7498 & 0.7498 & 0.7543 & 0.5538 & 0.5364 & 0.6052 & \textbf{0.9817} \\
        PGD-$\ell_{2}$ & 0.8597 & 0.8633 & 0.8707 & 0.7824 & 0.5519 & 0.6678 & \textbf{0.9961} \\
        FGSM-$\ell_{2}$ & 0.7265 & 0.7737 & 0.7704 & 0.5727 & 0.5546 & 0.5978 & \textbf{0.9997} \\
        BIM-$\ell_{2}$ & 0.8596 & 0.8602 & 0.8644 & 0.7701 & 0.5199 & 0.5462 & \textbf{0.9927} \\
        MM & 0.8505 & 0.8730 & 0.8786 & 0.6611 & 0.5340 & 0.6106 & \textbf{0.9928} \\
        AA  & 0.8535  & 0.8706  & 0.8733  & 0.6629  & 0.5351  & 0.6113  & \textbf{0.9936} \\ \bottomrule
    \end{tabular*}
    \label{appendix_low_attack_indentsity_test}
\end{table}

\begin{table}[ht]
    \caption{Comparison of cross-dataset \mymethod$^*$ under attack intensity $\epsilon = 2/255$ with other methods over WideResNet-28-10 on CIFAR-10, where \mymethod$^*$ utilizes a score model pre-trained on ImageNet.}
    \centering
    \small
    \begin{tabular*}{13cm}{@{}@{\extracolsep{\fill}}c|ccccccc@{}}
    \toprule
        Method & KD & LID & MD & \yoonwithnorm & \mymethodwithnorm~(Ours) & \mymethod~(Ours) & \mymethod$^* $ \\ \midrule
        PGD & 0.8871 & 0.8836 & 0.8815 & 0.9679 & 0.9950 & 0.9972 & 0.9987 \\
        FGSM & 0.9112 & 0.9010 & 0.9200 & 0.9902 & 0.9987 & 0.9994 & 0.9961 \\
        BIM & 0.8786 & 0.8878 & 0.8811 & 0.9268 & 0.9811 & 0.9914 & 0.9890 \\
        MIM & 0.8873 & 0.8909 & 0.8947 & 0.9676 & 0.9935 & 0.9975 & 0.9964 \\
        TIM & 0.8983 & 0.8735 & 0.9655 & 0.9603 & 0.9878 & 0.9747 & 0.9906 \\
        CW & 0.8634 & 0.8762 & 0.8968 & 0.9682 & 0.9953 & 0.9975 & 0.9971 \\
        DI\_MIM & 0.7810 & 0.7351 & 0.8153 & 0.9348 & 0.9764 & 0.9942 & 0.9918 \\
        PGD-$\ell_{2}$ & 0.8727 & 0.8935 & 0.9128 & 0.9931 & 0.9989 & 0.9997 & 0.9994 \\
        FGSM-$\ell_{2}$ & 0.9044 & 0.8927 & 0.9211 & 0.9166 & 0.9634 & 0.9976 & 0.9973 \\
        BIM-$\ell_{2}$ & 0.8675 & 0.8864 & 0.8946 & 0.8564 & 0.9333 & 0.9922 & 0.9896 \\
        MM & 0.8627 & 0.8843 & 0.8404 & 0.9497 & 0.9706 & 0.9727 & 0.9658 \\
        AA & 0.8754 & 0.8894 & 0.8392 & 0.9595 & 0.9775 & 0.9819 & 0.9782 \\ \bottomrule
    \end{tabular*}
    \label{appendix_cross_dataset_test}
\end{table}

\end{document}